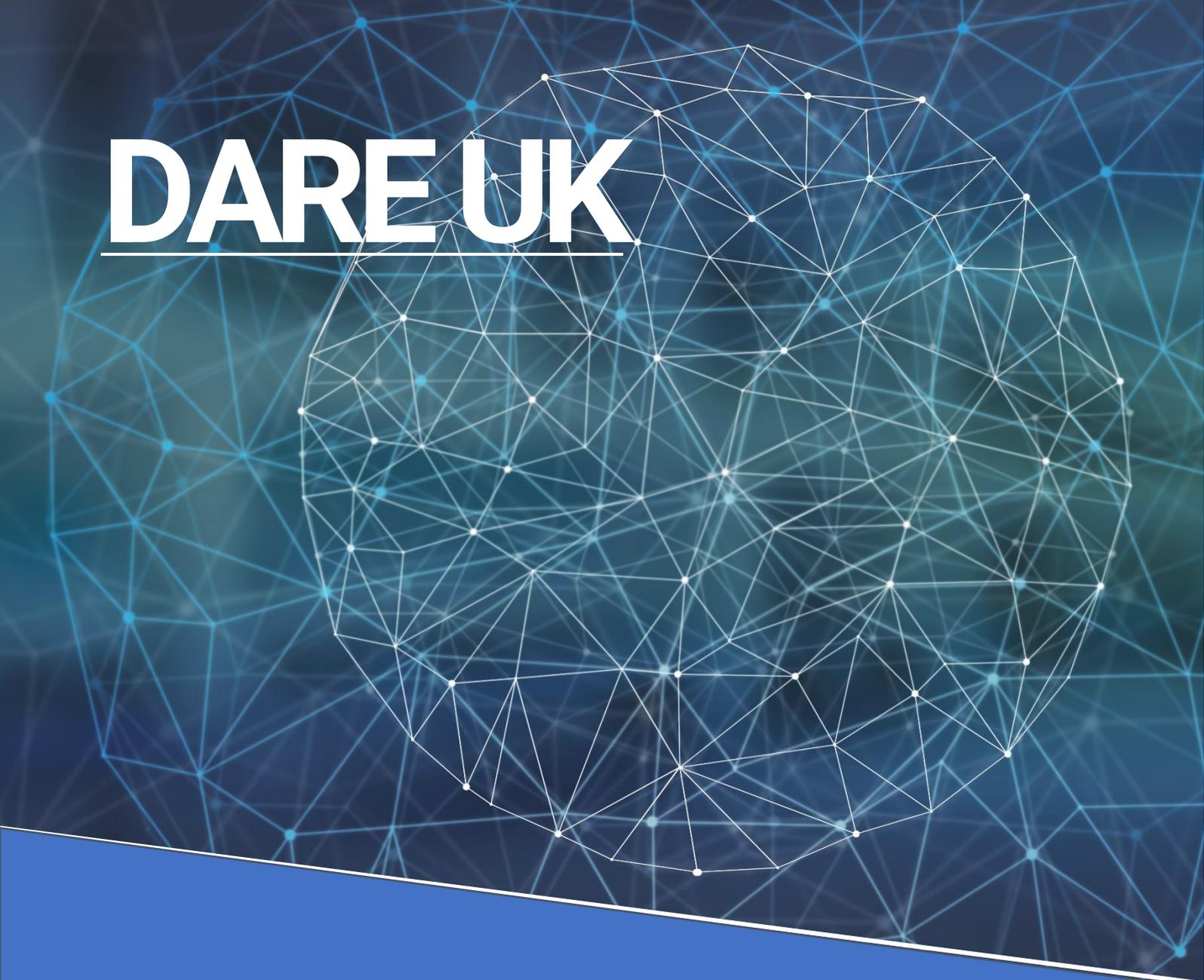

# Green Paper: Recommendations for disclosure control of trained Machine Learning (ML) models from Trusted Research Environments (TREs)

**DARE UK**

An output from the Guidelines and Resources for AI Model Access from TrusTEd Research environments (GRAIMatter) DARE UK Sprint Project

# 1 Contents

## 2 Current Status of Recommendations

To date, the development of these recommendations has been funded by the GRAIMATTER UKRI DARE UK sprint research project. This version of our recommendations was published at the end of the project in September 2022. During the course of the project, we have identified many areas for future investigations to expand and test these recommendations in practice. Therefore, we expect that this document will evolve over time.

Please contact the team if you would like to provide feedback or would like to learn more.

## 3 Recommendations Team

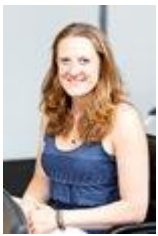

**Professor Emily Jefferson (Principal Investigator):** Director of Health Informatics Centre Trusted Research Environment (HIC TRE) and Professor of Health Data Science, University of Dundee

**Emily Jefferson** has run a Safe Haven/TRE for a decade and a major area of her research portfolio is the development of methods to enhance Safe Havens to support next-generation capabilities (such as handling of big data, scaling and performance as well as support for AI).

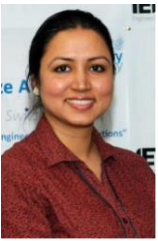

**Dr Smarti Reel (Project Manager):**

**Smarti Reel** is a postdoctoral researcher in the Health Informatics Centre (HIC), School of Medicine at the University of Dundee. Her research interests include machine learning and its use in the domain of medicine, imaging, and social media. She has worked on multi-omics biomarker discovery, multi-view image synthesis, broader image processing and other computing-based applications. She is experienced in the design, development, and evaluation of STEM projects.

**Work Package 1 - Risk assessment of AI models**

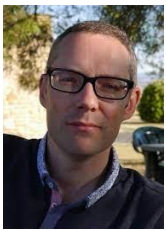

**Dr Christian Cole:** Senior Lecturer in Health Informatics, University of Dundee

**Christian Cole** is a senior health informatician and co-leads the research team on the development of the TRE Federation in Scotland and the development of next-generation capabilities enabling genomics, genetics and large data projects within HIC's Cloud TRE. He has 15 years of research experience in bioinformatics and data science.

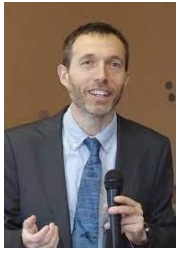

### Professor Josep Domingo-Ferrer

**Josep Domingo-Ferrer** is an international expert on information privacy and security and their interplay with AI: how to use AI to improve data protection/statistical disclosure/model disclosure control, and how to ensure privacy and security in machine learning. He is a distinguished professor of computer science at Universitat Rovira i Virgili, Tarragona, Catalonia.

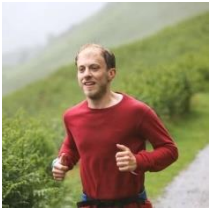

### Dr Simon Rogers: Principal Engineer - AI Models

**Simon Rogers** is currently a principal engineer within the NHS National Services Scotland Artificial Intelligence Centre of Excellence, where he supports the development and deployment of AI solutions across NHS Scotland. Before this, he spent over 10 years as a lecturer/senior lecturer researching the development and application of ML methods.

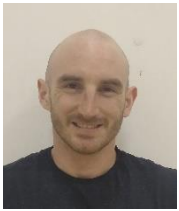

### Dr James Liley: Assistant Professor in biostatistics

**James Liley** is an assistant professor in biostatistics and has extensive experience in developing ML models in TRE environments through two years of work with the Alan Turing Institute and Public Health Scotland.

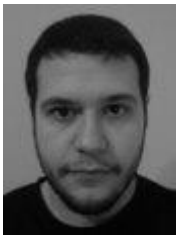

### Dr Alberto Blanco Justicia:

**Alberto Blanco-Justicia** is a senior postdoctoral researcher at Universitat Rovira i Virgili, Tarragona, Catalonia. His research interests include data privacy and security, privacy-enhancing techniques, and ethically-aligned machine learning, specifically including robustness, privacy, security and interpretability of (distributed) ML models.

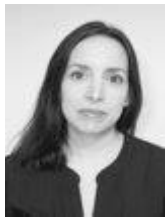

### Dr Esma Mansouri-Benssassi:

**Esma Mansouri-Benssassi** is a senior research fellow at HIC, the University of Dundee, working on several research topics such as dementia prediction, safe disclosure of machine learning models, and medical image feature extraction.

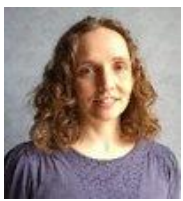

### Alba Crespi Boixader:

**Alba Crespi Boixader** is a postdoctoral researcher at HIC, University of Dundee. She carried out her PhD research in bioinformatics at the School of Informatics, University of Edinburgh, applying machine learning methods to genomic data.

**Work Package 2 (WP2) - Assessment of tools**

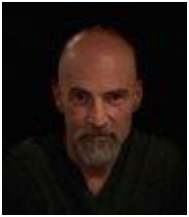

Professor Jim Smith: AI Models, the University of the West of England

**Jim Smith** has extensive theoretical and applied research experience in AI. He has worked closely with the Office for National Statistics for over 15 years, increasing their understanding of uncertainty/risk management, and developing AI-based tools they use for the SDC of published statistics.

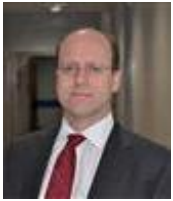

Professor Felix Ritchie: 5 Safes and Disclosure Control, the University of the West of England

**Felix Ritchie** is the author of the 'Five Safes' model and advises on data governance across the world. He is an internationally recognised expert on output SDC, having developed the original theory and first generic guide for research environments in 2006, and subsequently leading both theoretical and operational developments. He devised and delivers the national training programme in output SDC.

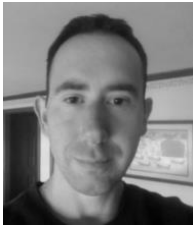

Dr Richard Preen:

**Richard Preen** received B.Sc. (Hons.) and M.Sc. degrees in computer science and a PhD degree in artificial intelligence from the University of the West of England, Bristol, U.K., in 2004, 2008, and 2011, respectively. He is currently a Research Fellow with the Department of Computer Science and Creative Technologies, the University of the West of England.

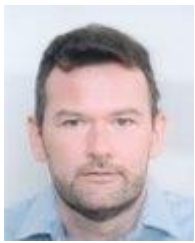

Andrew McCarthy:

**Andrew McCarthy** is a Research Fellow in Machine Learning with the Department of Computer Science and Creative Technologies at the University of the West of England, Bristol. He holds a B.Sc. (Hons) in Computing for Real-time Systems and an M.Sc. (Distinction) in Cyber Security. Currently, he is nearing completion of his PhD. His research interests include: Privacy and fairness of AI; Cyber Security; Secure machine learning; and methods for improving trust and reliability of AI models.

## Work Package 3 (WP3) - Legal and Ethics implications

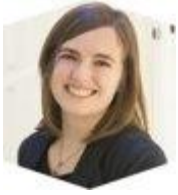

Professor Angela Daly: Regulation and governance of digital technologies, data protection, AI ethics

**Angela Daly** is an international expert in the regulation and governance of digital technologies, in particular data protection, AI ethics, intellectual property, and medical device regulation. She is a Professor of Law & Technology at the University of Dundee and the Chair of the Independent Expert Group to the Scottish Government on Unlocking the Value of Public Sector Data.

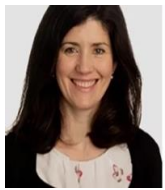

Maeve Malone: Lecturer in Intellectual Property law and Healthcare Law and Ethics, University of Dundee

**Maeve Malone** is the recipient of the Scottish Universities Law Institute (SULI) Early Career Academics grant for work on the legal and ethics framework of machine learning models.

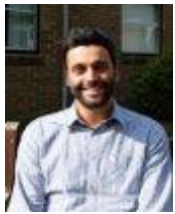

Dr Francesco Tava: Applied ethics, privacy and trust, the University of the West of England

**Francesco Tava** works at the intersection of applied ethics, political philosophy, and phenomenology. He is interested in issues concerning data and business ethics such as privacy, trust, and solidarity.

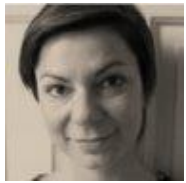

Dr Charalampia (Xaroula) Kerasidou:

**Xaroula Kerasidou** works as a post-doctoral Research Assistant at the School of Medicine at the University of Dundee. In her research, she explores the ethics, legal, social and political aspects of new technologies such as AI.

## Work Package 4 (WP4) - Patient and Public Involvement and Engagement (PPIE)

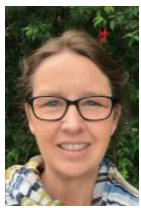

Jillian Beggs:

**Jillian Beggs** is an experienced public/patient advocate, driving the patient engagement work package. She is a lay co-applicant on GRAIMATTER and has led similar work packages for several other large research projects.

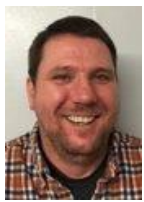

Antony Chuter:

**Antony Chuter** is an experienced public/patient advocate, driving the patient engagement work package. He is a lay co-applicant on GRAIMATTER and has led similar work packages for several other large research projects.

# 4 Stakeholder Engagement

The development of these recommendations has been an iterative process with input from many external stakeholders. We initially presented our recommendations to a workshop including attendees from many TREs across the UK and other DARE UK sprint projects. The first full draft version of this document was then shared with the workshop attendees, who were invited to provide written feedback. We updated the draft in response to this consultation, sharing the next draft for wider open consultation on Zenodo (https://zenodo.org/record/6896214#.Yt5i2HbMKHs) [1]. The document was widely publicised across the TRE and AI communities, utilising the DARE UK communication channels in addition to targeted communications via the networks of the authors. There were 469 unique downloads during the open consultation period. We further edited the recommendations based upon the open consultation feedback, resulting in this version at the completion of the GRAIMATTER sprint project.

# 5 Funding

This work was funded by UK Research and Innovation Grant Number MC_PC_21033 as part of Phase 1 of the DARE UK (Data and Analytics Research Environments UK) programme, delivered in partnership with HDR UK and ADRUK. The specific project was Guidelines and Resources for AI Model Access from TrusTEd Research environments (GRAIMATTER).

# 6 Lay Summary

Trusted Research Environments (TREs) provide a secure location for researchers to analyse sensitive data for projects in the public interest. TREs are widely and increasingly being used to support statistical analysis of personal data across a range of sectors (e.g., education, police, tax and health) as they enable collaborative research whilst protecting data confidentiality. TREs are often virtual environments where researchers can use their own laptop/desktop to access the remote secure TRE and the data relevant to their project. To ensure that an individual's personal data is protected, TREs apply a range of controls, such as:

- Projects are assessed to make sure they are in the public interest
- Only validated researchers who have undergone data governance training are allowed access to the data
- Only the data needed to answer the specific research question are made available to the researchers; only after being pseudonymised
- There is no access to the internet from within the TRE so researchers cannot inadvertently import or release any data without the appropriate checks
- Researchers must ensure that no one else can see their screens when they are working within the TRE and are not able to copy and paste from the environment
- Researchers must sign declaration forms covering their responsibilities when using the TRE and accessing data
- Normally, only aggregate, summary level data such as a trend or a graph can be removed from the TRE and used for purposes such as research publications. TRE staff perform manual and automated checks on outputs before release to prevent disclosure of individual's personal data. This process is called disclosure control.

TREs have historically only supported the analysis of data using classical statistical methods. These result in outputs using descriptors such as trends, averages, and counts, for which disclosure control methods are well understood.

There is an increasing demand to also facilitate the use of Machine Learning (ML) techniques for data analysis. Machine Learning is broadly defined as training a machine to perform complex tasks in a way that is similar to how humans solve problems. The result of using Machine Learning to analyse data is a (typically complex) piece of software called a trained model. The role – and benefit – of a trained model is to make a prediction when provided with a new example. ML models have been trained for many valuable applications e.g., spotting human errors, streamlining processes, helping with repetitive tasks and supporting clinical decision-making. To realise those benefits in practice, the models need to be trained on data held within TREs to represent the characteristics of the data. They then need to be released from TREs for use in the outside world. However, releasing trained ML models from TREs introduces an additional risk for the disclosure of personal data, including special category data under data protection laws, such as racial or ethnic origin, genetic, biometric, or health data. To meet legal requirements for data protection and ethics standards on fairness, accountability and transparency, particular care is needed for the safe release of such models. It is here that the size and complexity which give ML models their power present three significant challenges for the TRE's traditional disclosure-checking process:

1. Even models that are 'simple' in ML terms are usually too big for a person to view easily.
2. Our research shows that a person can't say whether a model is disclosive simply by eyeballing it.
3. Most significantly, ML models may be susceptible to external hacking using methods that reverse engineer the learning process to find out about the data used for training. These attacks can have greater potential to re-identify personal data than they would for conventional statistical outputs. This means that ML models trained on TRE data may be considered personal data(sets) and therefore fall under data protection laws.

The combination of, on the one hand, growing demand and potential benefits, and, on the other, significant challenges presented by using ML, creates a need to develop disclosure-checking solutions specifically targeted at ML models.

We evaluated a range of tools and methods to support TREs in assessing output from ML methods for personal data. We also investigated legal and ethics implications and controls. We have developed recommendations for evaluating risk, clearing models, and ensuring good practice when developing ML models. These recommendations have been developed with input from public representatives through a series of five workshops and input from two lay co-leads within the core team.

A summary of our recommendations for a general public audience can be found at DOI: 10.5281/zenodo.7089514

# 7 Scope

These recommendations provide disclosure controls for the safe release of trained machine learning models from TREs to protect personal data. These are different from the controls required for aggregate level results from classical statistical models. However, **the scope assumes that TREs already apply the high-level "Five Safes" controls** [2], **and implement more granular level controls** normally utilised for classical statistical studies such as those recommended in the Health Data Research (HDR) UK Principles [3] and Best Practices for Trusted Research Environments paper [4] and the Scottish Safe Haven Charter [5].

The 'Five Safes' is a popular way to structure thinking about data access solutions in the UK. Originally used mainly by UK statistical agencies and social science academics, in recent years it has been adopted more widely across the UK government, health organisations, and private sector bodies. The 'Five Safes' comprises:

**Safe People:** The researchers accessing the data through TRE are trained and authorised to use the data safely, follow guidelines, and report data safety concerns if any.

**Safe Projects:** TREs ensure that the research projects are approved by data owners, and that data are used appropriately and for public benefit.

**Safe Outputs:** TREs screen all outputs thoroughly and approve the release only after ensuring that it is non-disclosive of personal data.

**Safe Data:** The data are de-identified/pseudonymised before access is granted to researchers. It is ensured that researchers only see the data that they need to.

**Safe Setting:** TREs provide a safe environment to access personal data and prevent any unauthorised use.

**Many of our recommendations utilise and extend the controls already applied using the 'Five Safes' model**. For example, all research projects must have appropriate ethics approvals, but we have identified specific ways in which that process could usefully be amended. Rather than creating a new ethics approval process for assessing the implications of the additional disclosure risk posed by releasing trained ML models, the recommendations suggest that details of the additional disclosure risks be included within a standard ethics application. This will allow the risks to be balanced against the benefits of the activity along with the controls to mitigate the risks.

This report makes many recommendations, covering both technical and operational measures. At this stage, we are **not expecting that a TRE would adopt all the recommendations**. We have analysed the use of ML from a conceptual, worst-case perspective, and actual ML modelling in TREs. This will likely provide evidence for what works in practice. For example:

- if most modelling in TREs does not approach the worst-case scenarios analysed here, then unsophisticated checks, carried out by staff with lower training, may be more appropriate with only expertise brought in as needed; if, on the other hand, it appears that most ML models are high risk and bespoke (*sui generis*), then an equivalent level of expertise in protection may be necessary.
- we have proposed multiple solutions to give TREs the flexibility to tailor their output checking regime to their particular needs; it may be that operational tests will show that; e.g. responses A and B together or response C on its own provides adequate protection but that A, B and C is unnecessary.
- we have not considered in detail the implementation of restrictions by researchers but have used our experience as researchers and conceptions of what would be an acceptable output checking response; perhaps researchers will express a strong preference for response X over response Y because X is easier to incorporate in their code.
- ML models required to be 'safe' will have less predictive accuracy or usefulness than models which need not meet safety requirements. When researchers are aiming to release a 'safe' model they will typically wish to compare the 'safe' model with an unrestricted, potentially unsafe model within the TRE environment to assess attenuation in performance. The question of whether 'safe' models can maintain adequate levels of accuracy or usefulness as compared to potentially unsafe models is highly application-specific, so we defer the question to researchers rather than trying to answer it in general.

This report focuses on ML models exported from TREs. With a mandated point of release, the TRE's output checking routine, and compulsory output checking can be enforced, as well as other checks (e.g., for compatibility with the original ethics approval). However, the protection measures identified here are **also usable in non-TRE situations**.

As ML modelling generates the same confidentiality risks for a given dataset, whether inside or outside a TRE, good research practice would suggest these checks are applied by non-TRE modellers as well.

## 7.1 Out of scope:

The following are beyond the scope of our recommendations:

i.  The ethics of ML model training and use of the trained model i.e., the legal and ethics implications of using the trained model such as developing and using a medical device in a clinical setting. Such topics are assessed via the standard ethics approval process, whether or not a TRE is involved.
ii.  Medical device regulation, which may come into effect after a trained ML model has been released from the TRE.
iii.  Artificial Intelligence (AI) techniques other than Machine Learning (ML). ML is just one type of Artificial Intelligence (AI). Here we consider ML only.
iv.  The ML models are being trained entirely on data within the TRE as opposed to being adapted from models initially trained on external data (transfer learning using pre-trained models), that is we do not consider disclosure by differencing against external models.

Assumptions:

a)  It is assumed that TREs run a service and do not have intellectual property (IP) over or intellectual input into ML development, which is done by researchers.
b)  The data used for model training within the TRE are pseudonymised following good practice pseudonymisation methodologies.
c)  The existing best practices for doing research in TREs are followed, as per Health Data Research (HDR) UK Principles and Best Practices for Trusted Research Environments paper [4].
d)  ML models can be considered separate and apart from personal data, but can also be considered personal data [6].

## 7.2 Audience

The audience for these recommendations includes all of the different groups responsible for implementing these recommendations, i.e. researchers, TREs, Data Controllers, Data Governance Committees and Ethics Committees. These groups need to understand the recommendations, with specific groups responsible for applying the recommendation. For example, Data Governance Committees need to understand the controls which are used by TREs to assess a project application which describes these controls, researchers need to understand the controls to describe them in their project application, and a TRE needs to specifically apply the controls during the project. For each detailed recommendation, we have indicated the groups which need to understand the recommendation or are responsible for applying the recommendation.

# 8  Glossary

## TREs AND ACTORS

**TRE:** a trusted research environment sometimes called a 'data enclave', 'research data centre' or 'safe haven' is an analytical environment where the researchers working on the data have substantial freedom to work with detailed row-level data (see below) but are prevented from importing or releasing data without permission, and typically are subject to a monitoring and a significant degree of access control.

**Researchers:** someone who has permission to use a TRE and has access to data within that TRE. For this work, researchers are assumed to be interested in building machine learning (ML) models that they will then wish to remove from the TRE. We use the term "researchers" to mean researchers who could be academic or from a commercial or government setting – all of whom could be training ML models to develop a solution in the public interest.

**Data controller:** a person(s) or corporate body, who alone or jointly with others determine the purposes for which and the manner in which any personal data are, or are to be processed (or who are the controller(s) by virtue of the Data Protection Act 2018, section 32(2)(b) (by means by which it is required by an enactment to be processed)), and all of whom are officially registered on the Information Commissioners Office Data Protection Public Register [7]. Data controllers need to approve the use of their data for research projects. In this document, we are using the term Data Governance Committee to represent the group/person by which Data Controller(s) approve applications to use their data for the research project. TREs can be a service which is run by the same organisation that is the Data Controller or can be a service which is run by a separate organisation to the Data Controller. We have separated the roles of the Data Controller and TRE within these recommendations.

**Joint Data Controller:** a person(s) and/or body corporate, who jointly determine the purposes for which and how any personal data are or are to be processed (or who are the controller(s) under the Data Protection Act 2018, section 32(2)(b) (by means by which it is required by an enactment to be processed)) and all of whom are officially registered on the Information Commissioners Office Data Protection Public Register [7].

**Attacker or adversary:** a person or group of persons who attempts to extract, from the trained ML model, some or all of the personal data that was used to train it.

**Personal data:** 'any information relating to an identified or identifiable living individual' as per section 3 of the UK Data Protection Act 2018 (which implements the EU's General Data Protection Regulation or GDPR). Personal data is broadly interpreted, especially via the concept of 'identifiable' which is further defined as 'a living individual who can be identified, directly or indirectly in particular by reference to (a) an identifier such as a name, an identification number, location data or an online identifier, or (b) one or more factors specific to the physical, physiological, genetic, mental, economic, cultural or social identity of the individual.

**Special categories of personal data:** the term used in current UK and EU data protection legislation for certain more sensitive categories of personal data. The special categories are: race; ethnic origin; political opinions; religious or philosophical beliefs; trade union membership; genetic data; biometric data (where this is used for identification purposes); health data; data about sex life; and sexual orientation. There are additional legal requirements which must be met to process special category personal data.

**Synthetic data**: data artificially generated to replicate statistical properties of a given real-world dataset, ideally not containing genuine identifiable information i.e. no personal data.

**Row-level data:** a data set whose columns are different types of measurements/images/features, and where each row contains the record of an individual or organisation. It is synonymous with '**record-level data**' or 'microdata', and in contrast to aggregate-level data.

**Aggregate-level data**: summary data which is acquired by combining individual-level data and may be collected from multiple sources and/or on multiple measures, variables, or individuals. It is synonymous with '**aggregate(d) data**' or '**statistical results**'.

## MACHINE LEARNING

**Algorithm:** A set of instructions to execute a task. This is a very general definition; algorithms may be deterministic (always giving the same answer when presented with the same input) or stochastic (giving different answers with various probabilities). Algorithms are not necessarily run by a computer; humans also use algorithms implicitly when making decisions. We will usually use the term to mean a set of instructions which only refer to data generically, rather than a specific dataset. An example of an algorithm is the ordinary least squares (OLS) method for fitting linear models.

**Machine Learning (ML) model:** Some computer code which implements an algorithm that, when presented with some input data, processes it in some way, and produces some output. There are many possible ML models, differing by the particular type of process they implement, and how they implement that process. For example, a particular ML model might process images (the input data) to assign them into a category (the output), which might be useful when attempting to build an ML system for diagnosing disease from a medical image.

Formally, a model is a set of candidate distributions over the domain of a given dataset (we leave differentiation between classical statistical models and ML models unspecified at present: there is no general distinction, and whether a model constitutes an ML model is best determined on a case-by-case basis)[8].

**Trained Machine Learning (ML) model:** ML models are not usable until they have been trained. Training involves presenting the model with data that is relevant to the task at hand and modifying any parameters within the model to optimise its performance in the task of interest. For example, an ML model that is to be used for diagnosing breast tumours from mammograms will be trained with mammograms (input) with known tumour status (output). The training process will modify the parameters within the ML model such that the number of mistakes it makes on this "training" data is minimised. Once trained, the ML model can be used to generate an output for inputs that were not part of the training data: for example, to predict the tumour status for a new mammogram or predictive text on a smartphone.

Formally, a trained ML model is one of the candidate distributions of an ML model.

**Predictions:** The usable output of an ML model when given some data. Typically, this is the estimated chance of something happening given a set of inputs, where the estimation is made by the model. In the example above, the prediction would be the chance that the mammogram shows a real tumour.

Formally, a prediction is a (summary of a) conditional distribution derived from a trained ML model.

**Features:** independent variables, often organised in columns in a given dataset used to train the model; e.g. age, sex, medical history, heart attack incidence.

**Target model:** an ML model (untrained, trained or being trained) that is the target of an attack.

**Instance-based models:** Models which, to be able to make predictions, must 'remember' one or more training data samples exactly, rather than just summary data. These provide an immediate security risk, since specifying the model entails specifying individual samples. Such models are sometimes able to be made private by transforming training data samples randomly and only remembering the transformed samples [9].

**Ensemble methods:** the use of multiple methods, usually with their outputs combined through some form of the voting process, done to improve overall performance. This may involve the use of the same method on different parts of the dataset (e.g., random forests [10] gradient boosting methods (e.g. XGBoost [11])) or different methods applied to the same dataset (e.g., super-learners [12]).

**Kernel-based methods:** Group of model types that are used for pattern analysis. They use similarities between observations to build the model rather than the observations themselves. They are almost always instance-based methods, meaning that at least some of the training data must be saved within the trained model.

## TRAINED ML BEHAVIOURS

**Machine Learning (ML) model architecture:** The ML workflow specifies the various layers processes involved in the machine learning cycle: data acquisition, data processing, model engineering, execution and deployment. Broad categories of architecture Machine Learning are supervised learning, unsupervised learning, and reinforcement learning. Within each category, the architecture specifies the learning algorithm (e.g. neural networks, random forests, etc.) and its internal structure (e.g., number and type of layers in a neural network). A trained ML model is saved to a computer-readable file. Such a file could be loaded and used to make predictions or loaded to inspect the properties of the model.

**Hyper-parameters:** High-level parameters that can control aspects such as the model architecture (number of layers in a neural network, maximum depth of a decision tree, etc) and the learning process through which one particular trained model is chosen from all the possibilities

**Generalisation:** The ability of a machine learning model to make predictions on data that it did not see during training.

**Overfitting:** Situation in which a model fits and remembers the training data too well and does not generalise well for unseen data. Overfitting can facilitate membership attacks. Typically, small or unrepresentative training datasets can lead to overfitted models, especially if the data points have many features. A bad choice of hyper-parameters can also lead to overfitting (for example, excessively big neural network for simple classification tasks). Detection of subtle overfitting is difficult and a fundamental area of ML theory. More egregious overfitting can be readily identified by non-experts.

Methods to reduce overfitting include: increasing the training dataset size, possibly using data augmentation techniques; using 'regularisation' techniques during training, which penalise candidate models for complexity; and optimising the choice of hyper-parameters, possibly with cross-validation. In neural networks, it is often beneficial to include dropout layers, which randomly deactivate neurons during training (effectively making the training procedure noisier). Differentially private optimizers (such as DP-SGD) add noise during the optimization steps/training process and may lead to better generalization.

**Data augmentation techniques:** generate training samples from existing samples. In the case of images, a typical technique is to resize and rotate images in the original training set to generate new samples.

**Federated learning:** a technique that allows a machine learning algorithm to be trained on data that is stored in a variety of servers, devices, or TREs. The trained algorithm parameters (not data) are pooled into a central device which aggregates all individual contributions into a new composite algorithm.

## ATTACK TYPES

**White box:** a type of model attack where the attacker knows some information about the training data, the target model classifier, architecture and learned parameters of the target model. For example, this might include knowing the weights of a neural network, or the decision thresholds in a rule or tree-based model.

**Black box:** a type of model attack where the attacker has only query access to the model. That is, they can present input data to the model and observe the predictive outputs that the model makes. For example, in a model that detects the presence/absence of a tumour in an x-ray image, the attacker can present an image to the model and will receive the probabilities that a tumour is present or not. Black box attacks do not have access to the interior of the model.

## RISK TYPES

We have identified 3 major risk types (RT):
RT1: non-malicious researchers training models inside the TRE, naive to possible threats faced by those models and saving inappropriate information within the disclosed model file.
RT2: malicious researchers deliberately hiding data inside disclosed files.
RT3: an external attacker with access to disclosed model files after release.

## DISCLOSURE RISKS

**Membership Inference:** the risk that an attacker (of either a White or a Black box) can create systems that identify whether a given data point was part of the data used to train the released model. This risk is far more likely to be disclosive of special category personal data in cases of medical data (X was part of a trial for a new cancer drug) than it is for other forms of data TREs might hold (Y was part of a survey on educational outcomes).

**Membership Inference Attacks (MIA)**: a type of attack where an adversary wants to predict whether row data, which belongs to a single individual, was included in the training data set of the target model.

**Attribute Inference**: the risk that an attacker, given partial information about a person, can retrieve values for missing attributes in a way that gives them more information than they could derive just from descriptions of the overall distribution of values in the dataset.

**Attribute Inference Attacks (AIA):** a type of attack where the adversary is capable of discovering a few characteristics of the training data.

**Individual Disclosure:** occurs when outputs from an analysis segment a participant with a specific condition, e.g. rare genetic disease, or a unique combination of conditions that might put the data of this individual at high risk of being identified or disclosed.

**Group (class) Disclosure:** occurs where information about a group has been uncovered, and an individual can be identified as a member of that group; for example, the model might show that all males reporting for treatment aged 45-55 show traces of cocaine use.

**Disclosure by differencing:** occurs when two separate outputs from a TRE can be used to infer private information by *comparing* them to each other, even if neither output allows such inference on its own. For example, given a fixed set of patients, if we fit one model to predict heart attack risks and another to predict lung cancer risk and release both, then we may be able to learn about the patients by comparing predictions on both models, even if we could not learn anything private from looking at only one of the models. We do not explicitly consider disclosure by differencing in GRAIMatter.

**Mechanism:** a term describing a procedure which takes a dataset and outputs some information about it. As an example, given data D=(X1,X2,X3), a mechanism M might return a 'noisy mean' M(D)=(X1+X2+X3)/3+λ, where λ is a random variable.

**Differential Privacy**: a measure assigned to an output mechanism roughly stating how similar outputs can be when their training data differs by only a single sample (e.g., in the example above, looking at how similar (probabilistically) the values M(D1,D2) and M(D1,D2,D3) are to each other). Limiting differential privacy ensures that a malicious attacker with access to all-but-one of the training samples would have difficulty inferring the values of the last training sample. For a full definition and treatment see [13], particularly chapters 2 and 3. Differential privacy can be quantified as:

- **(Epsilon) ε-differential privacy**: the probability/density of seeing some output of the model never changes by more than a factor of **ε** when changing one sample.
- **(Epsilon,Delta) (ε,δ)-differential privacy**: the probability/density of seeing some output of the model usually does not change by more than a factor of ε when changing one sample, except for a set of values which have probability δ (delta) of being observed. The lower ε and δ, the more private the mechanism. Typically, lower ε and δ correspond to less useful mechanisms, in that they contain less useful information about the true dataset.

It should be noted that differential privacy is intended to bound the risk of disclosure of individual records and does not address the issue of group disclosure well. Although it can be applied additively (i.e., the epsilon value for a group of n people is n * the value for one person), this is of little use to a TRE/researcher planning to agree a value for the risk appetite (epsilon) other than in the unlikely case that the size of every potentially sensitive sub-group is known in advance.

**Data Governance Application:** An application form sent to a Data Governance Committee which explains how the data will be used for the research project. Details of the risks and controls to personal data need to be articulated within the form along with the benefits of the project.

**Data Governance Committee:** The group/individual who assesses the data governance application and approve/disapprove. Such committees include the Data Controllers responsible for approving the use of their data for the project. In some instances, this could just be a single responsible individual Data Controller who approves the application e.g., Caldicott guardian in the case of health data. In other instances, the committee may include representatives such as lay members of the public.

**Data Governance Approval Process:** The process by which researchers fill in a data governance application and send this to a Data Governance Committee for approval/disapproval.

**Ethics Application:** An application form sent to an Ethics Board which explains how the data will be used for the research project. Details of the risks and controls to personal data need to be articulated within the form along with the benefits of the project (e.g., individual, commercial, societal etc.).

**Ethics Committee:** The group/individual who assesses the ethics application for approval/rejection. Such committees include local research Ethics Committees (LRECs) based in specific universities and research centres as well as multicentre research Ethics Committees (MRECs), which were introduced in the UK following the publication of Department of Health guidance HSG (97)23 to deal with multicentre research.

**Ethics Approval Process:** The process by which researchers fill in an ethics application and send this to the ethics board for consideration.

**Contractual Agreement**: A binding legal document, signed, sealed and delivered by contracting parties.

**Linked Agreement**: This is an existing or contemporaneously agreed contractual agreement between contracting parties. A linked agreement can be incorporated into the terms of a contractual agreement. This might be necessary if the identity of the initial Data Controller or processor has changed, or if there was to be a change to the permitted purpose of the trained ML Model, as outlined in the original Data Governance and Ethics approval. An example of a permitted purpose assigned to a trained ML Model might be: functions designed to protect members of the public. An example of a linked agreement would be the initial (or updated) Data Governance and Ethics Approval Process for the development of ML model processing personal data. Another example of a linked agreement is a data sharing agreement.

**Data Protection Impact Assessment (DPIA):** is a process designed to identify risks arising out of the processing of personal data and to minimise these risks as far and as early as possible.

**Data Sharing Agreement:** Data sharing agreements set out the purpose of the data sharing, cover what happens to the data at each stage, set standards and help all the parties involved in sharing to be clear about their roles and responsibilities. Having a data sharing agreement in place helps demonstrate accountability obligations under the UK GDPR.

**End User License Agreement (EULA):** This is the license which a user of a trained ML model may have to sign to use the model. It will provide clauses which explicitly prohibit the hacking of the model to disclose personal data or to use it for another purpose other than the "permitted" purpose. The EULA will be drafted, finalised and put in place by a qualified and insured legal team. The license will usually be signed, sealed and delivered by two separate parties. These parties can be called the licensor and the licensee. All terms and conditions should follow the fair, reasonable and non-discriminatory (FRAND) [14] principles.

## OTHER

**Safe Wrapper:** code that unobtrusively augments the functionality of existing software for machine learning. Typically, when a safe wrapper is applied, the model will retain the 'look and feel' of its original version whilst adding functionality to:

- Automate the running of various attacks to assess the vulnerability of a trained model.
- Assist researchers in meeting their responsibilities, by warning when their choices for hyper-parameters or components are likely to result in models that are vulnerable to attack - and make suggestions for alternative choices.
- Detect when researchers have either maliciously or inadvertently changed important parts of a model (or hyper-parameters) between training and requesting release.
- Produce reports for TRE output checking staff summarising the above, to assist them in making good decisions about whether to release trained models.

**GitHub:** An open online platform that lets people work collaboratively on projects/software codes from anywhere while tracking and managing changes to software code.

**Encryption:** A process which protects personal information by scrambling the readable text into incomprehensible text which can only be unscrambled and read by someone who has access to a specific decryption key.

**Restrictive software:** Software that is subject to conditions or limitations imposed by a technology licensor or supplier and requires consent before its disclosure or assignment to third parties.

**Public release:** This means making the content of a work public through publication, presentation, broadcast or other means.

**Public representatives:** People who can represent the public interest and protect the integrity interests of the individual.

**Release** (when referring to a model): To export a trained machine learning model outside the TRE for deployment and for making predictions. Another term for this is "egress".

**Grant of a License or Transfer** (when referring to a model): To release a model to specific researchers, by way of the grant of a license. A license is legally and contractually binding.

**Deploy** (when referring to a model)**:** To set up the trained model within an environment where it can be efficiently used to make predictions.

**License:** a type of contractual agreement to authorise the granting of a license, to enable the use or release of a machine learning model. It can be perpetual or non-perpetual.

**Reproducibility:** means achieving a high degree of reliability or similar results when the study/experiment/ statistical analysis of a dataset is replicated.

**Model Disclosure Control (MDC):** Controls on privacy achieved exclusively through controlling aspects of the trained ML model, under the assumption that unlimited prediction queries may be made using the model by an attacker.

**Model Query Control (MQC):** Controls on privacy achieved by restricting access to or use of the trained ML model after release.

**End User:** A person who uses the ML model outside the TRE by obtaining access to the model either through access to a resource storing the code (e.g. a version control system such as GitHub), software that implements the model, or a web service that allows the model to be queried. This person is often a different person from the researchers who trained the model.

# 9  Executive Summary

## 9.1  Background

Trusted Research Environments (TREs), also sometimes referred to as safe havens, data enclaves or research data centres, have become widely used to support observational research on sensitive pseudonymised linked data within a secure virtual environment. There is a range of controls implemented by TREs, adhering to legal requirements (especially under data protection legislation) and the 'Five Safes' model [2]. These controls were expanded upon by the Scottish Safe Havens [5] the Turing Institute [15] and Health Data Research UK [4]. TREs assure Data Controllers that their data can be securely shared for research purposes without risking individual or patient confidentiality, resulting in a more scalable, streamlined process for population-level statistical studies using health and administrative data.

With the emergence of AI (a subset of which is Machine Learning, or ML), researchers are now pushing the capabilities of TREs to support the development of trained ML models. Applications of such models include clinical decision support, streamlining public sector processes and spotting overlooked issues in routine healthcare provision or other sectors. To support ML development, TREs need to provide additional tools and computing resources. They also need to ensure that the trained ML models to be released from the environment do not contain any record level, or personal data i.e. disclosure control needs to be applied to the trained ML model. The disclosure control guidelines developed for 'traditional' analyses are inappropriate for ML models [17]. ML model disclosure control is also needed to support Federated Learning across TREs.

[Appendix E](#) provides some real-world examples covering 4 different scenarios. The examples are written for a non-technical audience, with links to code [16] evidencing these examples for technical readers:

- o  **Finding out unknown personal data about a famous person:** we show that if a model has been trained using data from a famous person - for example, a Member of Parliament (MP) who attends a hospital in an area which provided the training data - and a hacker already knows some information about the famous person from data in the public domain, they may be able to query the trained ML model to find out the other data relating to the famous person. In our example, the hacker knew the MP is diabetic, asthmatic, smokes, and is 62 years old. The hacker could query the model many times in a systematic, automated way and find out that it is highly likely that the MP was also in an overweight BMI category and had slightly high blood pressure. This is an example of disclosure via attribute inference.

- o  **Identifying if someone famous has suffered from cancer:** Often, some details about the health of famous people are well known, either they admitted it themselves or someone leaked it to the media: for example, smoker, asthmatic, etc. If an attacker realises that the training data from a ML model to predict the outcome of cancer treatment means only people with cancer were included in the study, then the attacker could use the publicly available health status to prove a famous person had cancer. This is an example of disclosure via membership inference.

- o  **Successful candidates in a job interview:** In this case, we demonstrate that sometimes the outcome of the model can have easily identifiable patterns depending on whether a person was included or not in the training data, and it could have unexpected consequences if misused. We illustrate how researchers want to help drug users when they are at high risk of insolvency. An employment agency misused the model to determine if a candidate was part of this research, and therefore is a drug user (regardless of economic problems or not) and did not select the candidate for interview.

o **Hospital admission survival:** This example demonstrates how instance-based models contain data of some patients that were part of the training data and they can be retrieved from the model. In this case, the ML model was to predict the chances of survival for patients admitted to the hospital. Luckily, the TRE output checkers spotted the problem and did not allow this model to be released publicly, so the researchers had to decide on an alternative approach.

In earlier work, we interviewed 14 UK and 6 international TREs [17] to discover how TREs check ML models for personal data before approving release. It was found that universally TREs did not have mature processes, tools or an understanding of disclosure control for ML models. The existing processes were manual and did not consider various risks. A recent work discussed a wide analysis of the risks associated with ML vulnerability at prediction time [18]. We also reviewed different types of ML models for the potential to encode individual-level data or personal data and assessed security threats and vulnerabilities [19], [20]. Three major Risk Types (RT) were found:

- RT1: non-malicious researchers training models inside the TRE, naive to possible threats faced by those models and saving inappropriate information within the disclosed model file
- RT2: malicious researchers deliberately hiding data inside disclosed files
- RT3: an external attacker with access to disclosed model files after release.

Attacks carried out by such actors can include discovering whether someone's data was in the training set (Membership Inference), recovery of personal data given a partially complete record (Attribute Inference) or uncovering `whole records' likely to be in the training set (Model Inversion). This previous work proposed some very high-level mitigation strategies including restricting types of models used, avoiding overfitting, model/code inspections pre-disclosure, testing the model to be released on a subset of data withheld from the researchers, assessment of attack risks, and use of synthetic data and models.

The UK data protection regulator, the ICO, provides a helpful overview of the different types of risks and methods to mitigate these risks [21].

DARE UK (Data and Analytics Research Environments UK) has funded the GRAIMATTER (Guidelines and Resources for Artificial Intelligence Model Access from TrusTEd Research environments) sprint project to develop a set of guidelines and recommendations for how TREs should carry out disclosure control on ML models. DARE UK is a programme funded by UK Research and Innovation (UKRI) to design and deliver a more coordinated national data research infrastructure for the UK. **This recommendations green paper is one of the main outputs of GRAIMATTER.**

We assessed technical, legal and ethics, training, costing and PPIE (patient and public involvement and engagement) risks and controls. Section 9.3 provides a summary of the recommendations. A summary of the investigative work carried out within each category (Technical / Ethics and Legal Aspects (ELA) / Costing / PPIE / Training), is provided along with detailed recommendations in Sections 10 to 14. Recommendations for areas for future development and research are provided in Section 15.

High-level recommendations are numbered "R (n)". Detailed technical recommendations are numbered "TECH (n)", Legal and Ethics recommendations are number "ELA (n)", costing recommendations are numbered "C (n)", PPIE recommendations are numbered "P (n)" and training recommendations are numbered "TR (n)".

We have provided a range of Appendices which provide additional details of the investigative work (Appendix A), the attack simulation tool kit (Appendix B), an example data dictionary template (Appendix C), example constraints (Appendix D), case studies (Appendix E), lay explanations of metrics (Appendix F), and steps for running attack simulations (Appendix G).

## 9.2   Summary of investigative work

The GRAIMATTER project investigated several areas:

o **Quantitative assessment of the risk of disclosure from different ML models:** We trained models across a range of parameter/hyper-parameter values and data types and assessed the performance of disclosive membership inference attacks (MIA) in each case. We formalised and implemented different types of attribute inference attacks (AIA) and repeated the large-scale testing for this risk. This enabled us to identify factors for assessing risks in disclosure of these model types, and parameter and hyper-parameter regimes to avoid. In Predicting Vulnerability Risks (Section 16.4) we also present promising results from the use of Machine Learning algorithms trained on our experimental data, to predict the likely vulnerability of a trained model in advance. This could offer both a means for saving time and money (by not producing vulnerable models) or for advising researchers about the key drivers when managing the accuracy-vulnerability trade-off. Finally, we considered the practical relevance of these risks.

o **Controls and Evaluation of Tools**: We evaluated a range of tools to determine their usefulness in the semi-automating assessment of disclosure risk. Our evaluation considered the current 'effectiveness', requisite level of support/maintenance, and the risk of these tools themselves becoming part of an 'arms-race'. We considered approaches, where a model fitted to synthetic data [22], [23] is released instead of the true model, which partly shifts [24] privacy considerations to the synthesis process. We developed Python 'wrappers', around commonly used modelling functions (scikit-learn/Tensorflow), automatically assessing disclosure risk and producing reports to assist the output-checking team.

o **Legal and ethics implications:** We investigated the legal and ethics issues accompanying ML model release from TREs. We identified and assessed how current UK legislation applies to TREs supporting trained ML model release and the extent to which the legislation addresses ethics issues pertinent to the release of ML models out of TREs, and what happens after that release, developed based on personal data, such as transparency, privacy, data protection and non-discrimination. We looked at the main legal obligations incumbent on researchers and TREs, including protecting the confidentiality and security of the processing of the personal data held by the TRE and used for training ML models. We took into account the duty of Data Controllers to protect confidentiality but also share data in the public interest. We drew from the field of AI ethics and governance from an international level (UNESCO Draft Recommendation on the Ethics of Artificial Intelligence [28]) and EU level (proposed Artificial Intelligence Act) to inform our analysis, especially in considering the reform of applicable UK frameworks.

o **Public Engagement:** We ran 5 PPIE workshops – chaired by our lay co-applicants and included 8 members of the public selected with consideration for diversity including, but not limited to, gender, ethnicity, age and geography. The workshop structure included explaining the challenge with ML and AI, outlining the legal challenges, presenting project outputs and agreeing on the next steps. All workshops were followed up with a one-to-one call to check understanding and offer additional support if needed. The feedback we received fed into these recommendations.

Although risks of disclosure from trained ML models are widely recognised and there is significant research in this area both by academic and commercial groups, during our literature search we did not find any examples of a data breach from trained ML models. There were many examples of data breaches of sensitive data including special category personal data from other sources. This is because ML training and use are in relative infancy – particularly with health data. It is appropriate that recommendations such as these are implemented before such breaches occur.

## 9.3 Summary recommendations

To summarise the recommendations, we have grouped them into the stage of the project where they occur, and we have listed the relevant detailed recommendations which apply to each high-level recommendation. As explained within the Scope these recommendations utilise and extend the controls already applied using the 'Five Safes' model such as those recommended in the Health Data Research (HDR) UK Principles [3], Best Practices for Trusted Research Environments paper [4], and the Scottish Safe Haven Charter [5]. For example, the data provided to researchers is pseudonymised, there are checks on the individuals accessing them to ensure they are *bone fide* researchers, and the TRE is configured to limit open access to the internet.

'Researchers' could be either academic researcher(s) or researcher(s) from industry or government.

### 9.3.1 Pre-Project

Figure 1 shows the processes where recommendations apply pre-project.

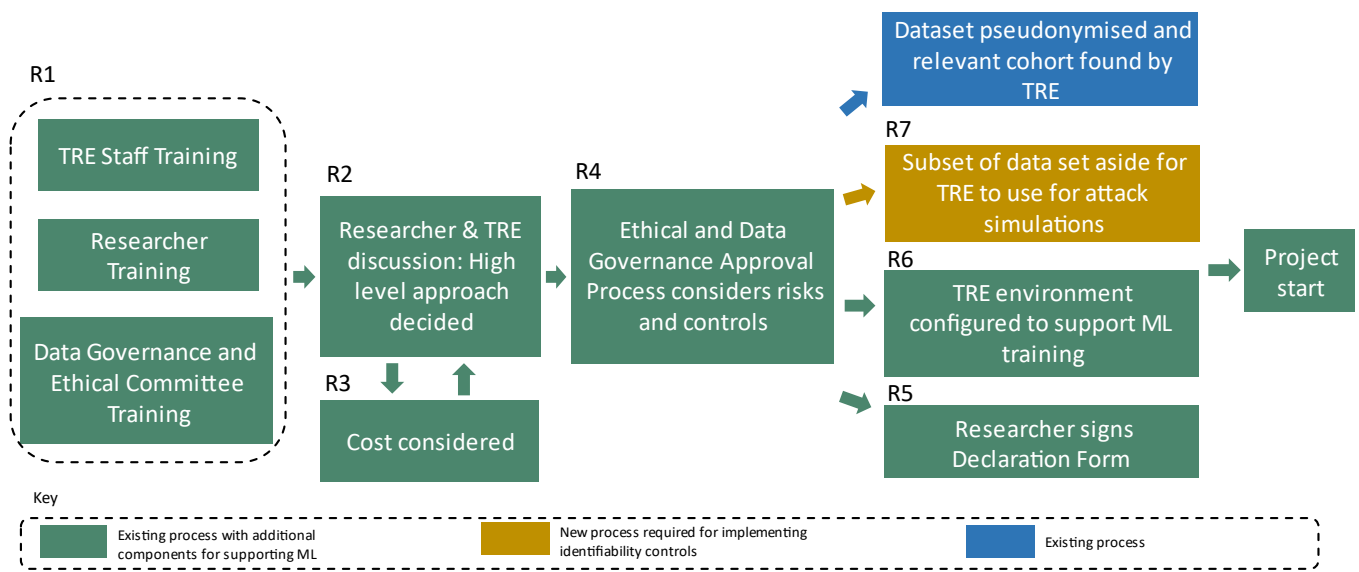

*Figure 1 - Pre-project process where the recommendations apply. Green boxes show existing process which take place for "traditional" TRE projects and where new recommendations apply to support ML training. The blue box shows a process which already takes place and where modifications to support ML training are not required. The mustard coloured box shows a new process which is required to support MDC for ML models. The "R" labels correspond to the high level recommendations within the main body of text.*

##### 9.3.1.1   R 1: All groups involved in the process should be appropriately trained

Before the commencement of ML training projects, all groups involved in the process (TREs, Data Governance Committees, Ethics Committees, and researchers) should undergo specific training relevant to their group to ensure they are aware of the risks and controls which could be applied re disclosure control of personal data, the legal and ethics implications, and their responsibilities within the process (TR 1, TR 2, TR 3, TR 4, TECH 5). There should be a group of TRE-trained experts on ML disclosure control who can work across TREs, reducing the burden on each TRE (TECH 5, C 5).

**Relevant Detailed Recommendations:** TR 1, TR 2, TR 3, TR 4, TECH 5, C 5

##### 9.3.1.2   R 2: Researchers should discuss their plans with TREs and decide upon a high-level approach which applies appropriate disclosure controls and mitigates risks

Once the TREs and researchers are trained in the risks and controls which could be applied, they should have an informed discussion regarding the relative merits and challenges of different approaches for the specific research project, and they should agree on an approach and high-level project plan (TECH 1). The TRE can support researchers to decide the approach to be taken.

We have identified two high-level groups of controls: Model Query Controls (MQCs) and Model Disclosure Controls (MDCs) (Figure 2). Both sets of controls, if applied robustly, result in the release of only anonymous data to the end-user. Such data does not contain personal data and therefore data protection legislation does not apply. The difference between the 2 methods is that for MDCs the exported trained ML model is considered anonymous itself.

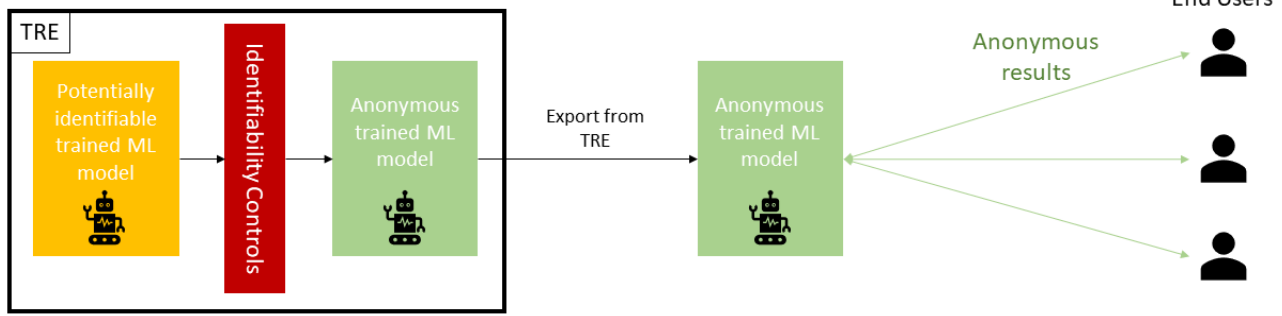

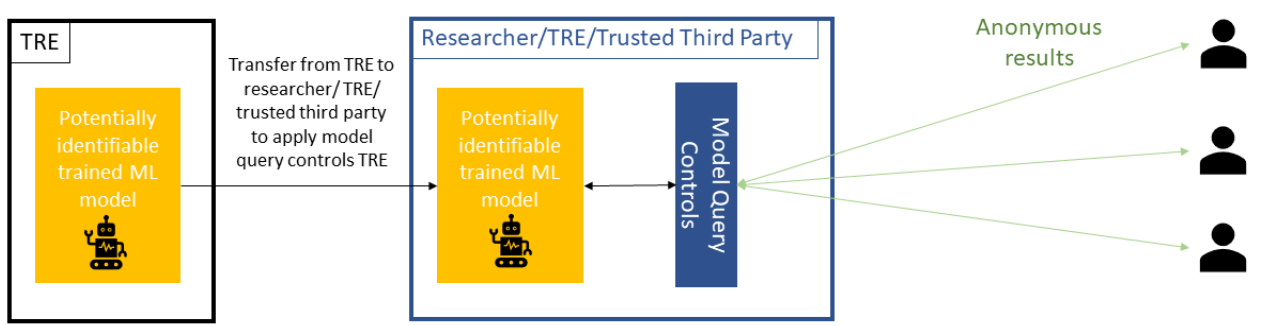

*Figure 2 - Overview of Model Query Controls (MQCs) and Model Disclosure Controls (MDCs). Both types of controls result in end-users only having access to anonymous data. For MDCs the model is checked for identifiable data, only being released from the TRE if the model passes the checks. For MQC the trained model may potentially contain personal data but access is managed in such a way that the end-users are only given access to anonymous data. It may be for some projects a mixture of both types of controls are utilised.*

Whereas for MQCs, the controls are placed around the trained model, restricting end-user access to only anonymous results. MQCs can be applied to either a potentially identifiable trained ML model or an anonymous trained model i.e. some projects may wish to apply both types of controls.

We are not recommending any one of these approaches over another; each has advantages and disadvantages. Both MQCs and MDCs, if applied correctly, will protect personal data. The decision will be a combination of researcher preference and whether the Data Controller is supportive of the approach.

The approach for managing disclosure control risk can be categorised into 4 high-level options, MQCs and 3 types of MDCs:

- **MQCs:** Rather than the model being released openly, researchers can choose to limit the queries on the trained model (TECH 17), e.g. the model could be hosted within a secure web service which technically limits who can query the model and the number of times. Such a web service could be hosted by the TRE, a trusted third party or the research group. Another method of limiting queries is by embedding the trained ML model within the software which technically applies controls. For example, if a trained ML model was incorporated into a medical device (classified as a Class IIa medical device (Chapter III Rule 11 EU Medical Devices Regulation) installed within hospitals, the software could constrain the number of requests and who had permission to query the software. However, it is recognised that such software-based technical controls are relatively easily hacked in comparison to a secure web service.

  Model query controls limit the model to only black-box attacks, i.e. the model is not exposed to white-box attacks.

  Legal constraints should also be utilised, e.g., an End-User License Agreement (EULA) should explicitly prohibit hacking of the trained model to determine potentially identifiable information and only allow the use of the model for the permitted purpose (ELA 6). If the model potentially contains personal data then legal agreements may be required between the Data Controller and the researcher if the model is to be hosted outside of the TRE such as data transfer/sharing agreements and Data Protection Impact Assessments (DPIAs) (ELA 1, ELA 3). It may be that the MQCs sufficiently mitigate the risks such that the MDCs (see below) are not required (more likely for a web service solution), e.g. TREs may not be required to run attack simulations or for researchers to use methods to protect personal data such as differentially private methods. However, it may be that for highly sensitive personal data some MDCs will also be required.

- **MDCs:** These are controls which reduce the risk that the ML model could be hacked to discover personal data (TECH 5). For each of the 3 high-level options of MDCs, TREs should run attack simulations on the trained model (mimicking possible hacking behaviour) to assess for vulnerabilities. The model should not be allowed to be exported if it is considered to be too vulnerable (based upon the thresholds agreed with the Data Controller during the data governance approval process -ELA 8). To run such simulations the TRE should be provided with basic information from the researchers regarding how the model was trained (TECH 8, TECH 7.2). TREs should agree to confidentiality agreements, should the researchers have concerns regarding their IP (ELA 10). TREs should also carry out other basic checks such as ensuring that the model is significantly smaller than the data used to train the model (to ensure that the full training dataset is not included within the model) (TECH 7.3) and eyeball the code used to generate the trained model (to ensure that the researchers are not maliciously trying to hide data) (TECH 7.4).

o **MDCs – use of methods such as differential privacy or training on aggregate level data**: Researchers can choose to use methods which are designed to protect personal data such as differentially private methods or training on aggregate level data ([TECH 11](#)). These significantly reduce the probability that personal data will be encoded within the model and increase the chance that the model will pass the attack simulations run by the TRE. However, such methods have been shown to reduce the accuracy of trained models; hence, researchers may choose to adopt other controls instead. Methods such as differential privacy or training on aggregate level data are indeed highly recommended for instance-based models (such as KNN, SVM and Gaussian Processes) or models where the high numbers of parameters risk memorisation (such as deep learning). For instance-based models, this is because the non-differentially private versions of these methods need all or some of the original data rows to be able to make predictions and thus personal data is highly likely to be encoded within the trained model. When DP methods or similarly randomised methods are used, randomisation should be dependent on a random seed, and the value of this seed corresponding to the released model should be chosen (randomly) by the TRE staff, rather than researchers. It should be noted that differential privacy is intended to bind the risk of disclosure of individual records. While it can provide similar protection against group disclosure, it may over-protect, in that protection against group disclosure must be attained through much more stringent control over individual disclosure (formally, a mechanism which is (ε,0) differentially private against individual disclosure is (Nε,0)-differentially private against group disclosure for groups of size N). It provides no useful protection against group disclosure for arbitrarily large groups.

o **MDCs – use of safe wrappers:** To increase efficiency over native training (below), researchers may wish to employ community-developed Safe Wrappers [25] ([TECH 9](#)). These Safe Wrappers provide the same functionality as the standard ML libraries from which they inherit but apply "safe" limits on the hyper-parameters of the model to reduce the risk the model encodes personal data. They also produce risk reports and can automate the running of targeted inference attacks, to assist researchers and TRE staff in making good decisions about what models should be proposed and approved for egress. Using Safe Wrappers requires minimal changes to researchers' workflows while increasing the chances that the trained model will pass the attack simulations, reducing the effort of both the researchers and the TRE staff to run multiple iterations of training and running attack simulations prior to allowing export. Researchers will have increased confidence that they are not being disclosive and have avoided over-fitting. The disclosure control process for TREs should be streamlined and less costly.

o **MDCs – native training:** Researchers can utilise a method of their choice to train the model. However, this may not pass the attack simulations run by TRE staff, and the model may need to be retrained multiple times before it is considered "safe" for release. Such iterations are likely to be time-consuming and expensive.

**Relevant Detailed Recommendations:** [TECH 1](#), [TECH 4](#), [ELA 8](#), [ELA 10](#)

### 9.3.1.3   R 3: Researchers and TREs should consider the costs of implementing the controls

Most TREs work on a cost recovery basis. Researchers should ask TREs to estimate and charge for any additional work to undertake the MDCs of trained ML models such as running attack simulations ([C 1](#)) and the additional costs of maintaining the data and development pipeline to support legal requirements ([C 4](#)), e.g. for a certified medical device. TREs should consider outsourcing some of the highly technical work and obtaining estimates from these other sources if they do not have the skills in-house (C 5) and to pass on the costs to the researchers.

The costs for limiting queries on the model should be considered (C 3), e.g. the costs of running a secure web service or a software wrapper.

### 9.3.1.4    R 4: Approvals processes should consider the risks and controls for disclosure control of trained ML models

Once the approach has been decided upon, researchers should include details of the approach within their Ethics and Data Governance Applications, providing sufficient detail to support the review process. Researchers should use standard text available from TREs which describe the controls which will be applied (ELA 9). This reduces the burden on researchers to draft such information and for TREs to review unfamiliar text as well as reducing the number of re-submissions of applications due to missing information.

Data protection legislation is generally considered not to apply to the anonymous [26] aggregate level releases from classical statistical data analysis from a TRE, as appropriate, widely applied, controls ensure that these releases do not contain personal data. In contrast for projects which train ML models, without the additional controls proposed by these recommendations, Data Controllers, TREs and researchers should consider that Data Protection legislation may apply to the trained ML model (ELA 1) as it may contain personal data.

Both the Data Governance and Ethics applications should consider:

- the risks and controls associated with the training and release of ML models (ELA 2.1)
- if the controls to be implemented will result in end-users of the model only having access to anonymous data (ELA 1)
- the trustworthiness of the organisations involved (ELA 2.1)
- if legal contracts are proposed to cover the responsibilities of each party (ELA 1, ELA 3)
- the length of time the model will be used prior to requesting new approvals (ELA 2.2)
- the requirement to keep the data and the pipeline available to meet legal requirements (ELA 2.3)
- the correct language to describe the process (ELA 2.4)
- if MQCs are required after release from the TRE (ELA 3)
- the requirement that details of a released model are recorded on a data use register (ELA 4, ELA 7)
- whether legal clauses will be added to the terms of use of any resulting trained ML model (ELA 6)
- whether researchers will sign additional clauses within researchers' declaration forms (ELA 7)
- whether confidentiality agreements are required to protect researchers' IP (ELA 10)
- whether agreements are required between the Data Controller and TREs (whether acting as the controller or data processor) and that TRE processes for applying MDCs have been detailed within the application and approved by the Data Controller (ELA 11).

It may be that Data Governance and Ethics Committees have the expertise to review such application after their training (TR 4) or they may wish to share the application with an external expert to independently review.

Researchers need to articulate the wider benefits of their work when they wish to release an ML model and explain how they will secure the models (ELA 2). This is especially important when ML models may be released as full consideration of benefits may add weight to the value of the research versus additional risks from disclosure. Approval processes may wish to consider a risk-based approach vis-a-vis personal data (ELA 8).

Public representatives should be involved in Data Governance and Ethics Approvals Processes (P 1). It may be that the Data Governance Committee or Ethics Committee requests that additional controls are placed on the process to mitigate the risks, e.g. the model cannot be released openly but has to be hosted within a secure web service.

Figure 3 shows a decision flow chart, summarising the key questions that Data Governance and Ethics Committees should consider when reviewing an ML project application

**Relevant Detailed Recommendations:** ELA 1, ELA 6.

### *9.3.1.5   R 5: Researchers and the Principal Investigator should sign modified Researchers Declaration Forms*

Researchers must read and sign researcher declaration forms to analyse data within most TREs, in addition to the Principal Investigator. Such forms explain the researchers' responsibilities and behaviours to which they must adhere. We recommend additional clauses are added to researchers' declaration forms to cover the training and release of an ML model (ELA 7), including:

- o   Researchers must abide by the controls which have been approved by Data Governance and Ethics Committees.
- o   Researchers agree that details of the project will be recorded on the data use register and that researchers are required to update the TRE with any changes to the use of the trained ML model once it leaves the environment i.e. if it is incorporated into a product with a CE mark and is being sold commercially.

Researchers should sign these modified researchers' declaration forms if they are running an ML project.

**Relevant Detailed Recommendations:** ELA 7

### *9.3.1.6   R 6: TREs should be enhanced to support ML training*

To support ML training the TRE environment should be configured with the software required to train ML models.

**Relevant Detailed Recommendations:** TECH 2

### *9.3.1.7   R 7: Data should be set aside for applying MDCs*

If the TRE is to run attack simulations as an MDC, some data should be kept to one side and not provided to the researchers, for use in attack simulations (TECH 6).

**Relevant Detailed Recommendations:** TECH 6

### 9.3.2   Project running within the TRE

There are 2 areas within the TRE:

- **TRE researcher's ML training zone:** This is the zone where researchers will carry out the training of their ML model once the project has been granted the appropriate permissions.
- **TRE export zone:** This is the zone where TRE staff check files for personal data prior to enabling them to be exported from the TRE if they are considered "safe" based upon any limits within ethics and data governance approvals.

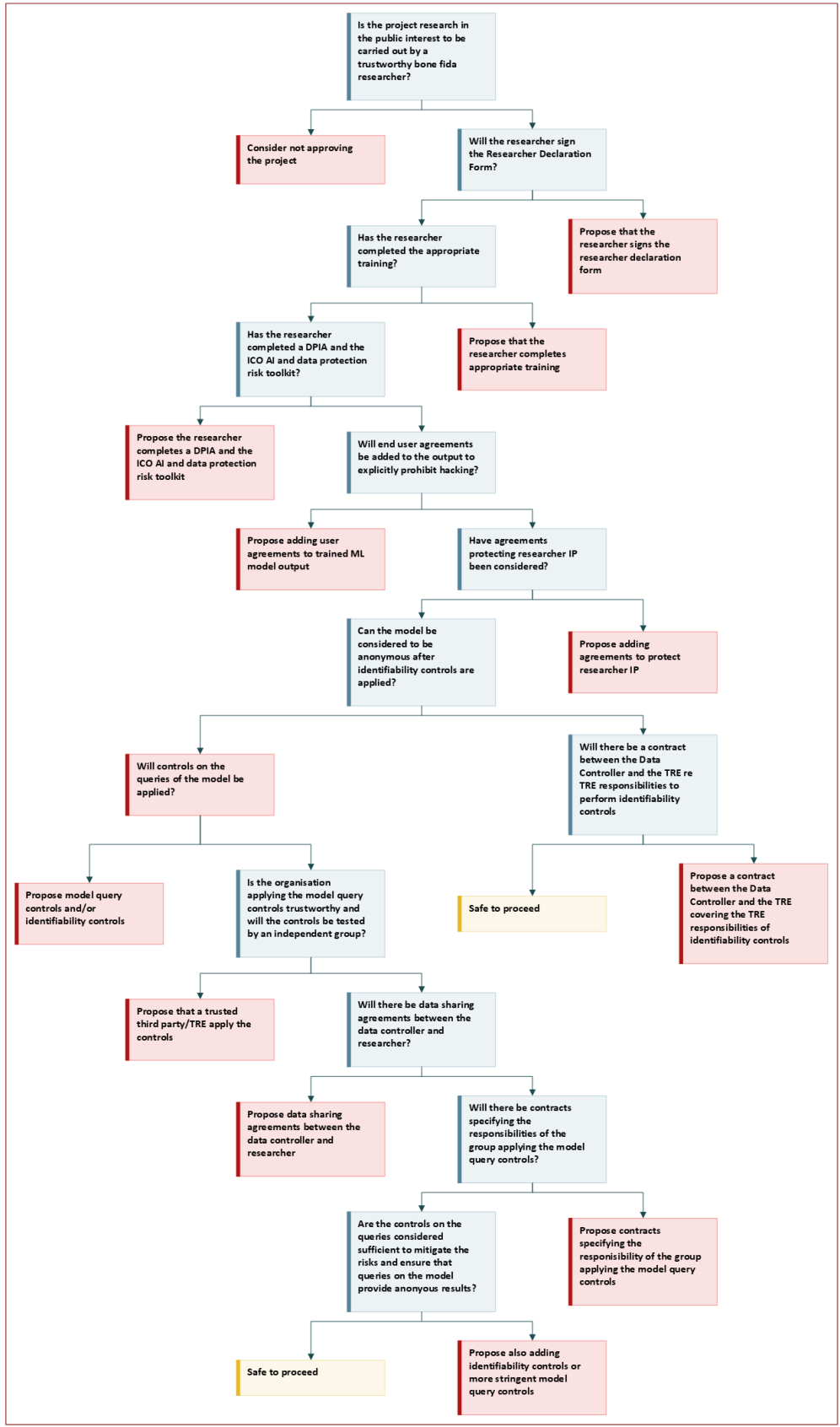

*Figure 3 - a decision flow chart, summarising the key questions that data governance and Ethics Committees should consider when reviewing an ML project application*

Figure 4 shows these 2 zones and the processes within them.

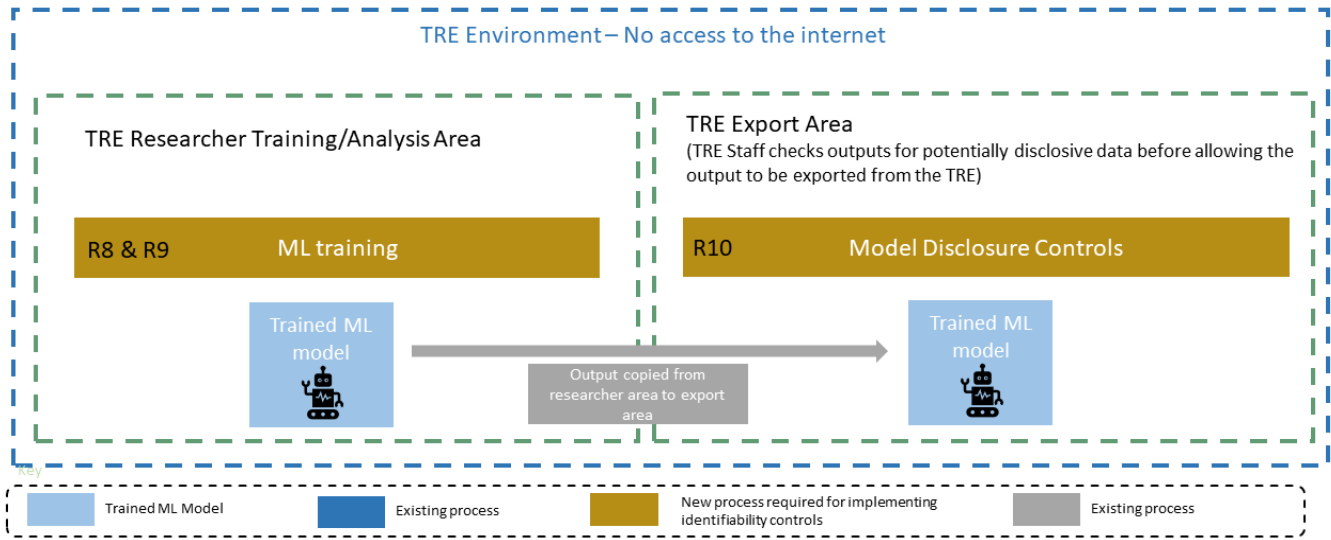

*Figure 4- TRE project process where the recommendations apply. The mustard-coloured boxes show a new process which is required to support MDCs for ML models. The grey box shows an existing process. The light blue box shows the trained ML model. The "R" labels correspond to the high-level recommendations within the main body of text.*

### 9.3.2.1 R 8: Within the TRE researcher's ML training zone, researchers should use differentially private methods or train on aggregated data for instance-based models or models where the high numbers of parameters risk memorisation (if MDCs are required)

If MDCs are required, instance-based models (such as KNN, SVM and Gaussian Processes) or models where the high numbers of parameters risk memorisation (such as deep learning) should have additional controls such as training on aggregated data, using differentially private methods, or training on synthetic data generated by a differentially private mechanism (TECH 11).

**Relevant Detailed Recommendations:** TECH 9

### 9.3.2.2 R 9: Within the TRE researcher's ML training zone researchers may wish to use safe wrappers (if MDCs are required)

If MDCs are required, to increase efficiency, researchers may wish to employ Safe Wrappers [23] (TECH 9) which have been developed by the community based upon a set of principles (TECH 10).

**Relevant Detailed Recommendations:** TECH 9, TECH 10

### 9.3.2.3 R 10: Within the TRE export zone, TREs should apply a range of MDCs (if MDCs are required)

For projects where MDCs are a requirement based on the Data Governance and Ethics approvals for the project, the TRE should apply a range of MDCs prior to the model being "approved" for release.

Researchers should provide a 'data dictionary' describing the inputs to their model in a standardised format (TECH 7.1, TECH 8). To assess the risk of a model containing identifiable data, TREs should run a range of checks on the model to be released including:

- utilising a risk assessment check-list (TECH 7.1)
- running the model to be released against set-aside data to ensure that it provides the expected result (TECH 7.2)
- checking the size of the trained model (TECH 7.3)
- eyeballing the code used to train the model (TECH 7.4)
- checking the file type of the model to be released to check that it is contained within a list of accepted file types for release (TECH 7.5)
- running attack simulations using the set-aside data (TECH 7.6):
  - passing the training data through the trained model to obtain predictive probabilities.
  - passing the data held out by the TREs through the trained model to obtain predictive probabilities.
  - running membership and attribute inference attacks on the trained model. This will result in a series of metrics describing the membership and attribute inference risk, that would be interpreted by the TRE staff.

- not relying solely on average performance metrics (e.g. accuracy, AUC, etc) when evaluating simulated attacks (TECH 7.7)

It is recommended that TREs do not assume that a safe ensemble implies safe base models or vice versa (TECH 12). TREs should take care to ensure that both the overall model is safe, and the base models that constitute it are safe. When using Federated Learning the final models should be tested for vulnerability (TECH 14).

Synthetic data could still be potentially disclosive (TECH 15). If the generated synthetic data encodes relationships between the variables accurately enough to use for ML model training, then the relationships encoded could also provide enough information to be vulnerable to hacking to find personal data. The software/pipeline that was used to generate the final trained model which has been approved for release should be stored as a snapshot so that it can be reproduced (TECH 3, ELA 2.3).

**Relevant Detailed Recommendations:** TECH 3, TECH 7, TECH 8, TECH 12

### 9.3.3 Model release

Figure 5 below shows the processes where recommendations apply after model release.

#### 9.3.3.1 R 11: When a model is exported from a TRE an entry should be made within a data use register

Data Use Registers should be extended to include information on trained ML models and details around their export and controls (ELA 4). The use of data for training ML models and the controls on the models should be visible to the public through searching data use registers (P 2).

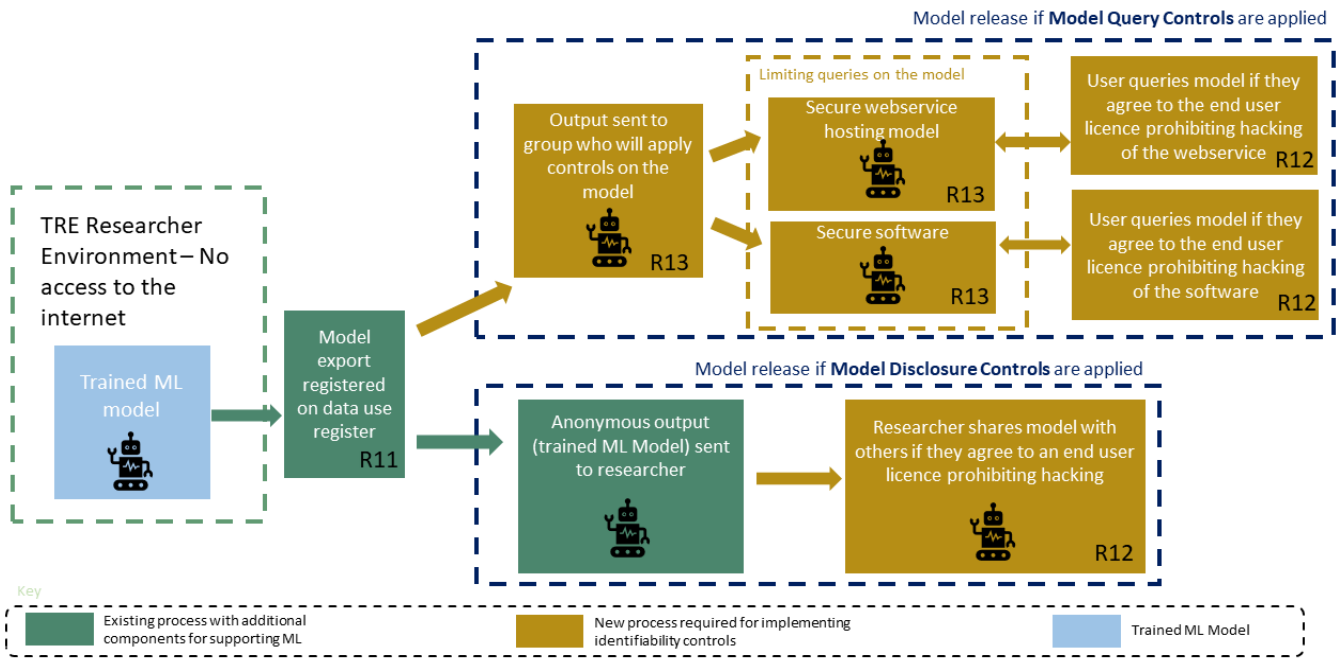

*Figure 5. Model release process where the recommendations apply. Green boxes show the existing process which takes place for "traditional" TRE projects and where new recommendations apply to support ML training. The mustard-coloured box shows a new process which is required to support MDCs for ML models. The light blue box shows the trained ML model. The "R" labels correspond to the high-level recommendations within the main body of the text.*

### 9.3.3.2    R 12: When a model is exported from a TRE clauses should be added to the End User License Agreement (EULA)

Clauses should be added to the EULA of any resulting ML model (ELA 6): for example, an EULA for web service users or an EULA for users of the software with ML models embedded. EULA should also be required if the model is to be released as open-source software. A legal term outlining the "permitted purpose" of the trained ML model should be included within the EULA.

### 9.3.3.3    R 13: Model query controls should be utilised if MDCs are not used/sufficient

If MDCs are not being applied or are not considered to be sufficient to mitigate the risks, MQCs should be considered once it is released from a TRE (TECH 16). For example, the model could be hosted within a secure web service or embedded within the software which technically limits who can query the model and the number of times. Such controls to limit the number of queries on the model should be tested by an external party (TECH 17). The costs for limiting queries on the model should be considered (C 3).

Legal contracts should be considered to cover the responsibilities of each party if MQCs are required after release from the TRE (ELA 3), such as data sharing agreements.

**Relevant Detailed Recommendations:** TECH 8, TECH 16, TECH 17, C 3, ELA 3

# 10 Technical

## 10.1 Technical Background

This section provides a high-level description of the field and the investigations undertaken by GRAIMATTER to inform the recommendations. More details regarding the investigations and experiments are provided in Appendix A.

Any model, whether hand-crafted by human researchers or created automatically by a ML algorithm, attempts to describe the underlying data in ways that facilitate the drawing of useful inferences on unseen data. In doing so, they risk disclosing information about individuals in the training set. These risks are well understood for tables of data – for example, no cell should just reflect the data from fewer than k people as that might allow an individual or his or her attributes to be disclosed (so-called k-anonymity) - and TRE staff are trained in spotting and rejecting disclosive outputs. However, ML models are typically far more complex, and rarely human-readable. For example, it is possible (albeit laborious) to manually check that no 'leaf node' of a single decision tree only refers to one person's data. However, this may be impractical or impossible when the model takes the form of a forest of such trees, each reflecting a partial view of the data, so it is the intersection of leaves from different trees that leads to disclosure. In some cases, it may be even worse, as trained models such as K-Nearest Neighbours and Support Vector Machines will typically directly encode row-level training data in the model output.

Thus, one aspect of the work undertaken has been to examine how the hyper-parameters that control the learning behaviour of different ML algorithms can result in models that pose different types of risk to an individual's privacy.

Since many trained ML models are represented as large arrays of numbers, another risk factor is that malicious researchers might attempt to manually edit models to 'hide' some of the training data within them – analogous to how text can be 'hidden' within an image (steganography).

Thus, the second strand of research has been concerned with exploring the role that different automated tools could play in helping:

- researchers (who may not be privacy experts) abide by best practices,
- TRE output staff make a rapid informed decision about the likely level of risks posed by different trained models for which release has been requested.

### 10.1.1 Quantitative assessment of the risk of disclosure from different AI models

Decisions on ML model disclosure from a TRE concern trade-offs between usability and privacy risk, particularly re-identification (RT3), necessitating prior assessment of risks [22]–[24], [27]. Factors such as ML model types, learning algorithms, data types and researchers' expertise can (sometimes provably) influence the risk of disclosure. We performed experiments to assess the risks of different models, for different data types, within different training regimes (Appendix A). We identified the following experimental factors as the best representatives of the risks and threats to ML models within TREs:

- Data types: different types of datasets including medical images, Electronic Health Records, and administrative data.
- Algorithm/training types: we have identified a range of representative model types spanning classical and AI settings: SVM, Decision Trees, Regression, K-means, MLP, CNN, and KNN. We have incorporated different training algorithms, including those based on Differential Privacy. We have simultaneously implemented methods for the generation of synthetic data [22]–[24], [27]

- Attack types: we have used the most common attack types conferring privacy risk: membership inference attacks (MIA), attribute inference attacks (AIA) and model inversion [28].

We trained models across a range of parameter/hyper-parameter values and data types and assessed the performance of disclosive MIA in each case (Appendix A). We have formalised and implemented different types of AIA and repeated the process of large-scale testing for this risk. This enabled us to identify factors for assessing risks in disclosure of these model types, and parameter and hyper-parameter regimes to avoid. Finally, we considered the abstract practical relevance of these risks, and how that should influence how we quantify them. Context matters: we know from 'traditional' Statistical Disclosure Control that risks which are meaningful in some environments are irrelevant in others. Evidence also matters: we need to understand whether the information burden placed upon an attacker is likely to be feasible (that is, we are not interested in attacks which are theoretically possible but meaningless in practice), or has historical precedents.

### 10.1.2 Controls and Evaluation of Tools

We evaluated a range of tools to determine their usefulness in the semi-automating assessment of disclosure risk (Appendix B). Our evaluation considered the current 'effectiveness', requisite level of support/maintenance, and the risk of these tools themselves becoming part of an 'arms-race'. We considered approaches, where a model fitted to synthetic data [22], [23] is released instead of the true model, which partly shifts [24] privacy considerations to the synthesis process.

Open-source privacy attack and defence tools we evaluated include:

- TensorFlow Privacy https://github.com/tensorflow/privacy
- Adversarial Robustness Toolbox https://github.com/Trusted-AI/adversarial-robustness-toolbox
- Fawkes https://github.com/Shawn-Shan/fawkes
- Diffprivlib https://github.com/IBM/differential-privacy-library
- IBM AI Privacy Toolkit https://github.com/IBM/ai-privacy-toolkit
- ML-PePR https://github.com/hallojs/ml-pepr
- ML Privacy Meter https://github.com/privacytrustlab/ml_privacy_meter
- ML-Doctor https://github.com/liuyugeng/ML-Doctor
- PrivacyRaven https://github.com/trailofbits/PrivacyRaven
- CypherCat https://github.com/Lab41/cyphercat
- AttriGuard https://github.com/jjy1994/AttriGuard
- MemGuard https://github.com/jjy1994/MemGuard

Factors considered in our assessment of the tools:

1. Project license
2. Project documentation and tutorials
3. Project is based on peer-reviewed publication(s)
4. Project version is numbered and has a version tagged as a release
5. Project recency and update frequency
6. Project popularity (e.g., GitHub stars and contributors)

7. Project quality assurance (e.g., use of unit tests, static analysis, and continuous integration tools)
8. Project distribution: location and ease of installation with current platforms/software
9. Project ease of use: including output analysis metrics and/or visualisations
10. Number of and quality of other projects that use/depend on the project

We developed python 'wrappers', around commonly used modelling functions (scikit-learn/Tensorflow), automatically assessing disclosure risk and producing reports to assist the output-checking team.

From a researcher's perspective, these behave just like the familiar library code, only adding functions to:

- Suggest changes if the researchers initialise the model with hyper-parameters or algorithm variants likely to lead to disclosure risk.
- *Request_release()* of a trained model: which saves a copy to file and produces an automated report for the TRE output checkers.

These 'wrapper' functions are controlled by a human-readable 'config' file that embeds the TRE and dataset-specific risk appetite. For example, they can be configured to restrict parameter choices to 'safe' regimes, preventing unintentional researcher-introduced risk (RT1).

In investigating the possibilities for functionality invoked when a researcher calls *request_release()*, we have implemented and evaluated mechanisms for detecting whether changes had been made to trained models *after* the ML algorithms had run. The latter risk might occur through misunderstanding or unintentionally (changing a parameter but forgetting to 'retrain')(RT1). However, it might also result from malicious attempts to hide data or subvert 'safe' choices for training parameters (RT2). The *request_release()* functionality can optionally be extended to run automated membership and attribute inference attacks on the specific model that the researcher wishes to export.

We investigated but discarded as impractical the possibility of hiding potentially 'vulnerable' parts of models (e.g., 'support vectors' of a support vector machine, or layers of a neural network) from either malicious researchers (RT2) or external 'white-box' attack post-release (RT3).

### 10.1.3  Differential privacy

We investigated the use of differentially private (DP) methods as a way to reduce the risk that a trained ML model contains identifiable data. Within our context, differential privacy is a mathematical way of measuring [13] the susceptibility of a trained ML model to an adversarial attack. Strictly, differential privacy considers a 'mechanism' which is an algorithm to convert private data to a public release – e.g. ML model training. Such a mechanism may be differentially private to a given degree. In this sense, DP is a property of the *algorithm* used to generate a trained

ML model, rather than the private data used or the specific TRE release; we may use (for example) a 'Differentially private linear model' mechanism to fit a linear model to some private data, to give us coefficients which we then release, but it is the *mechanism* which is DP rather than the data or coefficients (Figure 6).

DP measures the worst-case susceptibility of a trained ML model to an attacker with the aim that non-worst-case settings (for instance, where an attacker has imperfect or no knowledge of any private data) are automatically covered.

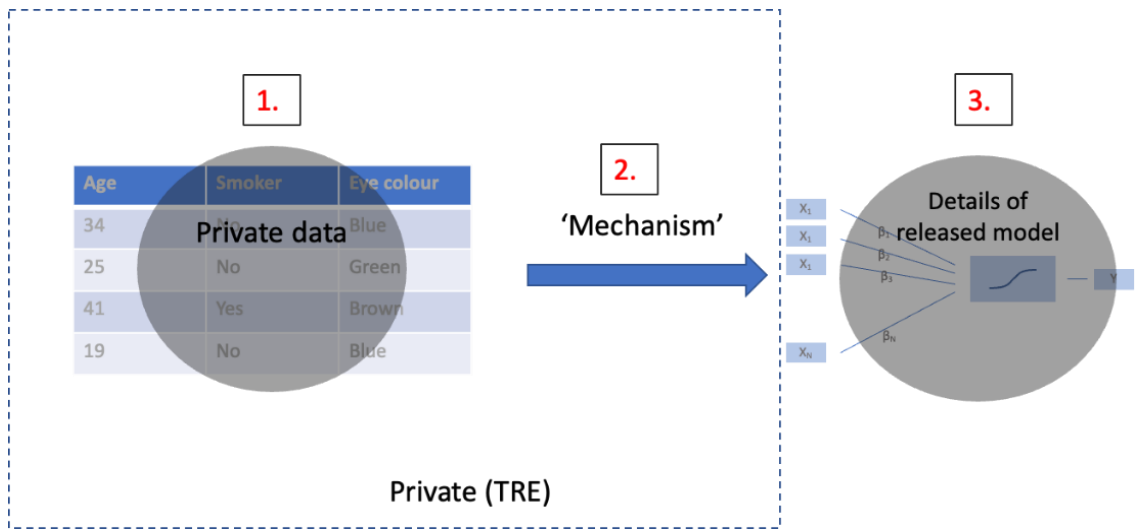

*Figure 6. Abstract description of the process of preparing a model for release. Differential privacy is a property of the 'mechanism' (object 2), whereas most output checks concern only the release itself (object 3).*

Suppose that we have some data for a range of samples within a TRE and we use some of this data to generate a public release of the trained ML model, using some mechanism. A hypothetical attacker has access to all of the data we have inside the TRE, except for one small thing: there are two samples, A and B, and only one of them was used in generating our release. The attacker knows everything about samples A and B except for which one we eventually used. They also know everything about the mechanism we are using to go from the TRE data to the release, and they can see the data we release. The attacker is interested in working out which one of A and B we used to generate our release. DP asks the question: to what extent can they work this out from looking at what we released publicly?

To see why this helps: let us suppose the attacker currently knows nothing about how we trained a model. Then the task of finding out whether A or B is in the dataset is essentially impossible, as there are so many other unknowns affecting the public release. If, on the other hand, the adversary knows 99% of what we did to train the model, the last 1% (whether A or B is in the dataset) is much easier. DP limits the ability of the attacker to find that last 1%. This also makes the already-difficult task of finding the first 99% much more difficult, thereby making the attacker's task much more difficult in general.

One important thing to notice is that if our mechanism is deterministic (that is, given the same input data, always gives the same output) then the attacker can work out which of A or B was included immediately – they would just run the mechanism with A included, then with B included, and see which of those matched the actual release (this also implies that releases like the mean or standard deviation of a dataset are **not** differentially private). Thus, differential privacy mechanisms introduce some randomness into the release.

A mechanism is differentially private if we can guarantee that the likelihood of seeing any particular release if we used A differs from the likelihood of seeing the same release if we used B by at most a particular multiplicative factor ε (epsilon, usually given on a logarithmic scale) and an additive factor δ (delta, usually given as a difference from 1). These factors specify the 'level' of differential privacy. The lower epsilon is, and the closer delta is to 0, the more private the mechanism is.

Generally, DP algorithms are adjustable, so we can set the level of (ε,δ). There is usually a fundamental trade-off in that the more private a model is, the less useful it is; that is, there is a necessary trade-off between privacy and accuracy,[13], [22], [27].

It should be noted that as interest in Differentially Private Machine Learning increases, a range of technical problems is arising, and variants of DP are being proposed. DP is a useful metric of risk but should be interpreted exactly as per its definition. Furthermore, the extent of noise that needs to be added to guarantee acceptable privacy controls via differential privacy typically leads to an unacceptable compromise in accuracy; an acceptable compromise generally needs to be decided between TRE staff and researchers on a case-by-case basis, which is potentially unrealistic to do at the research planning stage [29]. In fact, for federated learning, it is shown in [30] that anti-overfitting techniques can offer a better accuracy-privacy trade-off than DP.

Differential privacy is not directly comparable to statistical disclosure control practices typically used in non-ML TRE releases; for instance, the common statistical disclosure control practice of avoiding summary statistics calculated on fewer than 5 samples does not confer ε-differential privacy for any finite ε.

DP can be used to protect against group disclosure but it can be overly conservative for this purpose and cannot be used to protect against disclosure of arbitrarily large groups. Please see the glossary for discussion of DP in the context of group disclosure.

### 10.1.4  Synthetic data

We considered the evaluation of synthetic data generators (SDGs) in two settings:

- SDGs may be used inside a TRE in a setting for which the aim is to release a predictive model. In this case, the SDG would be used to generate synthetic data inside a TRE and a model for release trained to this synthetic data instead of the original data. Note that in this scenario, the performance of the final released model is heavily dependent on the quality of the generated synthetic data.
- The SDG itself could be released, either in addition to or instead of a predictive model, as an output.

In case 1, the SDG is used either instead of or in addition to a privatised predictive model and can be assessed for disclosure risk by simply assessing the predictive model that is released.

Case 2 is more difficult as an SDG is a model of the overall data distribution, which a predictive model is not. In other words, the maximal amount of disclosive information potentially contained in an SDG is higher than that in a predictive model. Released SDGs can be provably private (e.g. differentially private)[27], [31], [32].

Since synthetic data generators typically estimate a sampling distribution which resembles the distribution of the original data, they are not necessarily invulnerable to simply returning original data samples: indeed, if the estimated sampling distribution is sufficiently overfitted, this is exactly what they will do. For this reason, synthetic data generators should be treated with equivalent scrutiny to supervised machine learning models.

### 10.1.5  Instance-based models

We investigated row or instance-based models and found that particular care is required for such models. For example, the K-nearest neighbours (KNN) algorithm, when given a new data sample, begins by finding the K-closest training samples and thereby requires all training samples to work. Similarly, the commonly used Support Vector

Classifier (SVC) requires a subset of the training samples (known as the support vectors) to operate. SVC is an example of a kernel-based method (which requires a comparison of a new data sample with training samples) and all are dangerous in this way (other examples include Support Vector Regression and Gaussian Processes).

Such models are generally not appropriate for release from TREs, as the row-based data is automatically disclosive and can be essentially read off. Some model-specific adaptations are possible; for instance, KNN problems can be converted to 'radius-neighbours' classification [33], in which points within a fixed 'closeness' are counted, and SVC can use random transformations to hide the original data points [9]. Both methods compromise the accuracy of the predictive model. These examples are provably differentially private.

### 10.1.6 Imaging data

Imaging data creates particular issues when it comes to disclosure. In many cases, it will not be the image *per se* that holds any personal data (although some MRIs and chest x-rays might identify individuals and therefore constitute personal data), but some artefacts within the image. For example, we will rarely care about pixel values in the background of an image. This leads to a range of interesting questions about how aspects such as membership inference should be defined and whether or not anonymisation is useful and possible with such unstructured data [34]. Promising reconstruction attacks of training datasets of images based on model updates have been proposed in [35], [36]. Imaging data is out of scope of these recommendations due to the sprint project nature of GRAIMATTER. We plan to investigate this type of data as part of future research.

### 10.1.7 Limiting Queries on the Model

We identified two broad categories of the use of the model which necessitate different recommendations across each of the different control areas: unlimited queries on the model (for example, when the model is distributed to other end-users) and limited queries on the model (for example, when end-users can upload data to the model and run it, but the model runs on the servers owned by the TRE).

Typical assessments of disclosure risk [37], [38] operate through assessing the *behaviour* of a machine learning model, rather than directly *observing* it. More formally, when we assess disclosure risk, we characterise a machine learning model as a function from data to outputs: two distinct models which implement the same function can have equal susceptibility to attack when considered in this way. To be able to fully 'see' the function; all we need to do is to be able to repeatedly evaluate it on a range of inputs.

In this sense, an important dichotomisation of TRE-released models concerns whether their associated functions can be queried arbitrarily by parties external to the TRE, or whether such queries are restricted. This is distinct from the question of whether a released model is white-box (internal operations visible) or black-box (internal operations hidden), although the question of whether queries are restricted need only be considered for black-box models, as the associated function in white-box models is directly observable. Restricted-query models may be implemented in several ways: for instance, hosted on secure servers with the curation of queries to the model and restrictions around who may submit queries, or bundled with software which restricts queries and encrypted so that the model can only be accessed through the restrictive software. Restriction of queries may take the form of limiting the range of inputs for which the function can be evaluated (for instance, limiting a medical model to only use for data linked to real patients, rather than simulated data), or limiting the number of queries able to be made in a given time period.

An important appeal of restricting queries is the potential to reduce restrictions on the use of safe models, as restriction of model types can only weaken performance (see differential privacy section 10.1.3). Careful curation of queries which can be made to a released model restricts the capacity of potential attackers to accurately characterise the function implemented by the machine-learning model, and hence to perform membership- or attribute- inference attacks. This capacity to resist model attack through restriction of queries rather than restriction of model type can potentially allow more accurate models to reach the stage of public use, ultimately improving prediction performance (which may, e.g., result in better outcomes for patients when ML is used in health applications).

The category of model release type (unlimited and limited queries on the model) is important in determining the appropriate recommendations for model security.

## 10.2  Technical Recommendations:

### 10.2.1  TECH 1: Risks of disclosure need to be discussed at an early stage of the project application between the researchers and the TRE

Empirical analysis of a model can answer questions such as "are we able to successfully attack this model?" and "can we infer the value of this feature, given values for the other features?". It cannot determine whether this in itself represents a disclosure risk. TREs should therefore ensure that the risk pertaining to a particular dataset in question is discussed at an early stage and that researchers are aware of the kind of risks that will be checked for. An example of the kind of discussion that is required would be around attribute inference: by their nature, ML models learn correlations between attributes that can allow researchers to make predictions of one attribute based on the other. A released model may therefore allow people to make predictions of feature values for any individual, regardless of whether or not they were in the training set. A disclosure risk can occur when there is a discernible difference in the accuracy of those predictions for people in the training set, as opposed to those not in the training set. The TRE-researcher discussions should also include agreement on which attributes/features within a dataset are considered personal since not all will be. Similar discussions may take place around whether the risks of group disclosure need to be considered. For example, does the dataset include attributes related to groups who might be considered vulnerable to discrimination based on social factors (e.g. based on sexuality/religion etc.) rather than genotype etc?

Discussions between researchers and the TRE can help researchers to understand the risks and controls available to support their Data Governance and Ethics Approval Processes (see Ethics and Legal Recommendations:).

**Responsibility:** TRE staff and researchers; **Understanding:** Data Governance and Ethics Committees

### 10.2.2  TECH 2: TREs need to provide tools for training ML models within their environments

Many TREs have traditionally provided a relatively small number of software tools within the environment, e.g. statistical programs such as SPSS, STATA and R TREs [39]. Installation of such tools needs to go through security controls with regular patching (as they already do as part of best practices for managing any software within the TRE). To support ML training, TREs will have to install a range of additional tools and assess them for security risks, e.g. python ML-specific packages such as Tensorflow, scikit-learn etc.

**Responsibility:** TRE staff; **Understanding:** Researchers

### 10.2.3 TECH 3: The software/pipeline that was used to generate the final trained model which has been approved for release should be stored as a snapshot so that it can be reproduced for audit purposes

The data and pipeline that was used to generate the model should be saved so that it can be accessed if there are found to be any issues in the future. This may be required for regulatory processes.

This is a matter of accountability and tracking. Additionally, if 'inside' attacks are believed to be feasible, then clearly stating that production mechanisms must be auditable can be a disincentive to malicious attackers. In some scenarios – such as were governed by the Financial Conduct Authority or Medical Device certification, being able to account for and justify decision support models may be mandatory.

TREs should consider the technical requirements of maintaining the data and development pipeline to support these legal and regulatory requirements.

**Responsibility:** TRE staff and researchers; **Understanding:** Data Governance and Ethics Committees

### 10.2.4 TECH 4: Two broad categories of controls should be considered: controls to reduce the risk of personal data stored within the trained ML model (MDCs) and controls to limit the queries on the model (MQCs)

There are many advantages to releasing a trained model openly, including peer review, scientific reproducibility and re-use. However, openly sharing models increases the risk of releasing personal data. Therefore, to mitigate the risk of releasing personal data we are recommending many different MDCs which ensure trained models are anonymous. There can be some disadvantages to MDCs, such as the use of differential privacy for instance-based models, which may be viewed as too restrictive by some as they can also reduce the accuracy of models. Other controls require the TRE to understand the training process to run the required attack simulations which some researchers, especially those from industry, may be uncomfortable with regard to IP and trade secrets.

Rather than controlling the risk of releasing personal data within the model (MDCs), an alternative is to control the access to the model once it has been released by limiting the queries on the model (MQCs). For example, the model could be hosted within a web service which constrains who can query the model and how often. Utilising MQCs, the results returned from the model are anonymous to the end-user, even if the model itself contains personal data.

It is likely that for many projects either MDCs or MQCs are chosen. There may also be instances where a combination of MDCs and MQCs are employed. A combined approach might be appropriate, for example, where the model will be embedded within a software program which limits the queries on the model. There is a risk that such software can be illegally hacked, in which case some of MDCs may also be utilised as additional controls.

We suggest that the balance between MDCs and MQCs be guided by practicality, with the understanding that disclosure control is enforced with either option. Exclusive use of MDCs could facilitate easy use of released models by end-users, but possibly poor predictive model. Conversely, exclusive use MQCs could allow potentially stronger predictive models, but mean they are harder for future researchers to use.

**Responsibility:** TRE staff and researchers; **Understanding:** Data Governance and Ethics Committees

### 10.2.5   TECH 5: MDC: Disclosure control of ML models needs staff within TREs who are ML experts/trained

Our experiments (Appendix A) suggest that empirical testing needs to be performed on a model to determine how safe it is to be released (i.e., model disclosure risk cannot be reliably determined prospectively). This will require TRE staff to run experiments on the model. Although to a certain extent this can be automated, significant expertise is needed within TREs to - critically evaluate these results. The level of training and background knowledge required will likely be significant. A few days of training may be sufficient for someone with significant prior experience in applying ML to a range of complex datasets but is unlikely to be enough if they do not have this background 'know-how'. Each TRE does not necessarily have to employ someone directly with this advanced skill set. TREs could source this skill set from a pool of trained experts working across the network, or second an individual from another TRE to support a specific project.

It may be also that much of the checking could be carried out by less qualified staff. [40] categories traditional outputs for checking into 'runners' (outputs needing very simple approvals; can be done automatically), 'repeaters' (outputs requiring some human review at a low technical level) and 'strangers' (outputs requiring review by expert statisticians). As the expected share of outputs of each type is something like 90%, 9%, and 1%, TREs use this to manage output checking resources.

**Responsibility:** TRE staff

### 10.2.6   TECH 6: MDC: TREs should keep a proportion of relevant data to one side and never give this data to the research team to be used to assess the model

Running attack simulations requires two data sets – one data set containing examples that were used to train the model, and one containing examples that were not. To ensure that examples exist that have not been used for model training, the TRE should hold back some data from the researchers. In addition, it is important that the TRE staff must know which rows in the data given to the researchers were used for training. Note that to keep data separate, TRE staff will need to know how the data will be processed into rows. For example, if a row will represent individuals, then the TRE staff should set aside all rows corresponding to a subset of individuals. This is particularly important for episodic health data where each individual may have more than one row.

More research is required for recommendations on how the data should be selected. For example, should it be random and different for each project?

**Responsibility:** TRE staff; **Understanding:** Researchers

### 10.2.7   TECH 7: MDC: To assess the risk of a model containing identifiable data, TREs should run a range of checks on the model to be released

It should be noted that if the model will be released with additional controls on the queries of the model (e.g. hosted within a web service rather than released openly) (see Section 10.2.16), these checks may not be required. If the TRE has deployed, and the researcher used, 'safe wrappers'[25], then many of these checks are run automatically and so it may be advisable to do so, depending on the use case.

TREs often carry out risk assessments as part of their disclosure control process. These steps are additional steps to carry out on top of the existing risk assessments. Each of these checks is listed as sub-recommendations below.

**Responsibility:** TRE staff; **Understanding:** Data Governance and Ethics Committees, and researchers

### 10.2.7.1 TECH 7.1: MDC: Researchers should provide sufficient detail (in a standardised manner) to TRE staff to enable model assessment

Researchers should provide:

1. The trained model in a standard open file format (such as those recommended in scikit-learn/tensorflow/Karas or an open format such as ONNX).
2. The code that was used to train the model (the complete pipeline, including any data pre-processing).
3. Their assessment of model predictive performance.

Note that researchers must be made aware of this requirement at the start of the project as a condition of model disclosure (see ELA 2).

**Responsibility:** TRE staff & researchers; **Understanding:** Data Governance and Ethics Committees

### 10.2.7.2 TECH 7.2: MDC: TREs should run the model to be released against set-aside data to ensure that it provides the expected result.

TRE staff should assess the performance of the model on the data held out from the researchers. Poor performance relative to the performance quoted by the researchers is an indicator of over-fitting, which can indicate disclosive models. Performance that deviates significantly from that reported by the researcher should be reported back to the researcher and investigated before further analysis is undertaken.

**Responsibility:** TRE staff; **Understanding:** Data Governance and Ethics Committees, and researchers

### 10.2.7.3 TECH 7.3: MDC: TREs should check the size of the trained model

The model file size ought to be orders of magnitude smaller than the size of the training data (an exception is some large neural network models that can have more tuneable parameters than observations). If the model file size is of a similar order to the data size, researchers should be asked to justify this. It is worth noting that some ML packages by default (for instance, the 'glm()' function/class in R) include the training data within the saved model file (for reproducibility). This check will identify this issue.

**Responsibility:** TRE staff; **Understanding:** Data Governance and Ethics Committees, and researchers

### 10.2.7.4 TECH 7.4: MDC: TREs should eyeball the code used to train the model

TRE staff should have the access and expertise to inspect the code used to train the model to identify any (potentially accidental) poor practices (e.g. storing the data within the model). In addition, they should be able to check that the code is training the model in the manner described in the researcher's report.

**Responsibility:** TRE staff; **Understanding:** Data Governance and Ethics Committees, and researchers

### 10.2.7.5 TECH 7.5: MDC: TREs should check the file type of the model to be released to check that it is contained within a list of accepted file types for release

TRE staff should only accept model files in a set of common formats that can be loaded with common tools and inspected. Proprietary formats or formats that only allow the model to be run and not inspected should not be permitted.

**Responsibility:** TRE staff; **Understanding:** Data Governance and Ethics Committees, and researchers

### 10.2.7.6  TECH 7.6: MDC: TRE staff should run attack simulations using the set-aside data

Our results (Appendix A) show that model vulnerability cannot be prospectively determined for a model/dataset combination. Therefore, to have confidence that a model is safe, TRE staff need to empirically assess model vulnerability. The steps to do this are provided in Appendix G.

The GRAIMATTER DARE UK sprint project has developed as a minimal viable product (MVP) as a suite of attack simulations that can be applied by TREs (https://github.com/ai-sdc)[25].

The following principles should be utilised:

- Use of the same standardised data dictionary format.
- Definition of a format in which raw-data-to-design-matrix transformations should be specified; ideally both human and automatically readable.
- Make open-source on GitHub the code to run attack simulations which invites others to use it. Regular update frequency etc.
- Ideally, provide confidence intervals.
- Deal with different risk appetites for TREs.
- Add to existing GitHub site – (https://github.com/ai-sdc) [25].

**Responsibility:** TRE staff; **Understanding:** Data Governance and Ethics Committees, and researchers

### 10.2.7.7  TECH 7.7: MDC: TRE staff should not rely solely on average performance metrics (e.g. accuracy, AUC, etc) when evaluating simulated attacks

When evaluating the outcomes of inference attacks on a trained model, the 'use-case' of disclosure control has to be taken into consideration when determining which metric(s) to use to evaluate simulated attack performance.

Statistical Disclosure Control of 'traditional' statistical analyses does not permit the release of outputs that reveal information from a few respondents, even if the majority's privacy is respected. For instance, max/min values for attributes such as salary within a group are routinely disallowed.

Standard machine learning metrics (for example, AUC) measure the average performance of a model across large numbers of examples. Whilst such metrics can certainly indicate vulnerabilities, they cannot be used alone and should be used with metrics that target the extreme regions of the target model's predictive probability (for example, True Positive Rate at a fixed low False Positive Rate). This is due to the fact that a target model may give very confident predictions to a small number of training examples (perhaps they are somewhat abnormal and end up being memorised). The influence on AUC of this small set of examples may be insignificant with the result that the model appears safe. This argument was made in a recent study [41] and was backed up by our own experiments where we observed that metrics focusing on the extremes of the predictive probabilities picked up classifiers as significantly vulnerable that looked safe based on AUC (Appendix F).

**Responsibility:** TRE staff; **Understanding:** Data Governance and Ethics Committees, and researchers

### 10.2.8  TECH 8: MDC: Researchers should provide a 'data dictionary' describing the inputs to their model in a standardised format.

For any model to be useable for making predictions on new data, it is necessary to know what data should be provided to them, and in what format. If TREs (or preferably some higher body) agree on a standard format for this, then it is possible to write automated scripts that "attack" the models and provide the TRE output checker with useful information about the vulnerability of the model.

Information is needed because, to run various inference attacks on models, it is necessary to know the fields that they trained on and how they categorised them.

The data dictionary should include (note: similarities with Python pipelines and PyTorch data loaders):

- Name of the original dataset
- Source code function that takes as input the original dataset and performs any pre-processing (feature selection, feature creation, normalisation, one-hot encoding, handling of missing values, etc.)
- Description of the outputs of the pre-processing function (i.e., the inputs to the model): feature names, encoding (e.g., one-hot categorical or continuous), and a list of indices that represent each feature (e.g., feature indexes [1,2,3] may correspond to a single one-hot encoded feature).
- How the data is split into training, validation, and testing; the number of samples used?
- Description of the model outputs: type (classification, regression, etc.), encoding of the outputs (e.g., softmax, linear, etc.), number of outputs (e.g., number of classes).
- Description of the model: list of libraries and models used, architecture, training algorithm and how the model is initialised.
- Random seeds were used (for reproducibility).

Inputs

- Clear documentation of their model architecture, data used and how it is split
- A script that will enable the TRE to load the model and examine it
- A script that will enable the TRE to test the model

Furthermore, to help against inadvertently introducing the risk of backdoor or poisoning attacks, the TRE staff should check the transformation pipeline to ensure that:

1. The number of records is not increased unless specific data augmentation techniques have been agreed upon.
2. Transformations are not applied selectively to any subset of the records.

Appendix C provides an example data dictionary template.

**Responsibility:** Researchers; **Understanding:** Data Governance and Ethics Committees, and TRE staff.

### 10.2.9  TECH 9: MDC: To increase efficiency, researchers may wish to employ community-developed Safe Wrappers

Trained models should not be allowed to be released from TREs if they were trained with unsafe hyperparameters, with unsafe variants (for example, not using differentially private versions where these are available) or have been manually altered after the training algorithm had been run. Safe wrappers are methods which inherit from existing machine learning libraries, providing the same functionality but enforcing limits on relevant hyperparameters, and providing some traceability.

Safe wrappers can automate and report on many of the checks recommended above. Hence, they can act as a decision support tool for both the TRE output checkers and 'honest-but-curious' researchers wishing to explore the boundary between models' vulnerability and accuracy as they fine-tune their learning regimes.

A safe model version will check if the researchers have complied with a safe training practice without any malicious intent and a report is provided to the output checker.

While we can know that some hyper-parameter/algorithm variants typically lead to disclosive models, it is in the nature of logic that we cannot know the converse in advance for most algorithm-dataset combinations. Therefore, safe wrappers should also include functionality to run automated attacks on the specific trained model the researcher wishes to release.

The GRAIMATTER DARE UK sprint project has developed a set of minimal viable product (MVP) safe wrappers building on python scikitlearn/TensorFlow toolkits employing the safe parameters determined by experiments across a range of models, data types and hyperparameter settings (Appendix A). These are available to be used and tested by the community from *github.com/ai-sdc* [25].

The community of TREs and researchers should build many more such safe wrappers to meet the needs of the research community. e.g. R library versions. As the wrappers themselves do not contain any personal data, they can be widely shared across TREs rather than having to be developed bespoke for each TRE.

TREs should encourage researchers to utilise Safe Wrappers. This will reduce the length of time a TRE will take to verify that a model can safely be released. Furthermore, there is a distinction between a researcher knowing that *some* choices may lead to unsafe models (a concept that needs to be learned once), and the specific knowledge of *which* those choices are (a subject of ongoing research effort). If the definitions of 'unsafe' hyper-parameter combinations and algorithm variants are held in a central (human and machine-readable) file in the TRE, then this can be easily maintained and updated. The alternative is to rely on continuously updating and re-delivering the training provided to researchers.

The rationale for using wrappers:

1. We recognise that manually checking trained ML models will place a huge resourcing strain upon TREs and researchers. Without some form of support, this is likely to lead to lengthy delays in the release of models, or possible limited TRE appetite for supporting the use of ML and for researchers to use TREs.
2. Incentivisation (e.g. faster approval of results) will help to overcome resistance from researchers.
3. Many researchers will happily accept suggestions of 'safe' hyper-parameter ranges or algorithm variants. While researchers should be free to override these suggestions (with the understanding that they will still need, in a way that will be approved, to make the model 'safe' for it to be allowed to be exported from the TRE), those are conscious design choices with attendant risks which should be reported to the TRE output checkers to make a principles-based decision.
4. There is a high degree of correlation between a trained model's vulnerability to privacy attacks, and the risk of it 'over-fitting' the training set and hence failing to generalise. Therefore, the use of wrapper models is intended to encourage good practice and should not significantly impact the accuracy of the trained model.

5. Wrapper approaches form a natural way for triggering/embedding automated testing of the privacy risks associated with specifically trained models so that all the information is to hand when the TRE output checker comes to deal with a request for model release, rather than introducing further delays into the process.

**Responsibility:** Researchers; **Understanding:** Data Governance and Ethics Committees, and TRE staff

### 10.2.10 TECH 10: MDC: Safe wrappers should be developed following a set of principles

Clearly, there will be slight changes depending on the features supported by different languages. In the section below, we describe them with reference to python as this is the predominant language for Machine Learning development and libraries.

1. To ensure consistency of behaviour across different ML models or implementations, as much of the 'safe wrapper' behaviour as possible should be implemented within a superclass. This class should include functionality for checking hyper-parameter values, making checkpoints of the models to guard against malicious tampering, saving models, and producing reports for TRE output-checking staff.
   a. For example, in our implementation, we have called this class SafeModel.
   b. Classes for specific model types should then use multiple inheritances so they contain as little code as possible. **The philosophy here is to augment the functionality of existing code, not to re-implement it**.
   c. For example, our class SafeDecisionTreeClassifier() inherits from both SafeModel and sklearn.trees.DecisionTreeClassifier() as super classes.
2. All constraints on hyper-parameter values needed to prevent excessive disclosure risk should be stored in a human and machine-readable file held centrally by the TRE. We suggest the use of a single file, with write access limited to the TRE administrators for security and consistency reasons.
   a. An example of part of such a file is provided in Appendix D.
   b. We will be publishing recommendations for constraints for different model types as part of our full report. Naturally, these will be subject to change, so an online version will be made available.
   c. TREs should put in place schedules for periodically checking their constraints files are up to date, in the same way as they will have schedules for checking for updated versions of other software.
3. Whether there is a differentially private version of the optimisation algorithm used in training a model, such as for Support Vector Machines, and Artificial Neural Networks (e.g. Tensorflow privacy) this should always be used.
   a. This is one example of where the use of class inheritance naturally provides straightforward mechanisms for overriding default choices/optimisers. It would make sense for the differential privacy 'epsilon' factor to be defined in a single place in the constraints file to be consistent across different model types.
4. Safe wrappers should consider attributes or parameters of a model that researchers might unintentionally or maliciously change between 'fitting' a model, and the model is saved, and release requested.
   a. One obvious example is a researcher training a model with unsafe parameters and then changing the parameters to 'safe' values but not retraining
   b. Snapshotting a copy of the fitted model after the automated training and comparing this to the model the researcher requests to release is one useful mechanism.

    c. In our example implementation, we have found that although much of this process can take place in the superclass, knowing exactly which model attributes to check and what format they take in python needs to be done for different implementations of algorithms.

5. Safe Wrappers should be made open-source and shared with the community to maximise benefit and improve the ongoing maintenance

6. *Safe Wrappers should make sure training data is not included in the model: for example,*

    a. *a Safe_KNearestNeighbour() class might just return a message saying this algorithm is not permitted since it inherently encodes the training data.*

    b. *They should detect, and report on, any extra attributes/values that the researcher has added to the model in addition to those created by the underlying library implementation*

    c. *They should check that the existing model parameters/attributes only contain the expected information.*

7. Where Safe wrappers are being developed for different languages, they should provide equivalent functionality in different languages.

**Responsibility:** TRE staff; **Understanding:** Researchers

### 10.2.11 TECH 11: MDC: Instance-based models or models where the high numbers of parameters risk memorisation (such as deep learning) should have additional controls such as training on aggregated data or using differentially private methods

Instance-based models are popular within ML (such as KNN, SVM and Gaussian Processes). However, they require special treatment as, in their default use, they need all (KNN) or some (SVM) of the original data rows to be able to make predictions. It is vital that TRE staff are cognisant of this, and can ensure that, for example, researchers are unable to release a standard Support Vector Classifier. Researchers should be informed at an early stage of project scoping that release of instance-based models (in their standard form) will not be permitted.

Models where the high numbers of parameters risk memorisation (such as deep learning) may also pose an additional risk.

TRE staff require the expertise to be able to assess if the model under consideration falls within this category so that they can ensure that the researchers have taken the necessary steps to ensure disclosure.

There are several options for additional technical controls for instance-based models or models where the high numbers of parameters risk memorisation:

- **Train on aggregated/binned data instead:** although unlikely to be practical in many cases, if the training data can be de-identified (by, e.g. achieving k-anonymity via aggregating individuals and binning disclosive features) then there is no danger with rows being disclosed in the trained model.
- **Use differentially private (DP) training algorithms or training algorithms with anti-overfitting:** Some instance-based models have been adapted to use differentially private training algorithms. These place provable bounds on the disclosure risk of the trained model, and ought to be used where available. For example, the GRAIMATTER DARE UK sprint project has developed as a set of minimal viable product (MVP) safe wrappers for SVNs and neural networks utilising DP versions of the libraries. An alternative to DP model training is model training with anti-overfitting techniques (regularisation and dropout), since [30] shows that anti-overfitting techniques may yield better accuracy-privacy trade-offs than DP.

- **Train on synthetic data generated by a DP mechanism:** If neither of the previous steps is appropriate, a final option is generating synthetic data (in a private manner) and then training the model on this synthetic data. When trained in this way, we can think of instance-based models as no longer being instance-based, since the 'instances' they contain should be synthetic data points only. However, care must be taken that the synthetic data generator is generating samples sufficiently different from those in the training data; see notes on synthetic data.

Employing any one of these 3 additional technical controls can reduce the accuracy of the model. Researchers should consider instead limiting the number of queries on the model as an alternative option.

**Responsibility:** Researchers and TRE staff; **Understanding:** Data Governance and Ethics Committees

### 10.2.12 TECH 12: MDC: Researchers and TREs should make informed choices when using Differentially Private Methods

Since DP mechanisms are necessarily random, the material released from the TRE using the same method on the same data will be different each time. Importantly, DP guarantees are violated if several such releases are compared and one is chosen. As an example, if researchers firstly fit a non-DP model to some data, then set up a DP algorithm, generate fifty potential releases using this algorithm, choose the one of these fifty that most closely resembles the non-DP model, and release this 'best' model, then the net procedure is no longer DP. It is thus imperative that the randomisation inherent to the DP algorithm be dependent on a random seed, and that the seed corresponding to the final release is chosen by TRE staff who are agnostic to how well the released model performs.

We recommend that the use of DP alone is no substitute for following best practices, such as adhering to techniques to limit overfitting (e.g., data augmentation, regularization, dropout in ANNs, etc.) as well as performing attacks on trained models to empirically estimate their vulnerability. This recommendation stems from two main concerns with differential privacy in machine learning. First is the utility loss: differential privacy imposes an accuracy loss on the trained models, especially for small values of epsilon. This may not always be acceptable, especially in the biomedical sector. Second, the only available theoretical bounds on the privacy leakage of ML models are for epsilon values below 1 (at which point the accuracy loss is high), and these are not tight bounds. Values of epsilon above 1 still offer (empirical) protection, but since there is no theoretical bound on this level of protection, empirical risk assessment is still needed [30], [42].

**Responsibility:** TRE staff and researchers; **Understanding:** Data Governance and Ethics Committees

### 10.2.13 TECH 13: MDC: Ensemble methods should receive special treatment

Ensemble methods are ML methods in which several individual ML models are combined (e.g. taking the average of their outputs, or having them vote). TREs should take care to ensure that both the overall model is safe, and the base models that constitute it are safe. It is straightforward to make examples that show both that safe base models can create an unsafe ensemble, and that a safe ensemble can include unsafe base models. It is recommended that TREs do not assume that a safe ensemble implies safe base models or vice versa. This risk is only relevant in circumstances when an attacker has access to the inner workings of the model (white box attack), and not when all they can do is query the ensemble model.

**Responsibility:** TRE staff; **Understanding:** Data Governance and Ethics Committees, and researchers

### 10.2.14 TECH 14: MDC: When using Federated Learning the final models should be tested for vulnerability

Federated learning in this context describes an ML architecture where the training data is spread across multiple TREs and a 'local' model is trained in each TRE which is iteratively updated with learned information from all the other TREs. At each iteration, the local model is shared with a 'central' TRE which aggregates the multiple models into one which encompasses information learned over all the TREs. The aggregated model is shared back with the other TREs for further training there. This continues until training is complete and the aggregated model is the final trained model. Identifying if a model is disclosive needs to be done using the data on which it was trained, which is not possible for the final aggregated model as the central TRE does not have access to the other TREs' data. Therefore, it is recommended that within each TRE, the local models are tested for vulnerability. It can be assumed that the federated TRE network is trustworthy, following governance approval, and therefore only the final versions of the models are tested for disclosure by running attack simulations within all the TREs. The final aggregated model can be released only if all attack simulations are completed (see Figure 7).

**Responsibility:** TRE staff and researchers; **Understanding:** Data Governance and Ethics Committees

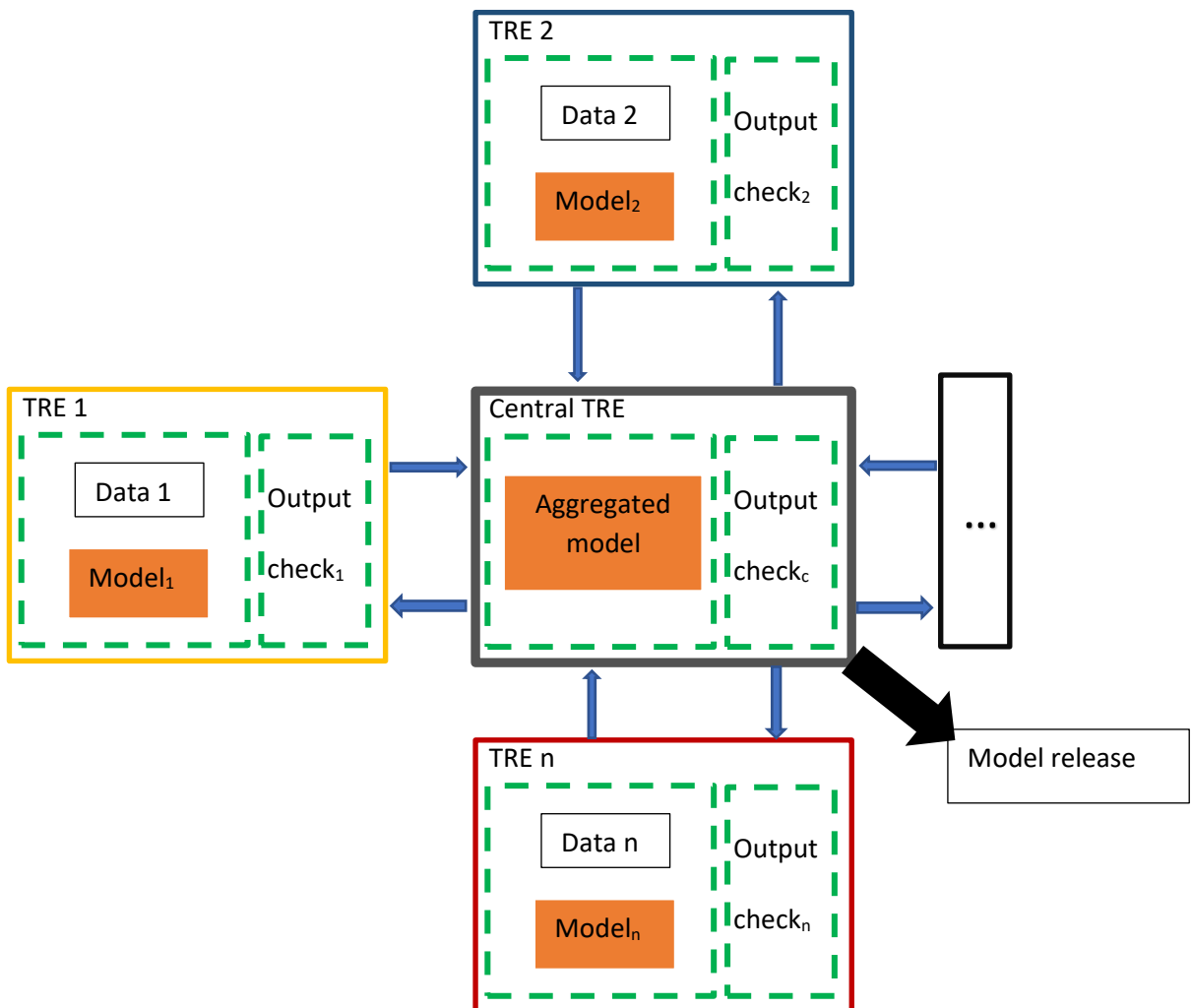

*Figure 7 Federated learning proposed checks*

### 10.2.15 TECH 15: MDC: Synthetic data should still be considered to be personal data

Synthetic data generators (SDGs) can often produce arbitrary amounts of data from a fixed sampling distribution, so should be thought of as specifying a *distribution* rather than a *dataset*. A classifier can be thought of as a conditional distribution of outcomes given covariate values, but an SDG specifies a joint distribution of covariates and outcomes.

We recommend that

- If an SDG is used exclusively inside a TRE, the aggregate process of generating synthetic data and fitting a classifier to this synthetic data should be considered as a classifier fitted to original data and tested as described above.
- If an SDG is to be released from the TRE, a provably differentially-private fitting mechanism should be used for the SDG. The level of differential privacy should be specified before beginning data analysis. Users of synthetic data generated by the released SDG should be made aware that the sampling distribution of the SDG may be different from that of the original data, and if models are fitted to the synthetic data we would expect attenuation in performance when used on non-synthetic data.

*If an SDG and a classifier are both to be released from the TRE, we are investigating if we should recommend that these be fitted to disjoint datasets.*

**Responsibility:** TRE staff and researchers; **Understanding:** Data Governance and Ethics Committees

### 10.2.16 TECH 16: MQC: Controls should be considered to limit the queries on a model once it is released from a TRE

Limiting the number and source of queries on a model is much safer than releasing it for external deployment as the usage can be tracked and any odd behaviour detected so that appropriate action can be taken.

There are several ways to accomplish this. For example:

- **Secure Web Service:** the model could be held securely within a web service which receives query data and returns only the answer from the model. Such a web service could be run by the TRE itself, a trusted third party or by the organisation of the researcher. Controls for a model deployed within the secure web service include:
  - o restricting access to a small number of IPs that can query it, e.g. if the model is only used by NHS Scotland, the web service could be configured to reject all other incoming query traffic.
  - o Maintaining a log of all queries received can help identify an attempted attack (e.g. if there are a high number of repetitive queries from the same IP this highlights an attempted attack).
  - o Restricting the number of queries within a time period.
- **Controls on a software program:** the software program which embeds the trained model should include controls around the number of times a particular end-user can query the model within a time period. However, there might be practical difficulties in ensuring that the program is used in this way and can't be reverse-engineered.

When an end-user query the model which has MQCs applied, the results returned can be considered to be anonymous and therefore not do not fall within data protection legislation.

Depending on the method selected to provide MQCs different groups could be responsible for the implementation. Who is responsible for implementing these controls should be clearly articulated within Data Governance and Ethics applications.

Where there is a risk that the controls could be circumvented (for example, by user collusion), additional technical controls, such as those listed within Sections 10.2.3(TECH 3) to 10.2.12(TECH 12), should be considered.

The scope of the GRAIMATTER project did not allow for a detailed investigation of MQCs and therefore these recommendations should be implemented with caution. We plan further research in the future.

**Responsibility:** Researchers; **Understanding:** Data Governance and Ethics Committees, and TRE staff

### 10.2.17 TECH 17: MQC: Controls placed on a model to limit the number of queries should be tested by an external party

In addition to internal security processes, TREs are regularly penetration-tested by an external party to check for security holes. If it is required that controls are implemented to limit the number of queries on the model. External penetration tests of these controls should be considered (consistent with model risk) to ensure that there are no security holes which could mean a hacker could circumvent the controls. For example, a web service which is hosting the trained ML model should be externally penetration-tested.

**Responsibility:** Researchers; **Understanding:** Data Governance and Ethics Committees, and TRE staff

# 11 Ethics and Legal Aspects (ELA)

## 11.1 ELA Background

We have investigated the legal and ethics issues accompanying ML model release from TREs. We identified and assessed how current UK legislation applies to TREs supporting trained ML model release and the extent to which the legislation addresses ethics issues pertinent to the release of ML models out of TREs. We investigated issues such as transparency, privacy, data protection and non-discrimination. We looked at the main legal obligations incumbent on the researchers and TREs, including protecting the confidentiality of the personal data held by the TRE and used for training ML models. We took into account the duty of Data Controllers to protect confidentiality but also share data in the public interest. In particular, given the aforementioned identification of data breaches as a major risk for ML model release from TREs, we investigated how current UK laws apportion responsibility for a personal data breach, misuse of private information, and breach of confidential information in research projects when releasing trained ML models from the TRE.

We drew from the field of AI ethics and governance from an international level (UNESCO Draft Recommendation on the Ethics of Artificial Intelligence [43]) and EU level (proposed Artificial Intelligence Act) to inform our analysis, especially in considering and guiding the reform of applicable UK frameworks. We considered whether a bespoke ethics impact assessment is needed before the release of a trained ML model, based on TREs data, as suggested in the UNESCO Draft Recommendation, and potentially also after the release through an ethics-based auditing system.

Our research on the security of the machine learning model considers the specific provisions of the Data Protection Act 2018 which aims to:

- demonstrate the need for a more detailed and particularised implementation of appropriate technical and organisational measures [7].
- acknowledge the requirements of the Data Protection Act 2018, that a level of security commensurate with and appropriate to the risks [7] arising from machine learning models is urgently required as these tools are increasingly used to process personal data.
- acknowledge the regulatory need for logging and, in particular, logging the identity of the person [7] who consulted the data, despite the lack of joined-up and interlinked registration systems to assist with the traceability of an ML model. As a result, these recommendations partially address the UNESCO call for traceability, human oversight and determination [43] for "any stage of the life cycle of AI systems".
- provide a stepping stone towards the roadmap to an 'effective AI assurance ecosystem'[44].
- incorporate the need for "greater algorithmic transparency and accountability"[45].
- incorporate the call for answerability, which can manifest itself in a continuous chain and designation of human responsibility [47], as proposed by Leslie in the Alan Turing Institute Guide for the Responsible Design and Implementation of AI systems in the public sector.

## 11.2 Ethics and Legal Recommendations:

The personal data used in TREs are governed by data protection law if it exists in a particular jurisdiction. Internationally, the European Union's General Data Protection Regulation (GDPR) [26] is the most prominent data protection framework, and the UK implemented it into domestic law before it left the EU as a Member State [48]in the Data Protection Act 2018. Given our focus on UK-based TREs, we proceed with some key points about their application.

The key challenge is responsibility for a possible data breach resulting from the release of a trained ML model, where the output checker does not have access to human-readable outputs. The aforementioned attacks (MIA and AIA) may lead to the identification of personal data, thereby rendering the model a personal dataset, to which data protection law would apply.

Understanding also that the disclosure of a trained ML model, which is not as human-readable as typical TRE releases such as the disclosure of standard statistics (e.g. graphs and tables etc.), raises certain risks and complications for TREs is important. The security of processing is a responsibility attributed to the controller and the processor, as outlined in Article 32 of the GDPR [26]. The controller and processor must take "state of the art", technical and organisational measures into account. These include "appropriate" measures commensurate with the risks, in terms of (a) pseudonymisation and encryption, (b) ongoing resilient systems and services, (c) ability to restore personal data (PD), and (d) regular testing for effectiveness of the security of the processing.

For a trained ML model, the chain of responsibility may be unclear where the initial Data Controller and TRE processor are no longer involved once the ML model has been transferred out of the TRE, and if the responsibility for any potential future personal data breach has not been identified nor allocated correctly in the terms of contractual agreement that a TRE operator has made with a researcher. These are issues we aim to address in these Ethics and Legal Recommendations:.

Our recommendations here relate to the application of data protection legislation to trained ML models once they leave the TRE, the insertion of new contractual terms in existing contractual agreements or linked agreements which researchers sign when accessing TRE data, and the ethics and data governance review process for projects using TREs to produce ML models.

### 11.2.1 ELA 1: Data Controllers, TREs and researchers should consider that Data Protection legislation may apply to the trained model and data sharing agreements may be required

Data protection legislation (in the UK, the Data Protection Act 2018 which currently implements the EU GDPR) is generally considered not to apply to the classical statistical anonymous aggregate level releases from a TRE as appropriate controls ensure that these releases do not contain personal data. A trained ML model, in contrast, may be considered to contain pseudonymised personal data including special category personal data, therefore requiring specific technical and organisational measures to ensure the processing is compliant with data protection law. In particular, it needs to be considered whether appropriate data security measures have been adopted to reflect the risk of a data breach.

Assessing the level of risk as to whether the trained ML model and/ or its output are inextricably linked to the personal data it was trained on will determine how the law categorises the model. We consider the following 3 categories:

1. A trained ML model can be considered to only contain anonymous data and therefore data protection law may not apply (as would be the case for aggregate level releases from classical statistical data analysis or if the MDCs (recommended in the Technical section) have been robustly applied to a trained ML model).
2. A trained ML model is considered to potentially contain pseudonymised personal data, therefore requiring specific technical and organisational measures to ensure the processing is data protection law compliant. In particular, whether appropriate data security measures have been adopted to reflect the risk of a data breach. Certain forms of attack (model inversion and membership inference) may render ML models as personal data(sets) [19].
3. A trained ML model is considered to carry more risk of including personal data given the risk of the deliberating 'hiding' of data or vulnerable parts of a model.

Data Controllers (through the Data Governance Committees) may need to consider legal protection in contractual terms to govern the transfer of responsibility, obligations and rights to a new Data Controller/processor associated with the release of category 2 or 3 trained ML models, ensuring prior written authorisation of the controller, or indeed the retention of responsibility, obligations and rights by the original Data Controller in the contract. In this instance, new contractual agreements and/or new terms in existing contractual agreements will be required between the Data Controller and researchers before the ML model is released from the TRE environment e.g. data sharing agreements.

**Responsibility:** Data Governance and Ethics Committees, TRE staff and researchers

### 11.2.2 ELA 2: Existing Data Governance and Ethics Approval Processes should consider a range of risks and controls with each application providing sufficient details to support the review process

Both Data Governance and Ethics Approval Processes should consider the risks and controls associated with the training and release of ML models, particularly as regards the possibility that such models may contain personal data, as this is a different scenario from conventional TRE releases which would not typically contain personal data. In this section, we term these processes as approval processes rather than spelling them out each time.

As described in the Scope section, both researchers and projects must go through both the Data Governance and Ethics Approval Processes to access TRE data. In their applications, researchers need to articulate the wider benefits of their work when they wish to release an ML-trained model and explain how they will secure the models. This is especially important when ML-trained models may be released as full consideration of benefits may add weight to the value of the research versus additional risks from disclosure.

It should be the responsibility of the researchers to provide sufficient details of the likely risk and controls for the Data Governance and Ethics Committees to review. These risks and controls should be revisited and finalised before the trained ML model is released from the TRE as a collaborative process between the researchers and the TRE. Any significant changes to the risk profile compared to what was outlined in the initial approvals should be notified to the approval bodies and new/amended applications for approval may be required in such circumstances.

**Responsibility:** Data Governance and Ethics Committees; **Understanding:** TRE staff and researchers

### 11.2.2.1  ELA 2.1: Approval processes should consider the additional disclosure risks and the controls used to protect data confidentiality and the specific legal and ethics implications

The following should be considered by the approval process:
- Information on the release of a trained model and any safe channels to be used to facilitate the release/deployment/license/transfer.
- The risks, the risk spectrum, controls and benefits for the ML model release:
  - How the technical recommendations provided in the Technical Recommendations Section will be implemented/observed e.g.
    - TRE staff will run attack simulations and the researchers will use safe wrappers to reduce the risk that the trained ML model contains personal data (MDCs)
    - Or, the risk the trained ML model contains personal data is significant and therefore controls will be placed on the queries of the model by hosting the release within a secure web service (MQCs).
  - The appropriateness of releasing the ML model considering the risks
  - The purpose of the release
  - The public benefits of the release
  - The benefits of the mode of the release e.g.
    - The trained ML model is shared openly, after being subject to appropriate disclosure controls, supporting peer review, scientific reproducibility and re-use.
    - Or, the trained model will be hosted within a secure web service. As the queries on the model will be controlled, privacy-preserving methods (required to safely release the model openly) will not be required resulting in a higher degree of model accuracy.
- Mechanisms for the traceability of released ML models, as ways to address issues of trust and trustworthiness.
- Whether the organisations involved are considered 'safe' e.g.:
  - The organisation responsible for applying any controls to limit the queries on a model
  - The TRE
  - The organisation the researchers and any collaborating researchers are from (academic, public sector, industry)
- Potential controls to stop the model from being used for another purpose than listed within the approval documentation e.g.
  - If the model is released completely openly with no conditions, then such controls would be added to an EULA (ELA 6).
  - If the model is embedded within software or hosted by a trusted third party, controls should be included to necessitate an amendment or a new approval process for an alternative purpose.

- o The consequence of legal sanctions in form of breach of contract if the model is used for other purposes as this would breach the Data Security Measures as per the suggested clauses to be added to the EULA etc.
- o Due diligence is to be undertaken to ensure the trained ML model does not disclose certain individuals or groups to safeguard their privacy [49].

In some cases, approvers might decide that the privacy risk associated with the disclosure may vary between (types of) attributes within the same dataset. In such cases, it may be appropriate to provide researchers and output checking staff with a 'risk appetite' on a feature-by-feature basis. It may be appropriate for some ML models not to be released from a TRE environment as they are too risky.

Experience shows that actively seeking to engage with data providers (rather than just making material available) tends to produce more positive outcomes. Data Governance Committees include representatives from the Data Controllers who can approve projects. We recommend that researchers engage with these Data Controller representatives prior to application submission, particularly if they view the project as high risk.

**Responsibility:** Data Governance and Ethics Committees, and researchers; **Understanding:** TRE staff

### 11.2.2.2 ELA 2.2: Data controllers/Data Governance Committees should be able to mandate a time limit for the use of a model, after the expiry of which the researcher needs to seek new approvals from the TRE

Where a model is not openly shared on release, it may be possible that controls are placed on the time the model can be used without requesting fresh approval. Approval processes should consider whether a time limit is appropriate and if so, the controls that would be put in place to ensure that the model is not used after this time period without fresh approvals being granted.

**Responsibility:** Data Governance and Ethics Committees; **Understanding:** TRE staff and researchers

### 11.2.2.3 ELA 2.3: Approval processes should mandate necessary requirements to keep the data and the pipeline available to meet legal requirements

There may be regulatory or legal requirements to keep a copy of the data and pipeline used to generate the trained ML for audit purposes for several years (e.g. if the model is used in a medical device such information may have to be kept for 15 years). The approval process should consider this requirement and the feasibility of meeting this need. The approval process should consider the implication for the persistence of all of the organisations involved to meet these requirements, i.e. what would happen should the TRE or another relevant party no longer exist within the time period.

**Responsibility:** Data Governance and Ethics Committees; **Understanding:** TRE staff and researchers

### 11.2.2.4 ELA 2.4: Consideration should be taken regarding the correct language to describe the process

Determining the correct language for different uses of the trained model is important. The following terms could be used to help explain how a trained model can travel from a TRE to another TRE or elsewhere: release, deployed, used, licensed and transferred. Language is also important for determining who retains ownership, and/or control of the trained model, and how the technical and organisational measures are logged under existing Data Protection Act 2018 rules, in addition to the traceability obligations in terms of the life-cycle of an AI system. These are key issues that need to be addressed by the Data Governance and Ethics application processes.

**Responsibility:** Data Governance and Ethics Committees, and researchers; **Understanding:** TRE staff

##### 11.2.2.5   ELA 2.5: Researchers should complete a DPIA, the ICO AI and Data protection risk toolkit, and the Ada Lovelace Institute' algorithmic impact assessement and provide these as inputs to the Data Governance and Ethics approvals process

Researchers should complete a Data Protection Impact Assessment (DPIA), the Information Commissioners Office AI and data protection risk toolkit [50] and the Ada Lovelace Institute's algorithmic impact assessment [51], and provide these as inputs to the Data Governance and Ethics Approval process.

**Responsibility:** Researchers; **Understanding:** Data Governance and Ethics Committees, TRE staff

##### 11.2.2.6   ELA 2.6: Data controllers should have the option of refusing to release a model or recalling a released model if it is deemed that the risks of release/continued release are too high

If the TRE (acting on behalf of the Data Controller) deems the model to be too risky to release, it should be able to decide not to release the model. The Data Controller should also be able to stop the continued use of a model after release if it is shown to pose excessive/unacceptable risks to data security, privacy or other human rights (this may not be technically possible if the model has been openly released after MDCs have been applied i.e. there are no MQCs applied).

**Responsibility:** Data Governance and Ethics Committees and TRE staff; **Understanding:** Researchers

##### 11.2.2.7   ELA 2.7: Data Governance and Ethics Committees may wish to outsource the expertise to review applications

It may be that Data Governance and Ethics Committees have the expertise to review such applications after their training (TR 4) or they may wish to share the application with an external expert to independently review.

#### 11.2.3   ELA 3: Legal contracts/contractual terms should be required to cover the responsibilities of each party if MQCs are required after model release from the TRE

If MQCs are required after release from the TRE then Data Governance and Ethics Committees should consider the requirement for legal contracts/ the insertion of new contractual terms in existing contracts, e.g. a data sharing agreement, to ensure compliance with the responsibilities of each party to implement and maintain such controls.

A data sharing agreement could include the responsibility of the researchers to ensure that these requirements are met in any subsequent agreements they enter into e.g. EULA or a sub-contract (ELA 6). The researchers will be in breach of relevant contracts with the Data Controllers if the end-user does not fulfil these responsibilities and may be liable for damages for breach.

For example,

- Researchers must implement these controls vis-a-vis the MQCs and the researchers/their organisation is liable for any loss or damage if these controls are not implemented.
- A company running a web service could be contractually required to ensure that the controls to limit the number and source of queries are in place.
- If a TRE is responsible for running the web service, the researchers who trained the ML model may wish to add an availability requirement into the sub-contract with the TRE.

- Exported ML models should not be allowed to be used for another purpose than the one listed in the approval within the approval documentation. This should be controlled by a contractual agreement (e.g. EULA and also other contractual agreements between the Data Controller and researchers) which forbids such activity. It would be the responsibility of the researchers who trained the ML model to add such clauses to EULAs with which they make the ML model available to other users.

**Responsibility:** Data Governance and Ethics Committees, and researchers; **Understanding:** TRE staff

### 11.2.4   ELA 4: Data Use Registers should be extended to include information on trained models and detail about their release and controls

It is recognised as good practice for TREs to keep a record of all of the research projects they support and which datasets were used. Health Data Research UK drafted a white paper providing recommendations for a data use register standard [3], supporting TREs to make such data publicly available within a standardised format.

We recommend that the data use registry standard is enhanced/extended to specifically include:

- A category of release is a trained model
- Details covering what the model was trained to do and the datasets it was trained on
- A new entry for the date prior to approval was granted for the release of the trained ML model from the Controller
- A new entry for the date the Controller granted permission to change the "Permitted Use" of the trained ML model to include a new entry note detailing the extent of the new "Permitted Use"
- A new entry for each time a newly trained model is released
- A record for the attribution of responsibility between controllers/processors, processors/sub-processors and how these changes over time to enable it to be added to any contractual agreement in the future
- The intended and actual uses of the trained ML model
- Information on any regulatory approvals of the model and associated dates e.g. medical device regulation or CE mark
- The controls are utilised to reduce the risk of personal data being released to an acceptable level.

The use of such a data use register could be extended by requiring the compulsory registration of the ML model to facilitate its auditing progress. This could be facilitated easily by modifying and increasing the information collected by the HDR UK data use register, as described above. This information could include updated registration numbers which are continually logged on the Information Commissioner's Office register of controllers and new processors. This information could include up-to-date certified audits (reference number(s)) by the Central Data and Digital Office for high-risk machine learning models [46]. This information could include "written advice" from the [7] Information Commissioners' Office following a "prior consultation" with the Information Commissioner for the use of a high-risk machine learning model [50]. This information can then be linked to a contractual agreement.

**Responsibility:** TRE staff; **Understanding:** Data Governance and Ethics Committees, and researchers

### 11.2.5 ELA 5: Researchers, TRE staff and Data Governance/Ethics Committees should be required to complete ML model training courses

Completing these courses (see Training on Training) should be an obligation that researchers must fulfil before accessing TRE data and releasing an ML model. TRE staff and those who sit on Data Governance and Ethics Committees should also complete the courses targeted to their specific audience.

The training courses of researchers, TRE staff and Data Governance/Ethics Committees should include legal and ethics aspects of ML model disclosure (as described within the Training Section).

**Responsibility:** Data Governance and Ethics Committees, TRE staff & researchers

### 11.2.6 ELA 6: Clauses should be added to the EULAs of any resulting trained ML model

Legal terms should be included within EULAs which

- prevent attacking of the trained model to disclose any personal data
- prevent use of the model for a different purpose other than the intended purpose
- Prevent transfer learning.

EULAs should apply to all types of end user access to the model regardless of the source e.g. software which is encrypted and embeds a model (such as within a medical device installed within a hospital), software which is shared openly (such as uploaded to a GitHub repository), and when the model is queried via a webservice.

The terms of use should include explicit legal terms which stipulate that appropriate data security measures must be taken to ensure the security of any personal data and to mitigate the risk of its disclosure. A legal term outlining the "permitted purpose" of the trained ML model should be added. This would restrict future use outside the "permitted purpose" assigned. To cater for the potential for a future change to the "permitted purpose", a condition should be added that prior approval and consent from the data controller of the ML model should be sought prior to such use and change of purpose.

It could be argued that such end-user terms of use are not required as the MDCs and the MQCs recommended here provide technical controls which means that end-users should only have access to anonymous data and hacking is illegal under the computer misuse act. However, these additional end-user clauses provide additional clarification and risk mitigation.

As part of the GRAIMATTER project, we have drafted template legal clauses which could be utilised within EULAs. It is important to note that these clauses require further research, input and final approval from a qualified and insured legal team. Such clauses would need to be approved by the relevant Data Controller. Please contact the authors for access to these drafts.

**Responsibility:** Data Governance and Ethics Committees; **Understanding:** TRE staff, researchers & end-users

### 11.2.7 ELA 7: Additional clauses should be added to researcher declaration forms

Researchers often must read and sign researcher declaration forms to analyse data within TREs. Such forms explain the researcher's responsibilities and behaviours to which they must adhere. We recommend additional clauses are added to researcher declaration forms to cover the training and release a ML model, including:

- Researchers must abide by the controls which have been approved by Data Governance and Ethics Committees.
- Researchers agree that details of the project will be recorded on the data use register.
- Researchers are required to submit a new or updated Ethics and Data Governance approval documentation with any changes to the use of the trained ML model once it leaves the environment i.e. if it is incorporated into a product with a CE mark and is being sold commercially. This may involve a new or modified DPIA to be undertaken.

For instances where the Data Governance and Ethics approvals require MDCs, these new clauses should be added:

- There is a responsibility on the researchers/their company or organisation to carry out due diligence of disclosive personal data within the trained model and in such cases must report it to the TRE and follow rules (whether those already in existence or augmented) about how to address such a situation and e.g. de-identify the personal data/otherwise mitigate the data breach
- Researchers agree that the TRE staff can run attack simulations on their model
- Researchers must be aware that to disclose a trained model they will need to provide information on their data-to-variable transformation to facilitate attack simulations by TRE staff in both human- and machine-readable ways.

As part of the GRAIMATTER project, we have drafted template legal clauses which could be utilised within researcher declaration forms. These clauses have been drafted to a stage where further research and input is required, and final approval from a qualified and insured legal team. Data controllers would have to agree such terms are acceptable. Please contact the authors for access to these drafts.

**Responsibility:** TRE staff; **Understanding:** Data Governance and Ethics Committees, and researchers

### 11.2.8   ELA 8: Approval processes may wish to consider a risk-based approach vis-a-vis personal data

We recommend a risk-based approach is taken by Data Governance and Ethics Committees in their decision-making. This risk assessment should be revisited and re-assessed by the TRE before the ML model is exported. A high-level example is provided in Table 1. Such risk assessment information may be included within the documents to be completed as per recommendation ELA 2.5.

**Responsibility:** Data Governance and Ethics Committees, TRE staff and researchers

### 11.2.9   ELA 9: Template text should be available from TREs covering the range of standard controls and processes which could be applied to streamline the approval process for researchers and Data Governance and Ethics Committees

Many TREs provide Standard Operating Procedures (SOPs) or template text which covers the controls that they apply to ensure personal data is protected. Researchers can utilise such information within application forms, for example by directly referencing the SOP rather than listing the controls in their own words. Data Governance and Ethics Committees who review many applications which refer to such SOPs are not required to re-review the SOPs for every new application as they are already familiar with the text. This reduces the effort on behalf of the researchers and reviewers as well as reduces the number of re-submissions of applications due to missing information.

*Table 1- High-level example risk template*

| | Model openly available and shared openly | Model embedded within the software with legal controls around hacking | Model embedded within the software with legal controls around hacking and limits on the number of queries | Model only accessible via queries to a secure web-server – limiting controls on the number of queries and users |
|---|---|---|---|---|
| Passes attack simulations to check for identifiable data and TRE checks | 🟧 | 🟩 | 🟩 | 🟩 |
| Passes attack simulations to check for identifiable data and TRE checks and use of Safe Wrappers | 🟧 | 🟩 | 🟩 | 🟩 |
| Fails attack simulations to check for identifiable data and TRE checks | 🟥 | 🟥 | 🟧 | 🟩 |
| Differential Privacy Instance-based models | 🟩 | 🟩 | 🟩 | 🟩 |
| Instance-based models e.g. SVM | 🟥 | 🟥 | 🟥 | 🟩 |

We recommend that the additional controls required for processing trained ML models are also provided within SOPs or template documents. This will help to streamline the Data Governance and Ethics application processes.

**Responsibility:** TRE staff; **Understanding:** Data Governance and Ethics Committees, and researchers

### 11.2.10 ELA 10: TREs should agree to confidentiality agreements should the researchers have concerns re their IP

As per recommendation TECH 8, researchers should share details on their inputs to their model and details of their training method to support TREs to efficiently run attack simulations. If researchers have concerns regarding the confidentiality of the TRE relating to their IP, then confidentiality agreements should be considered between the TRE and the researchers.

**Responsibility:** Data Governance and Ethics Committees, TRE staff and researchers

### 11.2.11 ELA 11: The Data Controller should approve the MDC processes applied by TREs

This information could just be included as links to SOPs within the data governance applications (as is often the case for non-ML projects) or could be pre-agreed as a set of approved processes.

**Responsibility:** Data Governance and Ethics Committees; **Understanding:** TRE staff and researchers

## 11.3  Additional recommendations

So far, our recommendations only focus on the issues specifically associated with risks and controls for protecting personal data (essentially the role of the TREs). In this section, we identify other aspects of ethics or legal aspects where we feel **additional recommendations** are required but are **out of this narrower scope**:

- Researchers should consider the issue of group privacy (and not only individual privacy) in the context of ML model release from TREs.
- TREs should carefully consider what additional technical and ethics/legal controls will be needed for updating existing ML models. TREs should carefully consider the checks, risks and possible controls needed if there is the possibility of allowing (knowingly or not) ML models initially trained on external data into the TRE (transfer learning using pre-trained models).
- Researchers should be made aware of their legal responsibilities in addition to the Data Protection Act 2018 and UK GDPR e.g. Computer Misuse Act 1990, Fraud Act 2006, National Security and Investment Act 2021, Human Rights Act 1998, Equality Act 2010.
- Researchers should be made aware of the current content of the National Security and Investment Act 2021, and considering the lack of specific regulation for machine learning models in current legislation, such a model could be considered a "qualifying asset" and thereby subject to the current vague terms of this Act and the voluntary provision on liaising with the Secretary of State in terms of a "voluntary notification procedure". An amendment to the provision of the Act could be considered in terms of allocating mandatory responsibilities in the event of a need for the notification of a trigger event (to take the place of the "voluntary notification procedure [52]") to private and public bodies working with Machine Learning Models trained Personal Data. This Act allows the Secretary of State to screen and prohibit "potentially hostile" investments that threaten UK national security. This Act has been described as "a sledgehammer to crack a nut [53]."

# 12  Costing

## 12.1  Costing Background

TREs generally work using a cost recovery model. They may charge for a range of different services such as:

- the work required by data analysts to extract and pseudonymise relevant data for the specific research project (this is often the case if the TRE is also the same organisation that owns the data)
- providing a linkage service as a trusted third party
- providing expertise on the data sets
- providing the secure hardware and software infrastructure for researchers to analyse the data without being able to be released without going through disclosure MDC and limiting access to the internet

- providing a disclosure control service where trained members of the TRE team assess files which researchers would like to release out of the environment for personal data.

Providing support for research projects training ML models will increase the support requirements above that of a 'normal' research project and as such these costs need to be considered.

## 12.2 Costing Recommendations:

### 12.2.1 C 1: TREs should charge for the additional work to undertake disclosure control of trained ML models and run attack simulations

TREs should estimate the additional effort to support the controls required to ensure personal data is not encoded within models to be released and to run attack simulations on models. It may be that for the first few projects that TREs support these additional costs are considerable (see additional funding requirement below, but over time this will become routine practice and result in efficiencies expected of mature processes).

**Responsibility:** TRE staff; **Understanding:** Data Governance and Ethics Committees, and researchers

### 12.2.2 C 2: Additional funding should be made available to support TREs to develop the tools and frameworks to support ML training

The design and development of new processes to support the recommendations listed within this document will take time and funding. We recommend that there are infrastructural funds made available for TREs to seek to support them to provide this new functionality to the community.

**Responsibility:** Funding bodies

### 12.2.3 C 3: The costs for limiting queries on the model should be considered

TREs could provide a service, such as a secure web service, to host the trained model and provide the security required for querying the model based upon the required controls. Such controls may include limiting the IP addresses which could submit a query, limiting the number of queries over a period of time, and guarding against denial-of-service attacks. Service level contractual agreements may be required. Such a service could also be provided by a trusted third party or the organisation of the researcher. Whether the service is provided by a TRE, a trusted third-party or the researcher organisation, the costs of providing the service would need to be covered by the researcher. The costs of designing, implementing and running the service should be considered.

**Responsibility:** Researchers; **Understanding:** Data Governance and Ethics Committees, and TRE staff

### 12.2.4 C 4: TREs should consider the additional costs of maintaining the data and development pipeline to support legal requirements e.g. for a certified medical device

There may be regulatory or legal requirements to keep a copy of the data and pipeline used to generate the trained ML for audit purposes for several years e.g. for medical devices such information may have to be kept for 15. TREs may have to consider the cost of supporting these requirements.

**Responsibility:** TRE for implementation. Researchers for the cost.

##### 12.2.5  C 5: TREs should consider outsourcing some of the highly technical work

Although highly trained experts are needed to support research projects involving ML model training, substantial expertise at each TRE is not necessarily required. The expectation is that ML releases are likely to be relatively rare (compared to traditional statistical analysis which generates multiple release requests daily). It may be efficient for TREs, particularly those who do not have many ML users, to share ML checking expertise, calling it in when needed. Some TREs already provide "peer review" access to projects, which could be a mechanism for a pool of trained checkers to share expertise. However, even in the case of outsourced checking, it would be advisable for TRE staff to have a basic conceptual understanding of ML modelling.

**Responsibility:** TRE staff.

# 13  Training

## 13.1  Training Background

Previous work [54] showed that uncertainties amongst TRE staff limited the uptake of ML modelling, including an understanding of how to carry out disclosure control on trained ML models. This project aimed to provide the tools and clear guidance to allow TRE staff to have confidence in releasing models. Use of those tools/guidance requires a good understanding of ML modelling, and so there is a need for training TRE staff in the specifics of assessment.

## 13.2  Training Recommendations

#### 13.2.1  TR 1: Training courses and documentation should be developed for TRE staff on ML, how to run attack simulations and the risks of disclosive data within trained models

These courses should provide sufficient detail for the TRE staff to both carryout tests and advise researchers on how to use the tools and guidance. The trained TRE staff should be able to engage with researchers as equals in ML. In line with current output checker training, this course should also cover how to engage with the researcher to build positive communications. As noted above (Costing section), not all TRE staff need to go through such a course (and possibly none if output checking is outsourced).

#### 13.2.2  TR 2: Introductory training courses and documentation should be developed for TRE staff on ML

All staff should be aware at an introductory level of the issues around ML models, how these are resolved, and who to contact for further advice (i.e. the specialist ML output checker). The purpose of this is to ensure that all TRE staff know how to process ML model outputs, even if they are not capable of running the tests themselves.

Such training should include:

- The risks of disclosure of personal data from trained models.
- The controls available to them to mitigate these risks such as
  - Attack simulations run by the TRE to check for unsafe practice
  - Use of safe wrappers [25]
  - How to limit queries to the model
- Legal and ethics implications

**Usage:** TRE staff

### 13.2.3  TR 3: Training courses and documentation should be developed for researchers on the risk of disclosure control when training models and should consider controls and legal and ethics components

Such training should be a requirement for access to the data for ML model training projects. Again, the courses should follow good practice in engaging researchers in the 'why' of output checking, to build a positive community of interest. Output checking is most efficient [40] when both the researchers and the checkers understand and agree on the basic questions: what is to be checked, how will it be checked, and why is it being checked? Researchers are unlikely to have considered the disclosure risk of models or be aware of tools developed for checking. Therefore, efficient output checking of ML Model requires the training of researchers.

Researchers should be made aware of the different types of contractual agreements and use that could apply to the deployment or release of a trained model, in terms of an EULA or a data sharing/transfer agreement. This is important as in the future there may be different categories of contracts required for different transactions (research/commercial). This is also important when a model is released under an open-source license (e.g., MIT license [55]). Additional clauses may be required to be added to such open-source licences to prohibit attacking of the model to disclose personal data and prohibit use for a different purpose other than the intended purpose. This could be implemented through a 'linked agreement' (a good example of how to draft this can be found in International Data Transfer Agreement published by the Information Commissioners Office [14]) or 'dual-licence'.

For commercial researchers, recognising and understanding that contractual agreements (including any linked agreements) must contain fair, reasonable and non-discriminatory (FRAND) [56] clauses is crucial. This clear understanding will help the interaction between technologists and lawyers in the drafting of these linked contractual agreements navigating machine learning model practices, which in turn, will help prevent or settle any potential disputes that may arise from unfair contract terms included in a subsequent contractual agreement (e.g. licenses, medical device agreements, patent licenses and trade secrets). Such training should include:

- The risks of disclosure of personal data from trained models.
- The controls available to them to mitigate these risks such as
    - o  Attack simulations run by the TRE to check for unsafe practice
    - o  Use of safe wrappers
    - o  How to limit queries to the model
- Legal and ethics implications

This training will help researchers to provide relevant information in their applications which will be assessed by Data Governance and Ethics boards/committees.

**Usage:** Researchers

### 13.2.4  TR 4: Training courses and documentation should be developed for Data Governance and Ethics Committee members to assess applications which include the training of ML models

Such training should include:

- How to assess the risks and controls of disclosure of personal data from trained models
- How to assess the risks and controls on the limitations on queries to the model

- Legal and ethics implications

Developing such training in combination with the target groups will help to ensure that the training is relevant and useful, as well as highlight issues which the training team might not have considered.

As Data Governance Committee members include Data Controllers who are responsible for approving the use of their data, such training should help to identify and forestall any concerns that Data Controllers may have.

Alternatively, Data Governance and Ethics committees could consider having applications reviewed by an expert(s) in assessing the disclosure risk from trained ML models who would then advice the committees.

**Usage:** Data Governance and Ethics process teams.

## 14 PPIE

### 14.1 PPIE Background

We established a PPIE Group, with 8 people with an interest in health data research. We worked to ensure that across the PPIE members there is diversity including, but not limited to gender, ethnicity, age and geography. The PPIE group met virtually 5 times throughout the project to discuss the challenges of machine learning and disclosure risk and how these challenges could be addressed. We covered the technical, legal and ethical challenges. Across the workshops, several tools were used to drive engagement with the group, including Menti, word clouds, online videos, project member pre-recorded videos and graphics. The PPIE discussions informed our recommendations.

The lay co-applicants were actively included in all workstreams and consulted on all recommendations.

### 14.2 PPIE Recommendations

#### 14.2.1 P 1: Public representatives should be involved in data governance and the ethics approvals process

**Responsibility**: Data Governance and Ethics process teams

#### 14.2.2 P 2: The use of data for training ML models and the controls on the models should be visible to the public through searching data use registers

**Responsibility**: Data Governance and Ethics process teams & TRE

## 15 Future Research and Development

The GRAIMATTER sprint project has identified many recommendations and provided initial research to support the safe training of ML models within TREs. However, the time and resource constraints of this project have left many areas for future research. This is a non-exhaustive list of areas we would like to focus on as part of future work:

**General:**

1. Testing the recommendations across a range of TREs and ML projects.
2. Development of training materials for researchers, TREs and Data Governance and Ethics Committees.

3. How the recommendations might be made more efficient in practice e.g. what are the runners/repeaters in ML models? Should we treat everything as a 'stranger' needing expert review until we get more experience? To what extent can mandating the use of 'safe wrappers' alleviate the need for TRE technical skills in implementing attacks, and checking for "runners/repeaters"? How can more expertise/ knowledge be embedded within the report generating process?

4. Assessment of trusted parties to host models – semi-disclosive model.

5. Commercial components, such as whether the recommendations would be acceptable to the industry and the views of the public in working with the industry.

6. Quantification of the risk appetite of Data Controllers.

7. Further investigation into health economics – analysis of the costs to support AI within a TRE.

8. Further investigation into benefit sharing if a company accesses public data to develop ML model.

9. Further investigation into the pay-for model for consented datasets.

10. Investigation into if additional controls are required for group disclosure from both technical and ethical/legal perspectives.

**PPIE:**

11. Work with the public to assess their opinion on what they consider to be disclosive.

12. Qualitative and quantitative PPIE research in the area.

13. Investigation of the possibility to refine correct metrics for risk influence by discussing with public representatives to understand their risk appetite e.g. evaluating if it was acceptable to risk disclosure of 1 person while benefitting 99 others.

14. Public education and outreach.

15. Additional research into trust and public benefit.

**Technical:**

16. Development of additional Safe Wrappers and the tool kit.

17. Development of more informative and more easily understandable risk metrics that balance information about the 'global' risk (e.g. mean accuracy of inference attacks) vs. the 'local' risk-specific subgroups of people.

18. Developing an improved understanding of membership inference risks, (and how to measure them), for unstructured data such as images and text etc., where it is inappropriate to focus specifically on exact combinations of feature values. In other words, under current definitions, a 'perfect' attack would say that someone's data was not in the training set if we changed one pixel in an image, or one decimal place in the calculation of BMI, both of which change continuously and are subject to measurement uncertainty. So we need a new understanding of membership that accounts for the inherent uncertainty in feature gathering. To explore and implement multiple criterion decision analysis by combining various evaluation metrics to quantify disclosure.

Further technical investigation into:

19. the risks and controls required to handle transfer learning.

20. the strengths and weakness of MQCs.

21. how the recommendations might be made more efficient in practice e.g. what are the runners/repeaters in ML models? Should we treat everything as a 'stranger' needing expert review until we get more experience? To what extent can mandating the use of 'safe wrappers' alleviate the need for TRE technical skills in implementing attacks, and checking for "runners/repeaters"? How can more expertise/ knowledge be embedded within the report generating process? how non-TREs can benefit from proposed recommendations.

22. how to balance the needs of researchers to have all the data vs enabling "worst-case" attacks – i.e. exploring the use of synthetic data (from the same distribution) as a proxy.
23. the possibility of water-marking the data to identify any data leakage while exporting the model.
24. the implications of federated learning.
25. the implications of regular updating of models, economical as well as technical.
26. the implications of disclosure control for genetics data.
27. the implications of disclosure control for imaging data.
28. encryption of the model and key management.
29. the possibility of using homomorphic encryption such that encrypted data is used in training the model instead of the original.
30. the impacts on the performance of the model when safe wrappers are used.
31. the methods through which TREs remain up-to-date with the challenges, ready to mitigate new risks and run new attack scenarios.
32. methods for stratifying/analysing disclosure risks for different sub-groups, and then the approach to making sure that certain groups are not more vulnerable to disclosure of their data.

**Legal and ethical:**

33. Consideration of the creation of a new regulated profession with categories for different industries, e.g. one could be the accredited NHS researcher, (see Goldacre Review)
34. A review of the disclaimer and warranty clauses of open-source licenses of the commonly used tools in GitHub Libraries / chosen commercial tools.
35. Consideration of crafting a codified "SAFE TRE Practice" or "SAFE Model Practice" with a view to it becoming a statutory framework for future legislation in AI or a sub-set of AI regulation called, Machine Learning, for example.
36. The lack of mention of "anonymous" practices in the current legislative provision (Data Protection Act 2018), albeit in a work-in-practice Code format provided by the Information Commissioners Office, needs to be addressed.
37. The necessity for clauses to be added to Open-Source Licenses including restrictive clauses on the lifecycle of the ML Model.
38. More dynamic models e.g. ethics audits for Data Governance and Ethics Approval Processes which follow a project's lifetime (and beyond) for AI-related projects without posing disproportionate additional burdens on researchers, TREs, Data Governance and Ethics Committees.

# 16 Appendix A: Risk Assessment of AI Models and Hyperparameters

The evaluation of risks associated with releasing trained machine learning (ML) models from TRE, is part of the Factor Analysis of Information Risk (FAIR) assessment. It considers the potential privacy risks and negative users (attackers and malicious) of the model. It assesses the potential vulnerability to a data leak in releasing the model.

ML models have several components as shown in Figure 8:

- *Architecture:* one can imagine it as a skeleton. It's how the model is structured, in the same way as there are many types of animals, and they all have unique skeletons, the same with models.
- *Training data:* it can be thought of as the filling of the skeleton: muscles, skin and so on.
- *Hyper-parameters*: can be thought of as the clothes that this person (or model) we are creating is the right size and style for the person/model. An overfitted model would be when the clothes are made to measure and only fits the specific individual and no one else, in which case the model generated is no good. Also, some parameters may not work well with some data, just as some styles of clothes are simply not appropriate for a certain type of event.

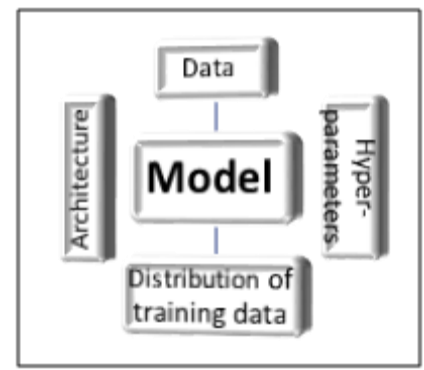

*Figure 8. Components of a Machine Learning model.*

- *Distribution of training data*: it can be thought of as the characteristics of the data, the maximum, minimum, average, if there are many outliers, etc.

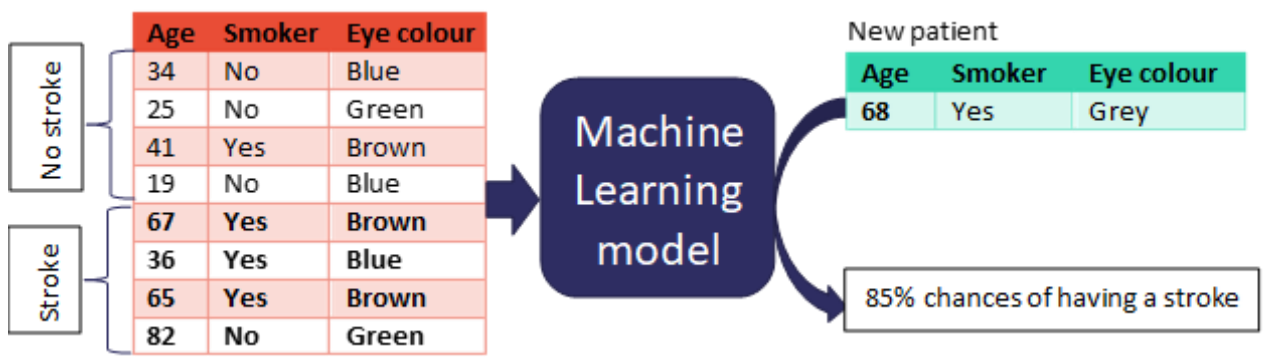

*Figure 9. Example of a Machine Learning model.*

Let's create an example. Imagine we want to predict whether patients will have a stroke or not. We can take health records of patients who did not have a stroke and patients who did and train the model with the selected data. Each row in the table corresponds to one patient. In a machine learning context, the columns are referred to as features. In the example (Figure 9) above, a model with 3 features is constructed (age, smoker, eye colour). Some features will be more relevant to the prediction than others. In the example below we can see that age and smoker might be informative for the prediction, but that eye colour probably isn't.

To query the model, it is necessary to have a row with the same type of data (same features) used to train the model which, in this case, would be age, smoker and eye colour. The model will be able to predict whether this new patient will suffer a stroke or not.

The minimum required for an attack is that the attacker or adversary needs to be able to query the model. The attacker will send rows of input data (age, smoker, eye colour) and the model will respond with either a predicted class (stroke / non-stroke) or scores (probabilities) associated with each class. The adversary will typically try and use this process (possibly repeated multiple times) to learn something about the personal data that was used to build the model.

White box attack                                      Black box attack

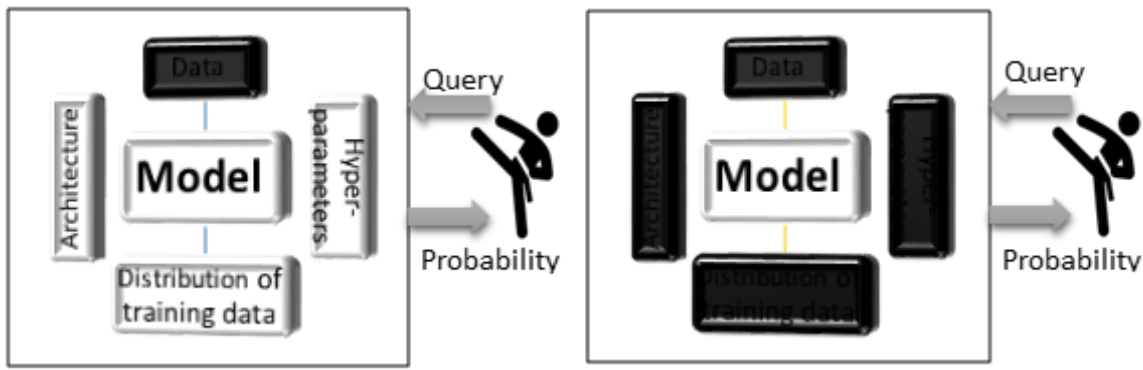

*Figure 10. Type of attacks.*

The situation in which the attacker can only query the model is known as a *black box* attack. Sometimes an attacker might have more access and be able to examine the model itself. This is known as a *white box* attack. The kind of information available in white- and black-box attacks is depicted in Figure 10 above.

Membership Inference Attacks (MIA) are when an adversary or attacker is trying to determine whether a set of input values (they have access to or have generated) are part of the original training dataset of the target model (referred to as target train). Our objective in assessing a model was to measure the highest potential MIA attack accuracy and, establish criteria in which models with identifiable data were safe.

Our simulated attack experiments proceeded as follows. After pre-processing, the target dataset was split into 3 equal parts: train, shadow and test. The split was repeated 5 times, varying the rows included in each part. For each classifier of interest, a set of values of hyperparameters to be explored was defined. Five target models were created for each combination of classifier hyperparameters, one for each data split. All of our experiments were on health record data.

Several attack scenarios were defined to determine the risk of personal data leak from ML models: Worst Case, Salem 1, Salem-synth and, Salem 2. The *Salem 1, Salem-synth* and *Salem 2* scenarios are based on adversary attacks described on Salem [37]. Table 2 contains a summary of their main characteristics.

*Table 2 Attack scenarios main characteristics.*

| Scenario | *Salem 1* | *Salem-synth* | *Salem 2* | *Worst case* |
|---|---|---|---|---|
| **Shadow model** | Same classifier as target model, same hyperparameters. | | | NA |
| **Shadow data** | Split of the target data (held out). | Simulated set from target data. | Unrelated to the target (breast cancer). | NA |
| **MIA model** | Random Forest Classifier | | | |
| **MIA data** | Shadow model predicted the probabilities of the shadow data. | | | Target model predicted the probabilities of the target train and target test data. |

The *Worst-Case* is a white box scenario, and it does not need any shadow model. It is described in Rezaei [57] as the easiest possible for the attacker. To perform a MIA, a new binary set of data (member, non-member of target train data) was created containing the predicted probabilities of the training and test data by the target model. A Random Forest Classifier was fitted with half of this new set. This scenario was not supposed to simulate a realistic attack (if the attacker has access to the data, they do not need to attack) but instead to assess whether there were potential vulnerabilities in the model that could potentially be leveraged by an attacker. This can give a good estimation of the maximum capability of an attacker to succeed. In some cases, the risk of data leakage could be overestimated, but it does guarantee (as much as possible) that any ML model allowed out of a TRE is safe. At the same time, it's easy to implement (see Figure 11).

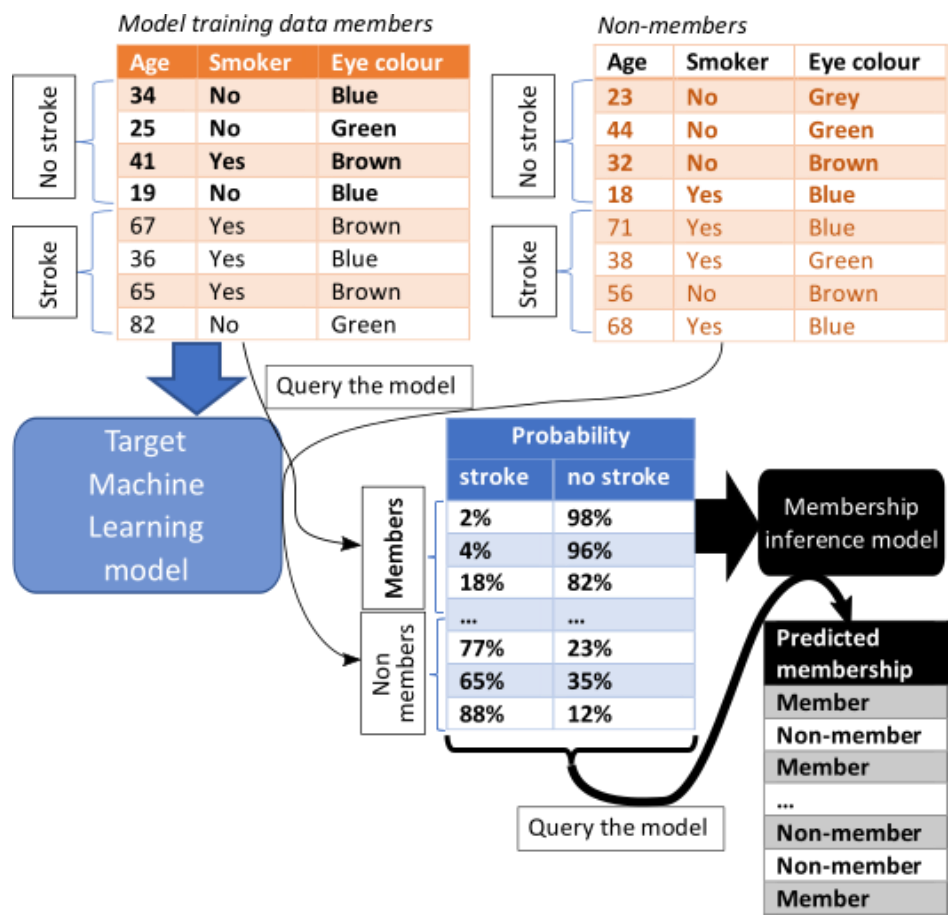

*Figure 11 Worst-case scenario diagram*

The three *Salem* scenarios work in the same way and need a shadow model, represented in Figure 12. The attacker does not have access to the data used for training and testing the ML model. The attacker can only query the model and obtain probabilities, and the attacker needs to create or find data for the "shadow model" to perform the attack.

The same combination of the target classifier with identical hyperparameters as the target model was used to generate the shadow model with the shadow data. In *Salem 1* the data to train and test the shadow model was the shadow data split from the target dataset; in *Salem-synth* the shadow data was synthesised from the target dataset (still data from the same distribution); and *Salem 2* used an unrelated set of different distribution, which was breast cancer data available in the python package scikit-learn. In all cases, the shadow data was split into two equal parts, train, and test shadow. Half of the predicted probabilities by the shadow model of the target train and test shadow data were used as a new set to train a binary classifier (Random Forest).

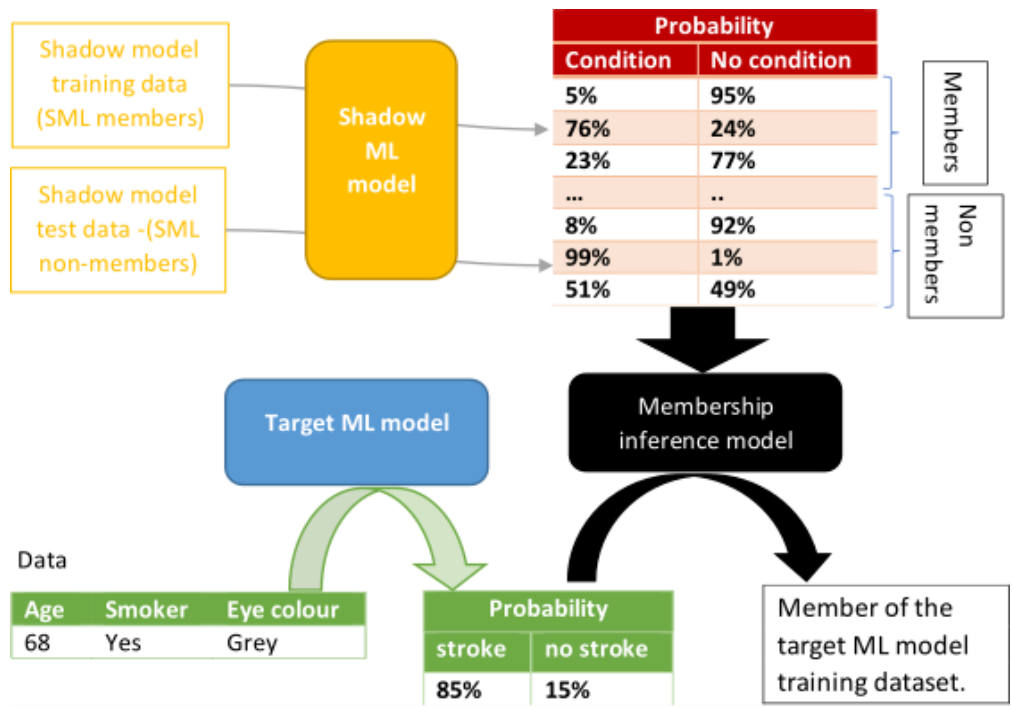

*Figure 12. Salem attack scenarios diagram.*

In all the scenarios (Figure 11 and Figure 12), the MIA model predicted whether a data point belongs to the target model training data and was validated with the test MIA set.

The predictions obtained from the target, shadow and MIA models were used to generate a confusion matrix and subsequently calculate the metrics of True Positive Rate (TPR), False Positive Rate (FPR), False Alarm Rate (FAR, also known as False Discovery Rate - FDR), True Negative Rate (TNR), Positive Predictive Value (PPV), Negative Predictive Value (NPV), False Negative Rate (FNR), Accuracy (ACC), F1Score, Advantage – which measured the capability of the attacker to distinguish if a data point belongs to the target training set [58] and was measured as the absolute value of the difference between TPR and FPR, and their corresponding probabilities were used to calculate AUC accordingly, FIDIF and the corresponding p-value (PDIF) which measured extreme cases.

The experiments were performed over several open tabular health datasets. For each dataset, the scenarios were compared with 7 target classification models (Ada Boost, Decision Tree, Random Forest, Logistic Regression, SVC (linear and rbf kernels), DPSVC and XGBoost). In each case, the membership inference attack classifier was a Random Forest with default sklearn parameters. Target model hyper-parameters were varied over a large range (between 12 and 3840 target models per classifier were created) and various attack metrics were calculated. In all, a total of 1,066,950 experiments were performed.

The results presented here include only those models which were considered to be "good target models", following the criteria of AUC>=0.75, TPR>=0.75, and TNR>=0.75. The figures below show the MIA AUC of gamma parameter of SVC with rbf kernel (Figure 13) and the metrics of MIA TPR averaged by classifiers and grouped by scenarios and (Figure 14). Our results highlight variability depending on the classifier and the target dataset. The Worst-Case fulfils the purpose of capturing the worst possibility of attack amongst the scenarios tested.

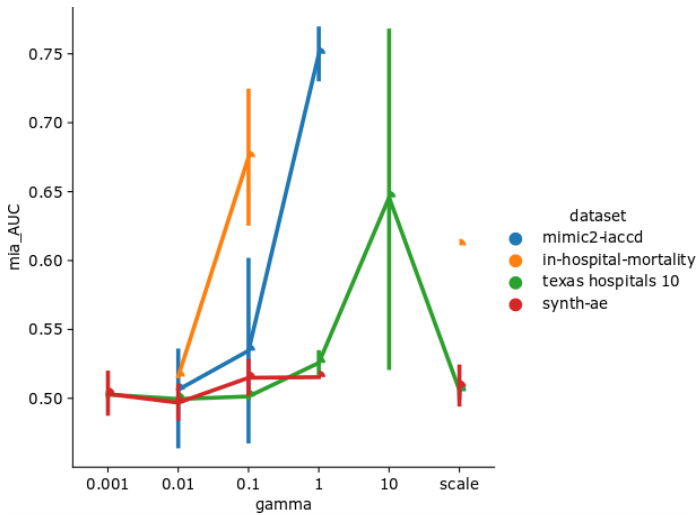

*Figure 13. Gamma parameter (SVC rbf kernel) MIA AUC by dataset (Worst Case scenario).*

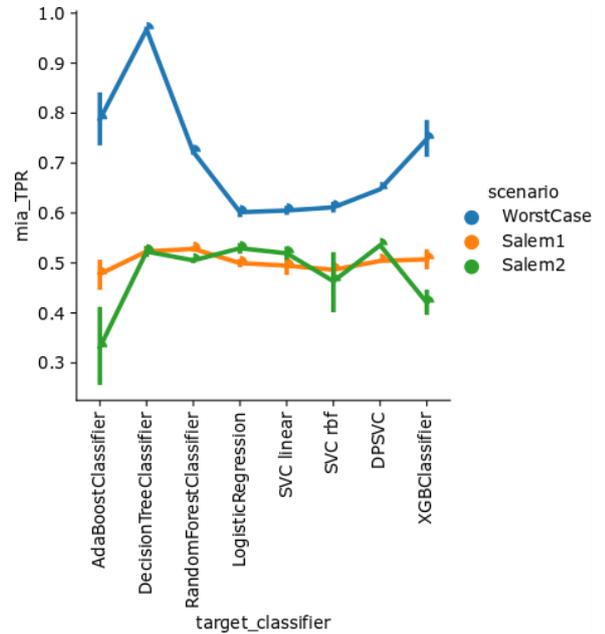

*Figure 14. Classifiers MIA TPR by scenario.*

## 16.1 Experiment 1: Scenario comparison

A comparison of the scenarios was performed to determine which scenarios ought to be included in an attack suite for the TRE staff. In particular, the scenarios were compared to find the scenario which gave the most conservative risk assessments.

The experiments showed that, on average, the Worst-Case scenario had the most conservative risk estimates. A summary of the results is shown here. We show Worst-Case versus Salem 1 (Salem 2 had slightly lower attack performance than Salem 1 in some cases, see Figure 14). Here we show three metrics: the membership inference advantage, the membership inference area under the ROC curve (mia_AUC), and the membership inference F-score (Figure 15, Figure 16 and Figure 17 respectively). In all cases, a higher score means more attack success. Each metric is represented by one pair of plots. The left-hand plot shows a histogram of the difference in the metric between the Worst Case and Salem 1 scenarios. A general positive trend suggests the Worst-Case scenario consistently gives higher risk values. The right-hand plot shows the metric for the Worst-Case scenario (x) versus the metric for Salem 1 (y). The key point from this plot is the absence of points in the upper-left quadrant (where Salem 1 would indicate riskiness and Worst-Case would not).

The following additional conclusions can be drawn from these results:

Even classical ML algorithms (decision trees, random forests, Support Vector Machines) can be vulnerable to membership inference attacks, as can be seen by points in the upper right quadrants of the right-hand plots.

There are many situations in which the Worst-Case scenario identifies risk when Salem 1 does not. It would be dangerous to release such models as they clearly have a degree of vulnerability, even if current attack methods would be unable to expose them.

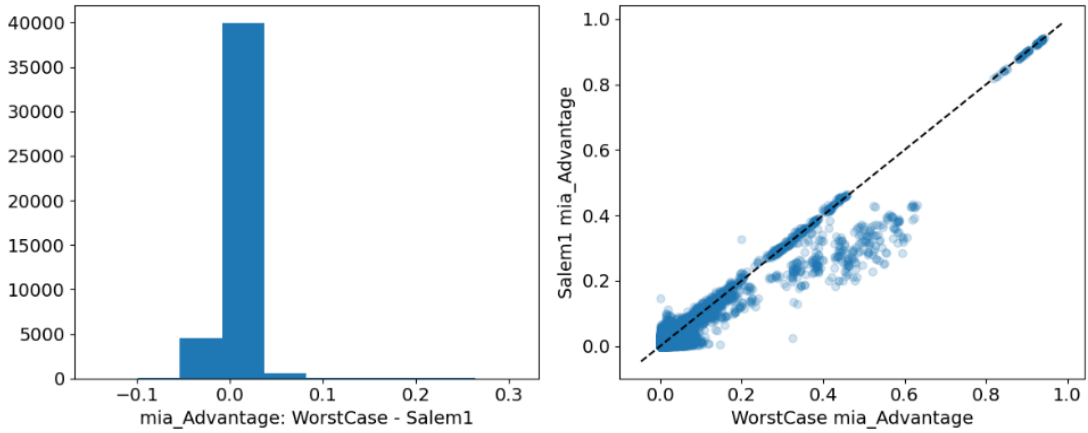

*Figure 15. Advantage comparison of the worst-case versus Salem 1 scenarios.*

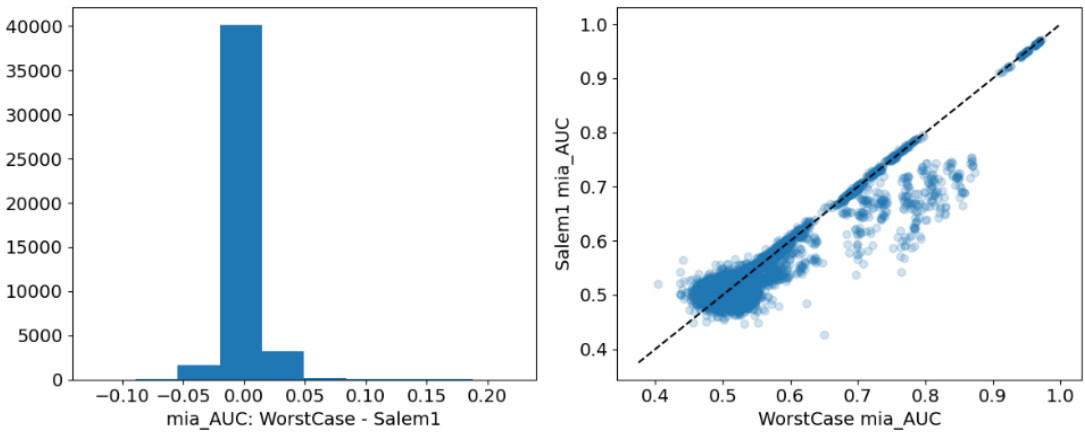

*Figure 16. AUC comparison of the worst-case versus Salem 1 scenarios.*

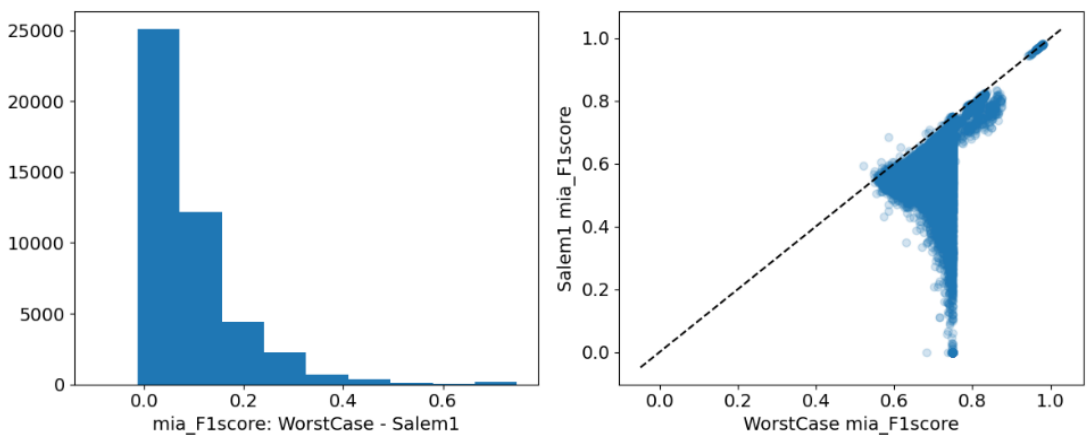

*Figure 17. F1 score comparison of the worst-case versus Salem 1 scenarios.*

## 16.2 Experiment 2: Risk analysis

In our second suite of experiments, we investigated the extent to which the riskiness of a set of hyper-parameters is generalised over datasets. This is an important consideration for TRE staff. If riskiness does generalise across datasets, then one could be confident that a model was unsafe based on experiments with that model on previous datasets, reducing the burden on TREs to perform their experiments.

We investigated a wide range of hyper-parameter configurations across a range of popular ML models for each of our datasets and then computed the minimum and maximum value of a range of attack metrics across the datasets. This resulted in, for each metric, minimum and maximum attack metrics across datasets and we were interested in identifying if there were hyperparameter values which were safe in one dataset and then very risky in others. Note that all datasets were of a similar type – tabular data.

The results showed that for popular model choices, there exist many hyperparameter values that appear safe for one dataset, but highly unsafe for others, suggesting that it would not be sufficient for TRE staff to rely upon previous analysis to determine if a new model was safe. Hence, it is advisable to run post-hoc attacks (or implement MQCs) on the specific model requested to be released.

Example results for the Random Forest Classifier are shown below (Figure 18). In each plot, the minimum value of the metric is plotted on the x-axis and the maximum on the y. Each point is a hyper-parameter configuration. Configurations with risk performance that generalises between datasets would have very similar minimum and maximum values and therefore exist on the y=x diagonal. We can see that although this is the case for some, there are many for which it is not. In particular, there are many cases where the attack AUC ranges from 0.5 (unsuccessful) to 1.0 (totally successful).

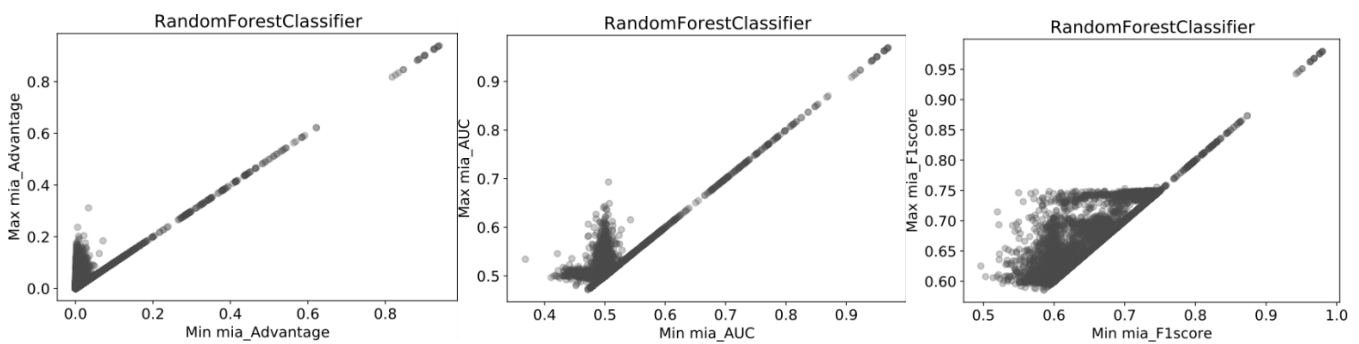

*Figure 18. Correlation between minimum and maximum MIA advantage, AUC and F1score.*

It is worth noting that although we can see that the riskiness of hyper-parameter configurations (at least as measured by the AUC metric) does not generalise overall, there appear to be many examples that are either always risky within the context of our experiments (top-right), or never risky (bottom-left), for all subfigures in Figure 18. Parameter configurations identified as being risky in at least one experiment can be used to form the ranges to be used by the Safe Wrapper classes (TECH 10).

## 16.3  Attribute Inference attacks

The experiments above refer to a framework for investigating Membership Inference attacks. A similar investigation was performed to explore the vulnerability of algorithm-hyperparameter-dataset combinations to Attribute Inference Attacks (AIA) and to investigate the relationship between these two different types of risk.

Informed by Experiment 1 above, we designed an AIA that assumes an attacker has access to a partial data record and the 'true' label, but that the value of one attribute for that record is missing.

- If the attribute is categorical, we then completed copies of the record with each possible value, and noted the predicted label, and the confidence (predicted probability for that label). If there was a unique attribute value that leads to a prediction with the highest confidence, the attack model output that label as its prediction, otherwise, it did not make a prediction.
- If the attribute is continuous, we estimated the lower and upper bounds on the prediction that an attacker could make as follows. First, we took $n$ (e.g. 100) samples from across the range and used the target model to make predictions using those samples in place of the missing attribute value. Next, we recorded the lowest and highest values where the trained model has confidence equal to its maximum. Finally, we marked the record as 'at risk' if the upper and lower bounds were both within k% of the true value for that attribute.
- Finally, for each attribute, we constructed risk measures, and provided comparisons of these for the training set, and a previously unseen test set.

To investigate the relationship between hyper-parameter values of vulnerability to attribute inference, we defined the following risk metric for an attribute, equivalent to a worst-case attack scenario used above:

1. Using the training set used to train the model, calculated the proportion of the samples for which the appropriate attribute (categorical or continuous) made a correct prediction. Denote this P_vulnerable(train).
2. Using the test set (I.e. not seen by the model) calculated the same proportion: P_vulnerable(test)
3. Reported the "Attribute Risk Ratio"(ARR) = P_vulnerable(train) / P_vulnerable(test)
   - If ARR was 0 then the predictions cannot be made for the attribute being considered.
   - If 0<ARR<= 1, then attribute values could be correctly predicted for some data, but this was no more likely to be the case for data 'seen' by the model than 'unseen'. In other words, the model was reflecting the general characteristics of the data which was allowing inferences to be made.

As ARR rises above 1.0, then correct inferences are increasingly more likely to be made for data 'seen' by the model than data not. In other words, the model is encoding something specific and non-generalisable about the training data and should be considered disclosive. The plots below, Figure 19, show typical results from a random forest classifier trained on the in-Hospital Mortality Data. Top plot shows for each continuous variable, the proportion of records (P_vulnerable) that can be estimated within +/- 10% of their true value, separated by training and test data. As can be seen for some (but not all) attributes the value of P_vulnerable(train) was significantly higher than P_vulnerable(test).

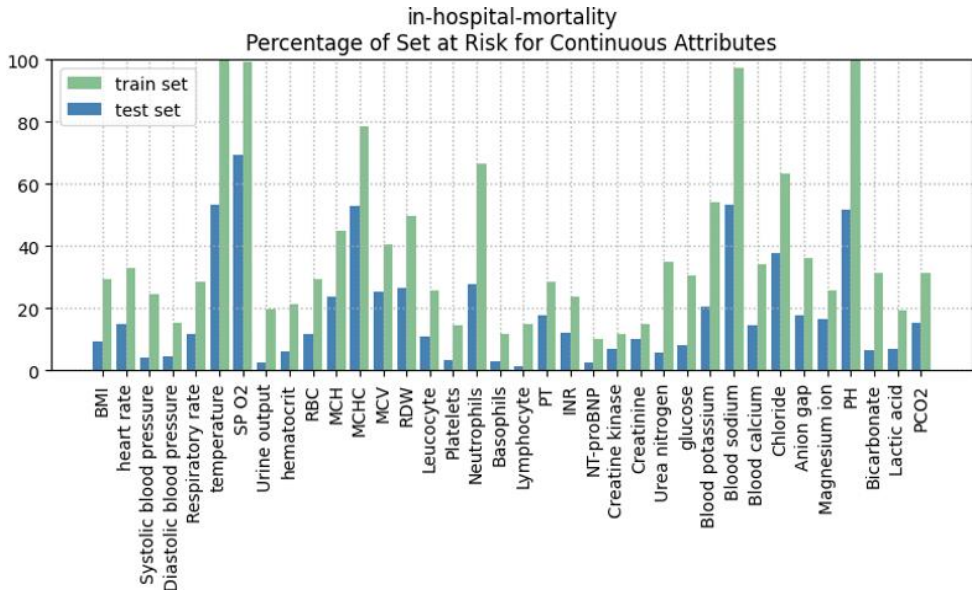

*Figure 19. Percentage of risk by continuous attributes*

The next plot, Figure 20, shows, for categorical variables, the proportion of records where the attack model would exactly predict the missing value of an attribute. Note this shows examples of both Attribute Risk Ratio below 1 (e.g. attribute *CHD-with-no-MI*) and above (e.g. *hypertensive*).

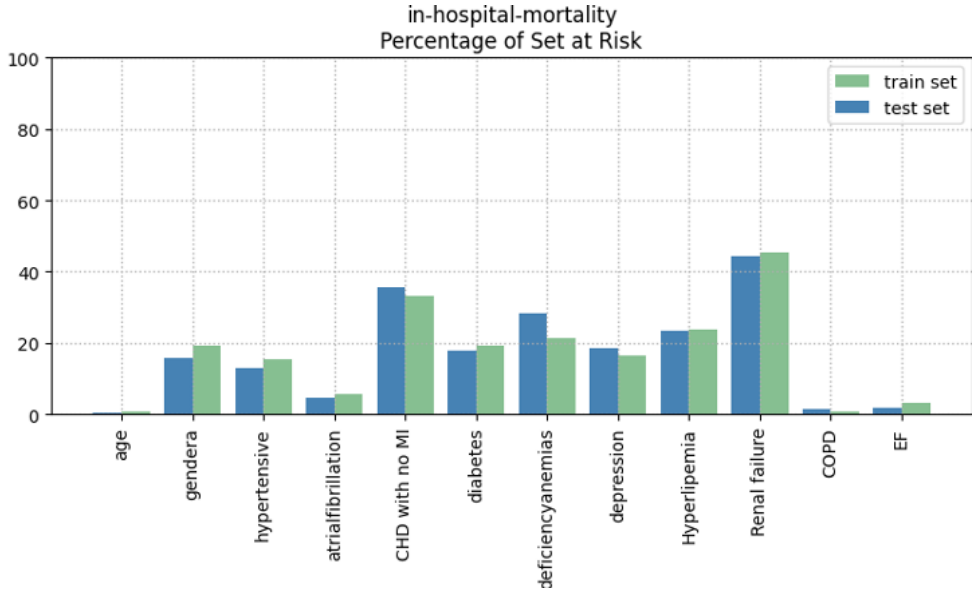

*Figure 20. Percentage of risk by categorical attributes*

The final plot, Figure 21, shows the improvement over a 'most frequent value' estimator achieved by the attack model, in the cases where it does make a prediction.

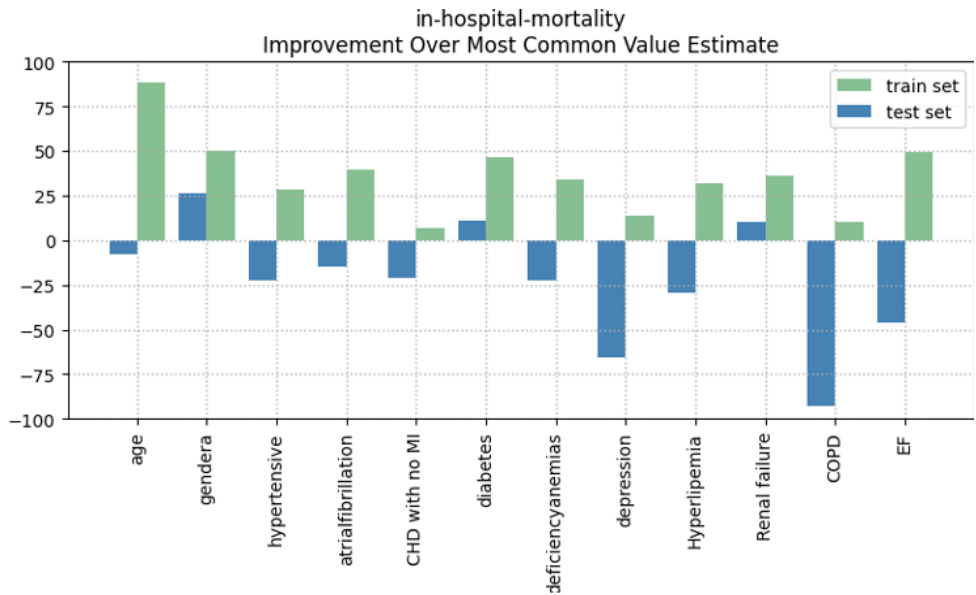

*Figure 21. Attack model improvement over most common values estimate*

These initial results demonstrate that AIAs can pose a significant threat to some trained models.

We subsequently explored the effect of different hyper-parameter and dataset choices as per the MIA work. This was rendered more complex by the fact that we could in effect build a separate model predicting risk from hyper-parameters for each attribute. Using a binary 'AIA risk value' (ARR) (i.e. Attribute risk ratio above 1.0 or not) we trained Decision Tree Classifiers to predict ARR from hyper-parameter values.

Our analysis revealed what appear visually to be some clear patterns of behaviour. For example, the risk of attribute inference appeared to be heavily influenced by the number and maximum depth of trees for a Random Forest Classifier. In the future we would like to explore ways of characterising attributes in a dataset to generalise risk predictions across groups of attributes.

## 16.4  Predicting Vulnerability Risks

Based on the outputs of WP1 and 2, we have a large archive of results with thousands of rows each representing a specific trained model, and fields (columns) denoting the dataset, model type, specific hyper-parameter values, metrics of the trained model's performance (accuracy metrics) and vulnerability (MIA and AIA attack metrics). Based on our analysis we used a combination of four rules (based on thresholds of attack metrics) to create a binary "vulnerable to MIA" field for each record, and a similar process to flag vulnerability to attribute inference. Note that in contrast to 16.2, here we focussed on a definition of 'vulnerability' based on FDIF and TPR/FAR metrics (Appendix F).

Results from training ML models to predict whether a given hyper-parameter combination is likely to result in a disclosive trained model were encouraging. On initial experiments with Random Forest as the ML model type, we found:

1: Generalisation across hyper-parameters:

- over 90% weighted accuracy at predicting the 'vulnerable to MIA' class of models,
- over 89% recall (i.e. correct identification) for the 'vulnerable to MIA' class,

- between 87% and 99% weighted accuracy for predicting 'vulnerability to AIA' (values differ for different attributes being attacked)
- between 67% and 100% recall of the 'vulnerable to AIA' class (over 90% for 42 of the 47 attributes tested)
- Both sets of figures were for the test accuracy/recall using a 90:10 % train/split on the same dataset. AIA were for 'in-hospital mortality', MIA results quoted were lowest from 'in-hospital-mortality' or 'mimic2-iaccd'

2: Generalisation across datasets:

- 88% weighted accuracy and 87% 'vulnerable' recall for models trained to predict 'vulnerable to MIA' using 90% of the in-hospital-mortality datasets then tested using all of the mimic2-iaccd
- 89% weighted accuracy and 78% 'vulnerable' recall for the reverse case
- Note these figures would be expected to improve if we trained on all of one dataset to avoid predicting for both unseen hyper-parameter combinations and a new dataset….

We would like to explore these results further as part of a new research project, however, it was clear that:

1. these initial findings appear to substantiate one of the projected benefits of the safe wrapper approach: saving time and money for researchers and TREs by detecting and flagging in advance hyper-parameter combinations which are likely to lead to disclosive models.
2. there is work to be done on helping users understand which hyper-parameter values to change if a warning is triggered, or if an attack on a trained model reveals that the model is disclosive.

# 17 Appendix B: Attack simulation tool kit for TREs staff to use

A SafeModel wrapper class has been developed in python, and versions produced for the following ML model types:

| Model *type* | Base Implementation/Library | Notes |
|---|---|---|
| Decision Tree Classifier | Sklearn.trees.DecisionTreeClassifier | Model also reports k-anonymity value for tree |
| Random Forest | Sklearn.ensembles.RandomForestClassifier | Model also reports k-anonymity value for forest |
| k-Nearest Neighbours | N/A | Safe model rejects attempt to use this algorithm |
| Support Vector Classifier | Sklearn + bespoke | Implements Differential Privacy version on top of sklearn |
| Artificial Neural Networks | Tensorflow Keras functional model | Enforces choice of TensorFlow differentially private optimiser |

Considerable effort has been put into the additional functionality and making this transparent to researchers.

From a researcher's point of view, the principal differences are two extra functions: save_model() and request_release(). The former is self-explanatory, the latter is the most important, and is the function they call to trigger the TRE output checking process.

Behind the scenes, the superclass does the following:

The object constructor init() function checks whether the parameters supplied by the researcher (or the base implementation defaults) match the constraints held in the TRE's rules.json file.

If they do not, it changes them where possible, and warns the researcher that these changes have been made, and any other changes they need to make

Note that the researcher has the option to over-ride these suggestions, and this will be flagged to the TRE output checkers.

The fit() method calls the base implementation's fit() method to train the model. It then saves a 'snapshot' copy of the model's attributes at that stage (or the architecure and weights for tensorflow neural networks)

Where appropriate (e.g. Decision Tree and Random Forest Classifiers it also calculates the k-anonymity of the model

The request_release function does the following checks:

- Do the model's current parameters meet the constraints in the TRE rules.json?
- Have the model's parameters been altered (maliciously or otherwise) between training and a request for release (e.g., to hide an unsafe training regime)?
- Have any other critical parts of the mode been changed since the fit() method was called? (for example the trees in a Decision Tree or Random forest, the support vectors in an SVM or the weight in a neural network)?
- (where appropriate) Was the DP-variant of the optimiser used?
- (where appropriate) What are the differential privacy guarantees (e.g. epsilon values)?
- What is the result of running a worst-case Membership inference attack on this model?
- What is the result of running a worst-case Attribute inference attack on this model?
- What is the result of running a Likelihood Ratio membership Inference attack on this model[41]?

The results of these tests are then collated into a human and machine-readable output file along with some recommendations. Below are some typical excerpts from respectively: a 'safe' model; a model trained with unsafe hyper-parameters; and one trained unsafely then maliciously changed to look superficially 'safe':

{

"researcher": "j4-smith",

"model_type": "RandomForestClassifier",

"model_save_file": "testSaveRF.pkl",

*"details": "Model parameters are within recommended ranges.\n",*

*"recommendation": "Run file testSaveRF.pkl through next step of checking procedure"*

*}*

*{*

*"researcher": "j4-smith",*

*"model_type": "RandomForestClassifier",*

*"model_save_file": "unsafe1.pkl",*

*"details": "WARNING: model parameters may present a disclosure risk:\n- parameter bootstrap = False identified as different than the recommended fixed value of True.",*

*"recommendation": "Do not allow release",*

*"reason": "WARNING: model parameters may present a disclosure risk:\n- parameter bootstrap = False identified as different than the recommended fixed value of True."*

*}*

*{*

*"researcher": "j4-smith",*

*"model_type": "RandomForestClassifier",*

*"model_save_file": "unsafe-malicious.pkl",*

*"details": "Model parameters are within recommended ranges.\n",*

*"recommendation": "Do not allow release",*

*"reason": "Model parameters are within recommended ranges.\WARNING: basic parameters differ in 2 places:\nparameter bootstrap changed from False to True after the model was fitted\nparameter min_samples_leaf changed from 2 to 10 after the model was fitted\n"*

*}*

We are developing a manuscript which details this summary information.

# 18 Appendix C: Example data dictionary template

The following is an example of how information about the encoding of different features could be provided:

data:{

 0: {'name': 'parents', 'indices': [0, 1, 2], 'encoding': 'onehot'},

1: {'name': 'has_nurs', 'indices': [3, 4, 5, 6, 7], 'encoding': 'onehot'},

2: {'name': 'assessment_completed', 'indices': [8, 9, 10, 11], 'encoding': 'onehot'},

3: {'name': 'children', 'indices': [12, 13, 14, 15], 'encoding': 'onehot'},

4: {'name': 'atrialfibrillation', 'indices': [16], 'encoding': 'int64'}

...

}

# 19 Appendix D: Example constraints file

Below is a snippet from a file rules.json that contains examples of how the 'safe envelope' for algorithm parameters can be stored centrally as a set of constraints within a human and machine-readable file.{

```
"DecisionTreeClassifier": {

 "rules": [

  {

   "keyword": "min_samples_leaf",

   "operator": "is_type",

   "value": "int"

  },

  {

   "keyword": "min_samples_leaf",

   "operator": "min",

   "value": 5

  }

 ]
```

```
        },
    "RandomForestClassifier": {
        "rules": [
            {
                "operator": "and",
                "subexpr": [
                    {
                        "keyword": "bootstrap",
                        "operator": "equals",
                        "value": true
                    },
                    {
                        "keyword": "min_samples_leaf",
                        "operator": "min",
                        "value": 5
                    }
                ]
            }
        ]
    },
    "SVC": {
        "rules": [
            {
                "keyword": "dhat",
                "operator": "min",
```

```
      "value": 1000

    },

    {

      "keyword": "C",

      "operator": "min",

      "value": 1

    },

    {

      "keyword": "eps",

      "operator": "min",

      "value": 10

    },

    {

      "keyword": "gamma",

      "operator": "min",

      "value": 0.1

    }

    ]

  },

....
```

# 20 Appendix E: Case studies

This section describes some simple scenarios to illustrate some of the main data privacy risks from machine learning models. To evidence these scenarios, we generated synthetic data (except for the hospital survival example which was based on a publicly released real anonymised data set) and trained ML models on the data. We then ran attack simulations to evidence what personal data we could find by attacking the models.

The data sets used were intentionally kept small for simplicity in explaining the underlying principle. However, the risks presented here apply to both small and large datasets. The first three examples (20.1 -20.3) discussed the risks of potential disclosure before the implementation of measures proposed in this document. Whereas example 20.4 presents a case of identification and prevention of disclosure risk, after implementation of proposed measures. The software code developed for all the scenarios can be found here [16].

## 20.1 Finding out unknown information about a famous person (Attribute Inference)

Early diagnoses of cancer save lives. Researchers from a hospital created a model that was capable of predicting the risk of cancer among people with multiple diseases. As per common scientific practice, the methods and description of the data were published in a specialised journal and the model was publicly available (see Figure 22).

The model was integrated into clinical systems enabling a clinician to query the model and receive a probability of an individual receiving a cancer diagnosis in the future. For example, a patient with a 24% response is at low risk of suffering from cancer, however, a patient with 95% can be identified as a high-risk individual.

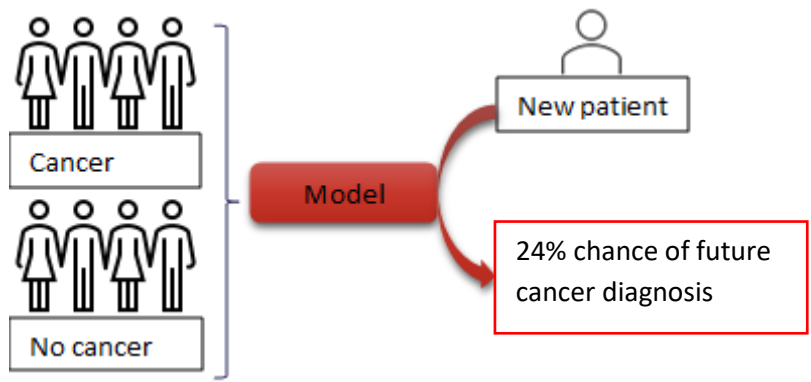

*Figure 22. Model to predict the risk of a future cancer diagnosis.*

The data used to create the model included 200 patients, 50% of which suffered from cancer. Their names and addresses were removed from the data to ensure privacy. The data employed to create the model was hosted in a safe environment and never made publicly available. Table 3 shows example data used to build the model for predicting cancer risk.

The researchers were unaware that the collected data contained the health records of a famous Member of Parliament (MP) who was local to the hospital. Since this MP is well known, a quick search revealed a lot of his health information e.g. he is diabetic, asthmatic, smokes, and is 62 years old

*Table 3 Data used to build the model for predicting risk to suffer from cancer.*

| Diabetes | Asthma | Weight | Blood pressure | Smoker | Age |
|----------|--------|--------|----------------|--------|-----|
| Yes | No | Obese | High | Yes | 72 |
| Yes | Yes | Normal | High | No | 83 |
| No | No | Obese | High | Yes | 63 |
| Yes | Yes | Obese | High | No | 77 |
| **Yes** | **Yes** | **Overweight** | **Slightly high** | **Yes** | **62** |
| Yes | No | Underweight | Normal | No | 50 |
| No | Yes | Normal | Slightly high | Yes | 66 |
| Yes | Yes | Normal | Slightly high | No | 44 |
| No | No | Normal | Slightly high | Yes | 61 |
| No | No | Normal | Slightly high | No | 56 |

Cancer

→ Famous MP

No cancer

A highly skilled technical person realised that this MP's data was potentially used to train the model and therefore it might be possible to *attack* the model to extract additional information about MP's health.

There were 5 different values of blood pressure (low, slightly high, high, very high, extremely high) and 4 different values of weight (obese, normal, overweight, underweight) recorded by the hospital, resulting in 20 possible combinations. The model was queried using the variables already known for the MP along with each possible combination of weight and height and its response confidence was recorded. A property of trained models is that they often provide higher confidence when the inputs are the same as seen by the training set and lower otherwise. It is this property which can be exploited by attackers. Table 4 presented the 5 responses with the highest confidence.

*Table 4 Model response with unknown variables*

| Weight | Blood pressure | Response confidence |
|--------|----------------|---------------------|
| **Overweight** | **High** | **93,8%** |
| Normal | Slightly high | 57,7% |
| Normal | High | 54,4% |
| Overweight | High | 54,2% |
| Obese | Slightly high | 54,2% |

It was evident that the first row had the highest confidence response which made it very likely to be the true values for the MP i.e. BMI group 3 (overweight) and blood pressure group 2 (slightly high). This is an example of an Attribute Inference Attack.

The researchers could have prevented this from happening using a different configuration of the model for training and a much larger training dataset so that there were many more examples of individuals with the same attributes as the MP.

## 20.2 Identifying if someone famous has suffered from cancer (Membership Inference)

Scientists trained a model that predicts whether patients would respond well to a specific drug used to treat cancer based upon multi-morbidities. Administering the appropriate treatment for cancer patients is crucial for recovery and use of this model would save lives. As per common scientific practice, the methods and description of the data were published in a specialised journal and the model was publicly available and easily accessible.

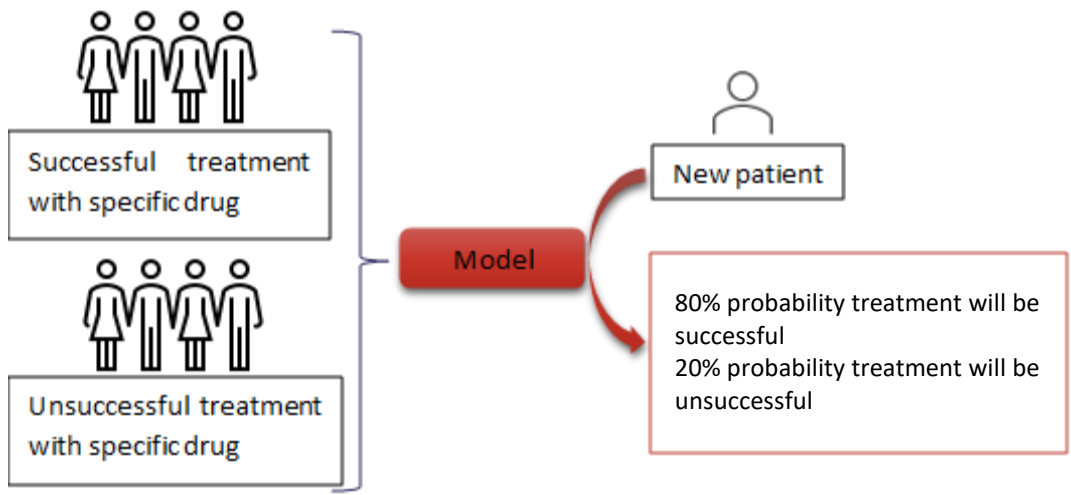

*Figure 23. Machine learning model to predict cancer treatment success.*

Querying the model with details about a specific patient returns the probability the specific treatment will be successful (see Figure 23), and could influence the treatment decision. This is a type of personalised medicine.

The researchers were unaware that the collected data contained the health records of a famous Member of Parliament (MP) who was local to the hospital as shown in Table 5. Since this MP was well known, many properties about him were in the public domain e.g. overweight, slightly high blood pressure, diabetic, asthmatic, smoker, and 62 years old.

What remains unknown is whether the MP suffered from cancer. All participants in the study suffered from cancer, so if it was established that the famous MP participated in the research study it would mean that he suffered from cancer – information that should not have been released.

*Table 5 Data to build the machine learning model to predict cancer treatment success*

| Diabetes | Asthma | Weight | Blood pressure | Smoker | Age | |
|----------|--------|--------|----------------|--------|-----|---|
| Yes | No | Obese | High | Yes | 72 | Successful treatment |
| Yes | Yes | Normal | High | No | 83 | |
| No | No | Obese | High | Yes | 63 | |
| Yes | Yes | Obese | High | No | 77 | |
| Yes | Yes | Overweight | Slightly high | Yes | 62 | Famous MP |
| Yes | No | Underweight | Normal | No | 50 | |
| No | Yes | Normal | Slightly high | Yes | 66 | Unsuccessful treatment |
| Yes | Yes | Normal | Slightly high | No | 44 | |
| No | No | Normal | Slightly high | Yes | 61 | |
| No | No | Normal | Slightly high | No | 56 | |

The following steps could be undertaken by an attacker to attempt to establish if the MP was part of the data used to train the model:

1) generation of some fake information about fictional patients (unidentical to the original set)

2) development of a prediction model as described in the publication (which remains the same as the original, except that it was trained on this fake data)

3) assess the confidence values given by the fake model on examples on which it was trained, and examples on which it was not trained. These confidence values for examples that were and were not used to train **the fake** model let the attacker see if the model gives different confidence values for examples that it saw in training compared to those it didn't see. The attacker sees that it does, and concludes that, given that the model is very similar to the original, the original will too. Because the attacker is a machine learning expert, they decide to use these confidence values to build another model that can predict from them, whether an example was used to train the model or not. The process is illustrated in Figure 24.

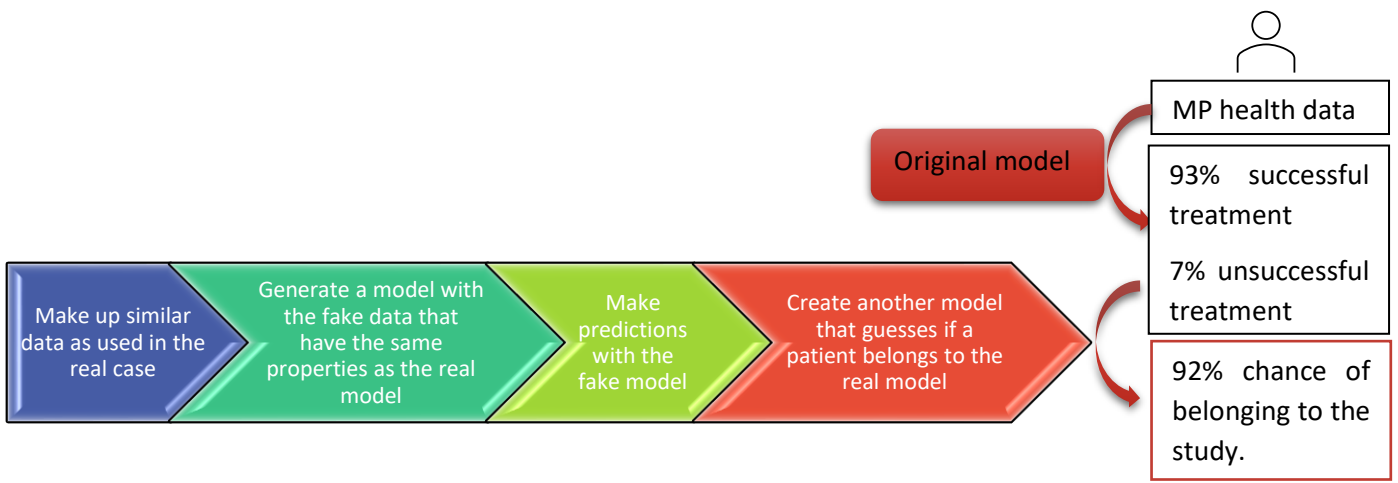

*Figure 24 Process that an attacker needs to follow during an attack.*

To determine if the MP was in the data and therefore suffered from cancer, the attacker would pass the MPs input data through the original model and then pass the confidence value through their attack model to classify it as part of the training set or not, and conclude that they were part of the original training set and therefore have had cancer.

This is an example of a membership inference attack used to breach someone's privacy.

## 20.3  Successful candidates in a job interview (Membership Inference)

A recruitment agency misuses a trained ML model to only offer interviews to individuals who are not likely to be drug users. The intended original purpose of the ML model was not this use and sensitive data was being breached in this use case example of membership inference.

The original purpose of the model was to help drug users who were in serious financial difficulties to improve their quality of life. The study recruits were promised that their data would be kept anonymous. The example of the recorded data is shown in Table 6.

*Table 6 Data used to create a machine learning model to predict insolvency for drug users*

| Name | Age | Education | Sex | Number of previous convictions | Housing | Number of previous rehabilitations | Solvent |
|------|-----|-----------|-----|-------------------------------|---------|-----------------------------------|---------|
| Michelle | 34 | Some secondary | Female | 2 | Other | 0 | No |
| Lindsay | 35 | Secondary | Female | 0 | Supported | 1 | No |
| James | 40 | University | Male | 2 | Other | 4 | No |
| Kenneth | 36 | University | Male | 3 | Other | 1 | No |
| Emily | 32 | None | Female | 1 | No fixed | 2 | Yes |
| Rebecca | 22 | Some secondary | Female | 0 | Supported | 0 | Yes |

The original model was trained and shared openly so that groups such as charities could use the model to assess the risk that a drug user would go into serious financial difficulties and take appropriate steps to support them (see Figure 25).

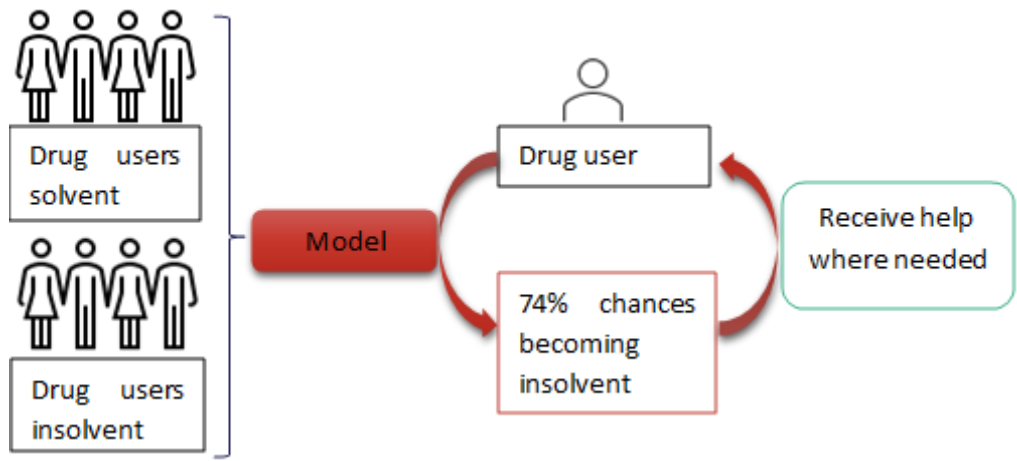

*Figure 25. Diagram of the machine learning model to predict insolvency for drug users.*

*Table 7 Example of response obtained by the model to predict insolvency for drug users*

| Name | Probability insolvency | Part of the study |
|------|------------------------|-------------------|
| John | 75.0% | Yes |
| Ashley | 74.9% | Yes |
| Elizabeth | 25.0% | Yes |
| Angela | 75.0% | Yes |
| Tyler | 74.9% | Yes |
| Jason | 24.9% | Yes |
| Christina | 50.0% | No |
| Jose | 50.0% | No |
| Thomas | 50.1% | No |
| Brittany | 49.9% | No |
| Nicholas | 49.8% | No |
| Megan | 50.8% | No |

The recruitment agency obtained a copy of the model, and knowing details about some of the people (since these people publicly shared that they participated in the study), found that for participants in the study the model returned values that were either around 75% or 25%, whereas non-participants had values of around 50% (Table 7). The agency used this rule to screen candidates and avoid inviting the candidates for a job interview where the model's probability was close to 75% or 25% since they were likely to be participants in the study, and therefore probably drug users. Figure 26 shows the selection process.

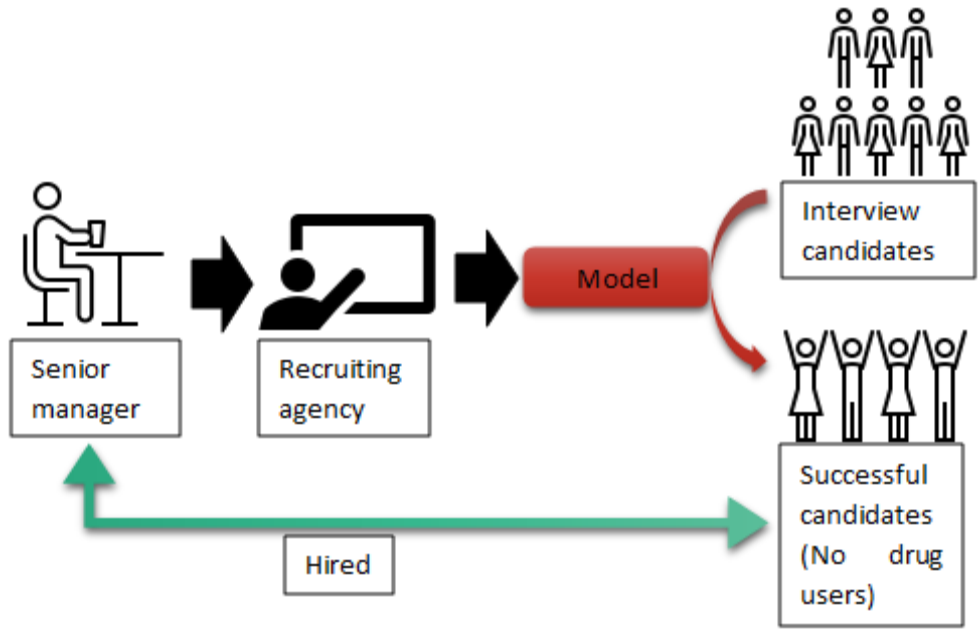

*Figure 26. Candidate recruitment process by employing a machine learning model.*

In this case, the model violated the privacy of people who volunteered to be part of the study and as a consequence were not offered a job interview.

### 20.4 Hospital admission survival (instance-based model vulnerability)

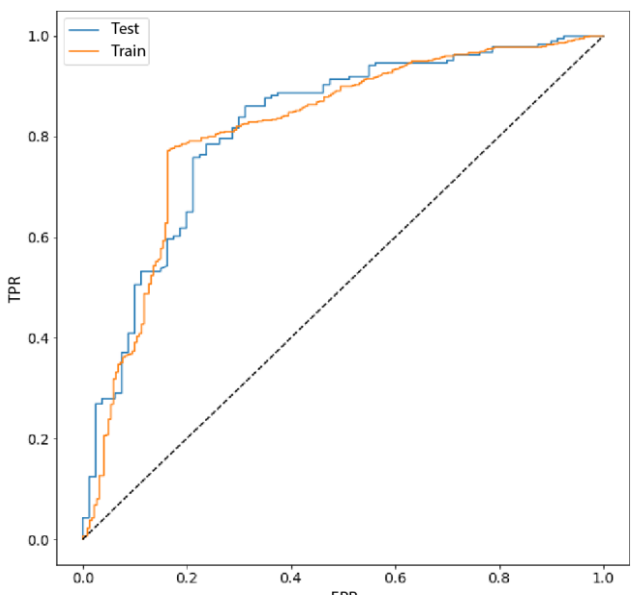

*Figure 27- ROC curve.*

Hospitals wanted to improve the care for admitted patients, by predicting the probability of their survival. A Support Vector Machine (SVM) model was used, which is "instance-based", powerful, and capable of making very accurate predictions and thus highly popular in machine learning. The high performance of the SVM is obtained as shown in the ROC curve in Figure 27. The True Positive Rate (TPR) represents the proportion of correctly identified survival of patients versus the False Positive Rate (FPR), which represents the proportion of incorrectly identified survival. The further the curves deviate from the diagonal black line, the better its performance, suggesting the model could be extremely useful.

After the model was trained within a TRE, the researchers requested an export of the model so that it could be deployed within hospitals. The TRE staff checked the model and found it contained 441 exact copies of original patient information out of a total of 798. The researchers were requested to remove the sensitive data from the model, however, the researchers explained that the model would not work without this information. There were two options proposed: 1) to create another model using a non-instance-based method or a differentially private version of an instance-based model (if the model is to be exported), 2) not to release the model openly but instead add additional controls such as limiting the queries to the model by, for example, hosting the model in a secure web service. This issue could have been addressed much earlier in the development life cycle through discussions between the researchers and the TRE during the project planning phase.

# 21 Appendix F: Lay explanations of metrics

The following descriptions are proposed for use in the report provided to TRE staff when the attack simulation code has been run.

## 21.1 Introduction

Researchers train and release a model. The model was trained on data where each row represents an individual patient or in the case of aggregated data a set of grouped patient information (in the following, "row" always corresponds to data about a single individual). See an example of this type of data in Table 8.

*Table 8 Example of the tabular type of data*

| Diabetes | Asthma | Weight | Blood pressure | Smoker | Age | Cancer |
|---|---|---|---|---|---|---|
| Yes | No | Obese | High | Yes | 72 | Yes |
| Yes | Yes | Normal | High | No | 83 | Yes |
| No | No | Obese | High | Yes | 63 | Yes |
| Yes | Yes | Obese | High | No | 77 | Yes |
| Yes | Yes | Overweight | Slightly high | Yes | 62 | Yes |
| Yes | No | Underweight | Normal | No | 50 | No |
| No | Yes | Normal | Slightly high | Yes | 66 | No |
| Yes | Yes | Normal | Slightly high | No | 44 | No |
| No | No | Normal | Slightly high | Yes | 61 | No |
| No | No | Normal | Slightly high | No | 56 | No |

The researcher might need to extract the labels which later will be used to construct the model. The labels will be the column identified as "cancer". To evaluate the model, it is essential to split the data in training (data shown to the model) and test (data kept aside for evaluation).

The researcher or TRE output checker will perform some attack simulation to test the level of disclosure before the model release. During this process, an attack system is built that will enable them to predict whether any given patient information (full row of data as shown in Table 8), was used to train the model or not. For example, a process of attack can be done as explained in Appendix A, i.e. Figure 11, which consists of using the training and test data to query the model and their output will be used to generate an attack model.

## 21.2  Probability

The attack system can assign a score to each row. When the score is 1, the system is certain that the row was part of the training data otherwise it assigns score 0.

The attacker will make a decision based on this score by thresholding the score. If the score exceeds the threshold, the attacker concludes that the row was in the training set. If it doesn't, the attacker concludes it was not.

Useful questions to ask are "How probable is it that a row **was** from the training set if the attacker concludes that it was?" and "How confident is the attacker about their conclusion being right?". If this probability is close to 1, the attack is very successful – the vast majority of examples that the attacker predicts as belonging to the training set do. If the probability is close to zero, the attack is unsuccessful. To compute this probability, we need to assume an unknown quantity. We assume that either the attacker has access to the rows that they are making predictions about, or are generating them themselves based on some knowledge of the underlying dataset. The quantity we do not know to calculate our probability is "what proportion of the rows the attacker has access to / is generating" are from the training set. We will call this quantity A. The proportion that is not will therefore be equal to 1 – A.

In most real attack situations, this quantity will be very small. For example, if an attacker is generating synthetic data in the hope that they will be generating some that perfectly match things in the training sets, the success rate is likely to be very small. If (for whatever reason) they have access to a large set of these rows (maybe data for Scotland) and are interested in their attack to find which of those rows were in the training set, the value will still be low (assuming that the training set is a small proportion of the population). One sensible way of setting this value might be as the proportion of the total possible rows (e.g. the population) used for training. Our probability is calculated by combining the true positive and false positive rates.

## 21.3  True and False Positive Rate

True Positive Rate (TPR) = the proportion of the examples that **were** in the training set that the attacker correctly predicts as being in the training set.

False Positive Rate (FPR) = the proportion of the examples the attacker has that **were not** in the training set that the attacker incorrectly classifies as being in the training set.

Armed with these quantities, we can compute the probability of interest (P):

P = (A x TPR) / (A x TPR + (1 – A) x FPR)

Here are some examples (Table 9):

*Table 9 True Positive Rate, False Positive Rate and Probability for a proportion of data.*

|   | A | TPR | FPR | Probability |
|---|---|-----|-----|-------------|
| 1 | 0.5 | 0.6 | 0.4 | 0.54 |
| 2 | 1/100 | 0.6 | 0.4 | 0.01 |
| 3 | 1/100 | 0.8 | 0.2 | 0.04 |

Considering the third row on – this could correspond to a scenario where the attack model performs well (high TPR, low FPR) but because the actual training set constitutes a small proportion of the overall data (1/100 = one in 100 individuals in the population are in the training set) the probability that an example that the attacker predicts as being in the training set is only 0.04 (i.e. 4%).

## 21.4 AUC (Area Under the ROC Curve)

Consider presenting two rows to the attack system. One of the rows was from the training data, and other was not. One way of assessing the performance of the attack system is to ask "What is the probability that the system will give a higher score to the row that *was* from the training data?". This probability is known as the AUC[1]. An AUC of 1 means that the system **always** gives higher scores to rows that were in the training data than rows that were not. The ability to build an attack system with an AUC of 1 suggests that the original model was highly vulnerable to attack. By way of contrast, an of 0.5 means that if given two rows, the system gives the row that was in the training data a higher score half of the time – it might as well be tossing a coin. This would be of no use to an attacker.

There is no single threshold above which an attack AUC can be considered bad. However, we can compute, for a given number of rows that were and were not in the training set the distribution of AUC values we would expect by chance and can therefore compute the probability of obtaining an AUC **higher** than that observed in the attack by chance.

We report the AUC as well as the probability of achieving an AUC of that value or better by chance.

## 21.5 FDIF (Frequency DIFference)

AUC provides a useful *average* performance metric. However, from an attack perspective, average performance is not always the most relevant metric. An attack which is unsuccessful on average may be successful at the extremes.

As described above, AUC gives the probability that the attacker's model will give a row from the training data a higher score than a row that was not from the training data. Note that the AUC doesn't mention any particular score values, it's all about relative scores (higher and lower). An attack model could have poor average performance but perform well in the extremes: say that if the score is above 0.9 the model is always correct. This kind of situation could represent a successful attack but might be missed by AUC.

The following thought experiment can help us understand this. Consider an oracle who has access to 100 rows that were used for training and 100 that were not. The oracle picks one of each and passes it to the attacker and asks the attacker which is which. The attacker uses their attack system to score each row and tells the oracle that the higher scoring row was the one from the training set.

Now, let's consider scenario 1, here the scores that the attack system gives are completely random. The attacker will give the correct answer 50% of the time. The AUC will, on average, be 0.5, and we can calculate the range of AUC values we might expect – three standard deviations from the mean (0.5) gives a range of values from 0.38 to 0.62.

Now scenario 2. Here the attack model is such that 10 rows were in the training set for which the model score is exactly 1. And ten rows that were not for which the model scores exactly 0. The other 180 rows all result in random scores greater than 0 and less than 1. If the oracle picks a row from one or both of these sets of 10, the model will be correct (if it picks one of the 10 that were in the training set that gives a score of 1, the score for the row that didn't have to be lower (and vice versa). The probability of picking one or other of these rows is 0.19. The probability of not picking one or the other is 0.81. The AUC is therefore, on average, 0.19 * 1 + 0.89 * 0.5 = 0.595. This is higher than scenario 1 but crucially, it's well within the range we might expect by chance (see above).

---

[1] Note that AUC is a probability calculated on the basis of all the data, not just the vulnerable examples. As such, like all ways of aggregating lots of data into one number, it loses a lot of information.

Situations like scenario 2 call for a new metric which we call FDIF. FDIF looks at all the scores and computes the difference in prevalence between true training rows in the top 10% (we can use other percentages) and the bottom 10%. If the attack model is completely random, we'd expect to see, on average, the same number of true training rows in the top and bottom 10% and FDIF would be equal to 0. In scenario 2 above, FDIF would be 1. We report FDIF and a p-value that tells us how likely the observed FDIF value would be if the scores are completely random.

## 21.6  Other useful metrics

### 21.6.1  Accuracy

Simply the proportion of test examples that the attack model classifies correctly (0-1).

Let's imagine that our example model is presented with a set of 10 dogs and 10 muffin images. And the predictions obtained are:

| | | Reality | |
|---|---|---|---|
| | | Dog | Muffin |
| Predicted | Correct | 8 | 9 |
| | Incorrect | 2 | 1 |

$$\text{Accuracy} = \frac{Correctly\ identified}{Total} = \frac{8 + 9}{8 + 9 + 2 + 1} = \frac{17}{20} = 85\%$$

The result from the previous table tells us that the class *muffins* are predicted correctly 90% of the time and class *dogs* 80%, and the overall accuracy is 85%.

### 21.6.2  True Negative Rate

It can be interpreted as the proportion of examples that were not in the training set that were correctly classified. Also known as the specificity (0-1). From an attacker's point of view results closer to 1 are good.

### 21.6.3  False Alarm Rate

FAR can be interpreted as the proportion of things incorrectly classified as being in the training set (0-1). From an attacker's point of view results closer to 0 are good.

### 21.6.4  Advantage

The advantage of an attribute inference adversary is its ability to infer a target feature given an incomplete point from the training data, relative to its ability to do so for points from the general population. The Advantage is defined as ABS (TPR - FPR) (0-1). From an attacker's point of view results closer to 1 are good.

# 22 Appendix G: Steps for running attack simulations

The steps for running attack simulations recommended are as follows:

- Passing the training data through the trained model to obtain predictive probabilities.
- Passing the data held out by the TREs through the trained model to obtain predictive probabilities.
- Attempting to build a new model that can predict whether a particular example is in the training data or the held-out data. This simulates a Worst-Case membership inference scenario (Appendix A) where the attacker has access to training and non-training data. Note that this is not necessarily a realistic attack scenario, but serves to provide a conservative assessment of the model's vulnerability.
- This will result in a series of metrics that describe the membership inference risk, that would be interpreted by the TRE staff.

An equivalent worst-case attack for attribute inference would be: Pass each training set record through the trained model with different values for that attribute, and note whether there is a unique value for that attribute that yields the highest predictive probability and whether that value is the 'true' attribute value for that record. Report the percentage of the training set for which there is a unique attribute value that gives the highest confidence and the accuracy of those inferences. Repeat for the hold-out data and report on the differences. This description applies to categorical variables. Similar attacks for continuous variables work based on the upper and lower bounds of the prediction attack.